\documentclass{article}

\RequirePackage{amsthm}
\usepackage{authblk}
\usepackage[T1]{fontenc}
\usepackage[margin=1.5in]{geometry} 
\usepackage[utf8]{inputenc}
\usepackage{amsmath}
\usepackage{amssymb}
\usepackage{ulem}
\usepackage{mathtools}
\usepackage{url}
\usepackage{multicol}
\usepackage{bbm}
\usepackage{tabularx}
\usepackage[colorlinks=true,citecolor=black]{hyperref}
\usepackage{enumitem}
\usepackage{colortbl}
\usepackage{ulem}
\usepackage[usenames,dvipsnames]{xcolor}
\usepackage{tikz}
\usepackage{tikz-cd}
\usetikzlibrary{trees,automata,positioning,calc}
\usepackage{cleveref}
\usepackage{subcaption}
\usepackage{array}
\usepackage{multirow}
\usepackage{diagbox}
\usepackage{natbib}

\newcommand{\ntrees}{B}
\newcommand{\replace}{\texttt{replace} }
\newcommand{\nodesize}{\texttt{min\_nodesize} }
\newcommand{\width}{\ensuremath{W}}
\newcommand{\depth}{\ensuremath{\text{D}}}
\newcommand{\card}[1]{\ensuremath{\# #1}}

\newcommand{\mtryrandomcart}{\texttt{mtry\_random\_cart}}
\newcommand{\mtrycartcart}{\texttt{mtry\_cart\_cart}}
\newcommand{\mtryrandom}{\texttt{mtry\_random}}
\newcommand{\mtrymode}{\texttt{mtrymode}}
\newcommand{\mtrymodeaf}{\texttt{fixed}}
\newcommand{\mtrymodenf}{\texttt{not-fixed}}

\newcommand{\cartcart}{\texttt{include\_cartcart}}
\newcommand{\rootcell}{\ensuremath{\mathfrak{t}} }

\newcommand{\simcartcart}{\texttt{include\_cartcart}}
\newcommand{\simntrees}{\texttt{num\_trees} }
\newcommand{\simwidth}{\texttt{width} }
\newcommand{\simmtryrc}{\mtryrandomcart }
\newcommand{\simmtrycartcart}{\mtrycartcart}
\newcommand{\simmtryrandom}{\mtryrandom}
\newcommand{\simreplace}{\texttt{replace}}
\newcommand{\simminnodesize}{\texttt{min\_nodesize}}

\newcommand{\zwei}{\ensuremath{\mathbf{S}}}

\newcommand{\nsimsvar}{ \ensuremath{N} }
\newcommand{\nsims}{ \ensuremath{100} }
\newcommand{\modelPureType}{\text{(pure-type)}}
\newcommand{\modelHierarchical}{\text{(hierarchical)}}
\newcommand{\modelAdditive}{\text{(additive)}}
\newcommand{\modelPureZwei}{\text{(pure-2)}}
\newcommand{\modelPureDrei}{\text{(pure-3)}}

\newcommand{\dsimvar}{\ensuremath{d}}

\newcommand{\simsamplesize}{ \ensuremath{500} }
\newcommand{\rsrf}{\text{RSRF}}
\newcommand{\simrsrfnf}{\texttt{RSRF}}
\newcommand{\simrsrfaf}{\texttt{RSRF (af)}}
\newcommand{\simrf}{\texttt{RF}}
\newcommand{\simif}{\texttt{INTF}}
\newcommand{\simet}{\texttt{ET}}

\newcommand{\simmeanY}{\texttt{mean-Y}}
\newcommand{\simeinsnn}{\texttt{1-NN}}

\newcommand{\robot}{\texttt{robot}}
\newcommand{\airfoil}{\texttt{airfoil}}
\newcommand{\calhousing}{\texttt{california housing}}
\newcommand{\abalone}{\texttt{abalone}}
\newcommand{\concrete}{\texttt{concrete}}

\newcommand{\simrsrfnfRealdata}{\texttt{RSRF}}
\newcommand{\simrsrfafRealdata}{\texttt{RSRF (af)}}
\newcommand{\simrfRealdata}{\texttt{RF}}
\newcommand{\simifRealdata}{\texttt{INTF}}
\newcommand{\simifnoreplaceRealdata}{\texttt{INTF (noreplace)}}
\newcommand{\simetRealdata}{\texttt{ET}} 
\newcommand{\simetnoreplaceRealdata}{\texttt{ET (noreplace)}}
\newcommand{\simrfnoreplaceRealdata}{\texttt{RF (noreplace)}}

\newcommand{\simrsrfnfnoreplaceRealdata}{\texttt{RSRF (noreplace)}}
\newcommand{\simrsrfafnoreplaceRealdata}{\texttt{RSRF (af) (noreplace)}}

\newcommand{\N}{\ensuremath{\mathbb{N}}}

\newcommand{\R}{\ensuremath{\mathbb{R}}}   
\newcommand{\E}{\ensuremath{\mathbf{E}}}   
\newcommand{\PP}{\ensuremath{\mathbf{P}}}  
\newcommand{\1}{\ensuremath{\mathbbm{1}}}

\newcommand{\dx}{\ensuremath{\mathrm{d}}}

\newcommand{\cov}{\ensuremath{\mathbf{Cov}}}
\newcommand{\corr}{\ensuremath{\mathbf{Corr}}}

\newcommand{\quot}[1]{\text{``#1''}}

\makeatletter
\newcommand*{\rom}[1]{\expandafter\@slowromancap\romannumeral #1@}
\makeatother

\newtheorem{theorem}{Theorem}[section]

\newtheorem{proposition}[theorem]{Proposition}

\newtheorem{definition}[theorem]{Definition}

\newtheorem{remark}[theorem]{Remark}

\makeatletter                              
\def\blfootnote{\xdef\@thefnmark{}\@footnotetext}
\makeatother

\setenumerate{label={\normalfont(\alph*)}}

\allowdisplaybreaks

\newcolumntype{C}{>{$}c<{$}} 
\newcolumntype{D}[1]{>{\centering\let\newline\\\arraybackslash\hspace{0pt}}m{#1}}

\providecommand{\keywords}[1]
{
	\small	
	\textbf{\textit{Keywords---}} #1
}
\title{\textbf{Pure interaction effects unseen by Random Forests} }

\author[a,1]{Ricardo Blum}
\author[b,2]{Munir Hiabu}
\author[a,3]{Enno Mammen}
\author[a,4]{Joseph~T.~Meyer}
\affil[a]{Institute for Mathematics\\
	Heidelberg University\\
	Im Neuenheimer
	Feld 205\\
	69120~Heidelberg,~Germany}
\affil[b]{Department of Mathematical Sciences\\
	University of Copenhagen\\
	Universitetsparken 5\\
	2100~Copenhagen~Ø,~Denmark}
\affil[1]{e-mail: \texttt{ricardo.blum@uni-heidelberg.de}}
\affil[2]{e-mail: \texttt{mh@math.ku.dk}}
\affil[3]{e-mail: \texttt{mammen@math.uni-heidelberg.de}}
\affil[4]{e-mail: \texttt{josetmeyer@googlemail.com}}

\date{}
\begin{document}
\maketitle

\begin{abstract}
		Random Forests are widely claimed to capture interactions well. However, some simple examples suggest that they perform poorly in the presence of certain pure interactions that the conventional CART criterion struggles to capture during tree construction. We argue that simple alternative partitioning schemes used in the tree growing procedure can enhance identification of these interactions. In a simulation study we compare these variants to conventional Random Forests and Extremely Randomized trees. Our results validate that the modifications considered enhance the model's fitting ability in scenarios where pure interactions play a crucial role.
\end{abstract}
{\keywords{random forests;regression tree;cart;pure interaction;functional anova}}

\section{Introduction}\label{sec:introduction}
\begin{figure}[h]
        \centering
          \begin{minipage}[c]{0.42\linewidth}
          \includegraphics[width=\linewidth]{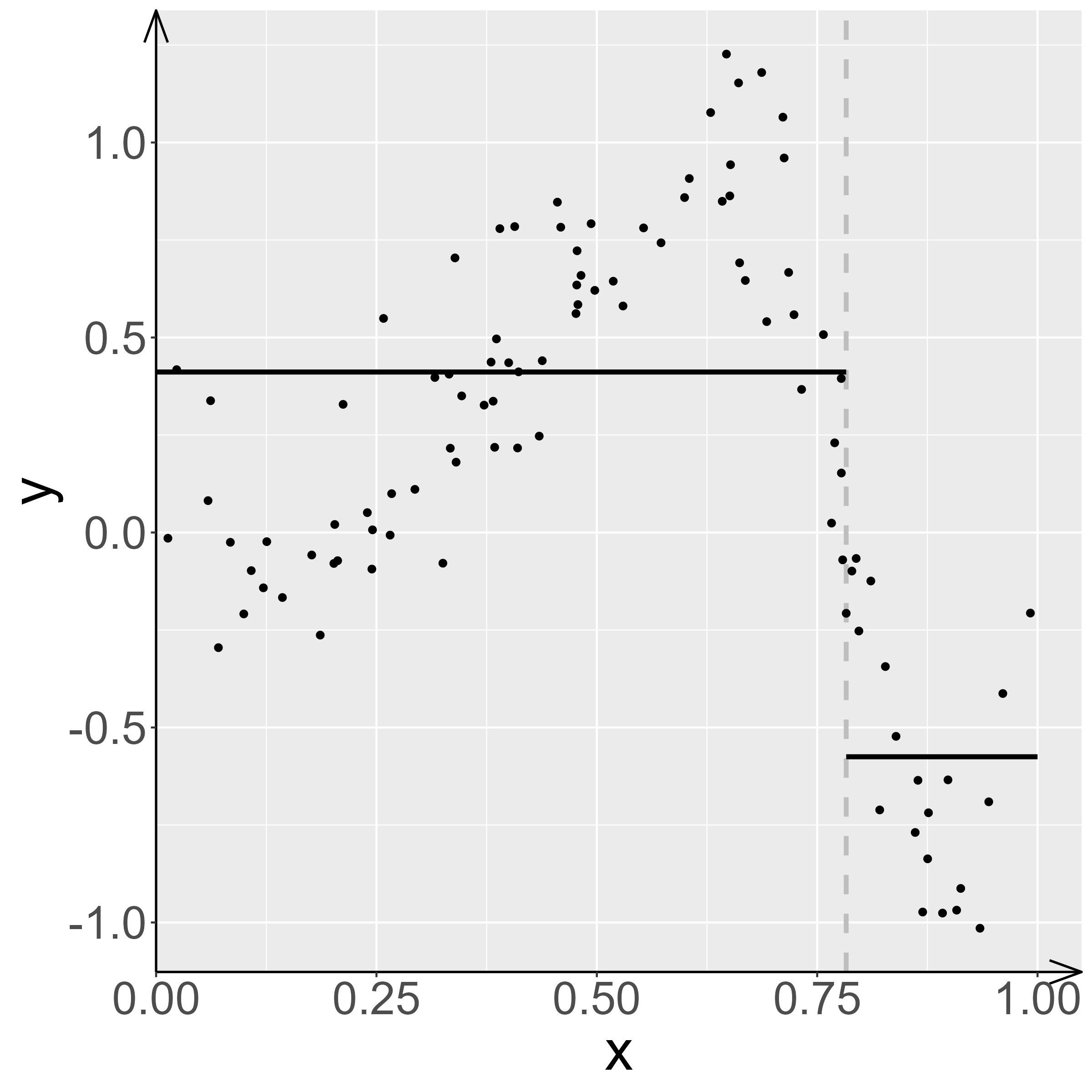} 
          \end{minipage}
          \hspace{1cm}
    \begin{minipage}[c]{0.42\linewidth}\includegraphics[width=\linewidth]{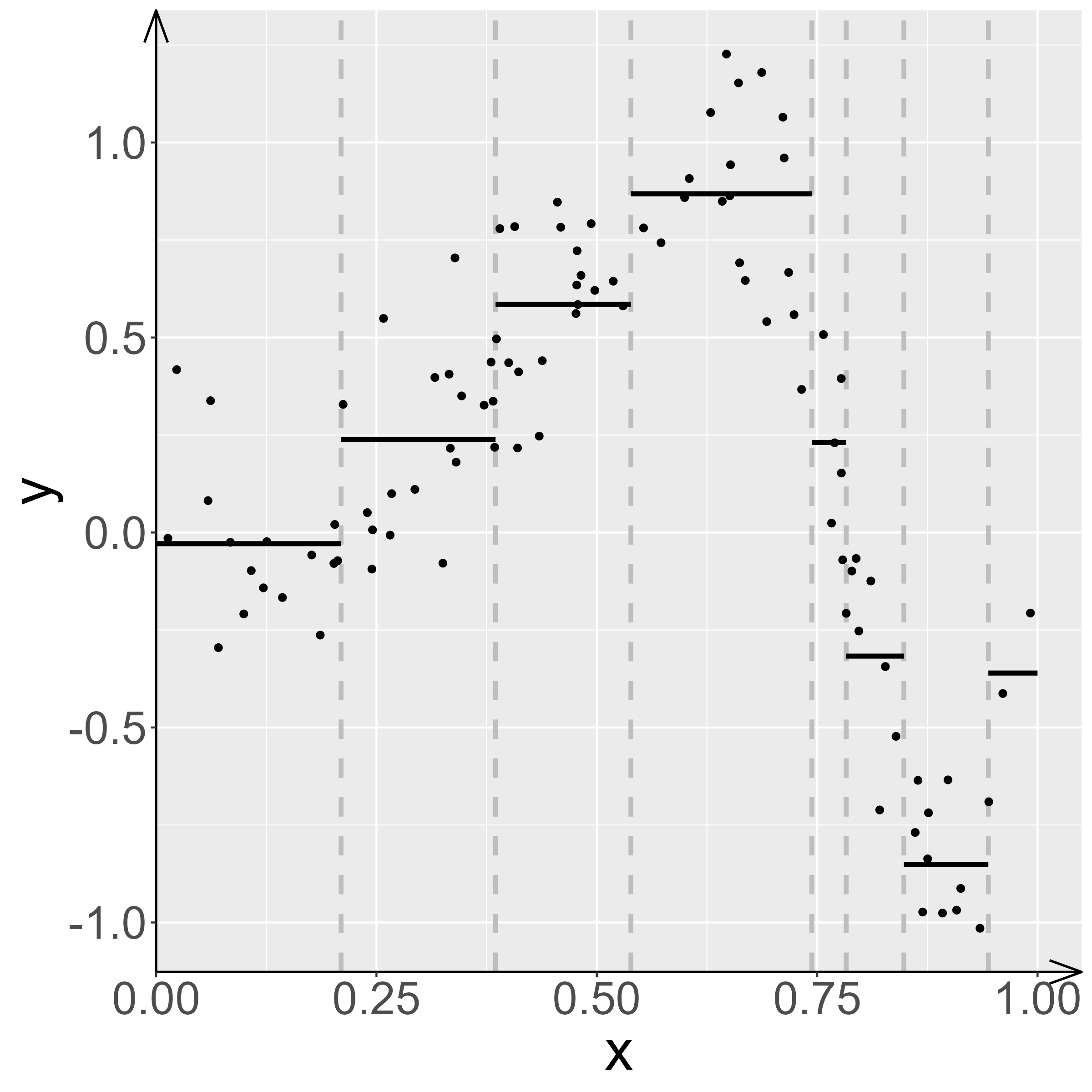}
    \end{minipage}
        \caption{Illustration of the regression tree algorithm for $d=1$. On the left hand side, a single CART split is placed on the $x$-axis (grey dotted line). On the right hand side, the $x$-axis has been split iteratively, resulting in a piecewise constant estimator (black solid line) of the unknown regression function.}
        \label{fig:illustration_cart}
    \end{figure}

Throughout the rise of machine learning over the last decades, decision tree ensembles have captured significant attention. Notably, Breiman's Random Forests \citep{breiman} gained widespread popularity among practitioners and has been applied within various fields, e.g. finance, genetics, medical image analysis, among many others \citep{gu2020_finance, GDA2, Qi_bioinformatics, criminisi_mda, criminisi_unified}. In this paper, we present a simulation study revealing limitations of Random Forests when the target function exhibits certain pure interactions, and we show that adaptions of the algorithm such as Interaction Forests \citep{interactionforests} or 
Random Split Random Forests \citep{tpaper-arxiv}
considerably improve in these scenarios. Moreover, we compare the different variants with Random Forests on real data examples.\\
   Consider a nonparametric regression model
    \begin{align}\label{eq:regmodel}
		Y_i = m(X_i) + \varepsilon_i,
    \end{align}
    $i =1,\dots, n,$ with i.i.d.\ data, (unknown) regression function $m:\R^d \to \R$ which is measurable and $\varepsilon_i$ is zero mean and independent of $X_i$. A regression tree is constructed by partitioning the support of $X_i$ (feature space) via a greedy top-down procedure known as CART \citep{cartbook}. First, the whole feature space (root cell) is split into two daughter cells by placing a rectangular cut such that the data is approximated as well as possible by a function that is constant on each daughter cell. This step is then repeated for each daughter cell and so on, until some stopping criterion is reached. The procedure is called greedy since one optimises the next split given a previous partition instead of optimising the entire partition. We refer to \Cref{fig:illustration_cart} for an illustration. 

    In many situations estimators constructed this way adapt well to high dimensional functions including complex interaction terms. However, difficulties arise in the presence of certain \textit{pure interactions}. We call interactions between multiple covariates pure if there are no marginal effects present containing exactly one of these covariates. Thus, they are hard to detect when employing a step by step procedure using CART, see e.g. \citet{wright2016}. For a formal definition, see \Cref{sec:pure_interactions}.\\
    In this paper, we consider estimation based on regression tree type methods when pure interaction terms are present. We argue that simple regression trees and Random Forests (even with small \texttt{mtry} parameter value; see Section \ref{sec:rf} for a definition of the \texttt{mtry} parameter) are not able to properly approximate pure interactions. In a large simulation study, we show that different modifications of the tree growing procedure lead to algorithms outperforming Random Forests in these cases.\\ 
    More precisely, we focus on the Interaction Forests algorithm \citep{interactionforests}, Random Split Random Forests \citep{tpaper-arxiv} and Extremely Randomized Trees \citep{extra} which have recently been proven to be consistent for regression functions lying in algorithm and data specific function classes \citep{tpaper-arxiv}. Noteworthy, for Random Split Random Forests (RSRF), the function class where the algorithm is consistent includes regression functions with pure interactions.\\
    While the algorithms have in common that they stick to some of the main principles of Random Forests such as aggregation of individual estimators, the tree growing procedures differ: The modifications include additional randomness when choosing splits, allowing partitions into more than two cells in a single iteration step, a combination of both, and usage of other sets than rectangles for the partitions.\\
    The difference between trees in the Interaction Forests algorithm \citep{interactionforests} and usual CART is that the partition into two cells in a single iteration step is allowed to be constructed through certain cuts along \emph{two} directions (cf. \Cref{fig:intf}). The authors have shown in a large real data study that Interaction Forests improve upon Random Forests and related methods, in terms of predictive performance.\\
    The RSRF algorithm is based on the following principle: Split a current cell at random into two daughter cells. Then, split each of the cells using the CART criterion. By repeating this for the current cell, but with a different random split, a number of candidate partitions into four cells is given, among which the one which provides best approximation is chosen. Afterwards, for each of the four new cells, the procedure will be repeated, and so forth.\\   
    For the Extremely Randomized Forest algorithm, one tries several random splits in a single step and chooses the best. In its extreme case, every split is randomly chosen.\\
Our contribution can be summarized as follows. We show via simulations that Random Forest, independent of hyper-parameter choices, cannot adequately deal with pure interaction terms. In addition, we show in our simulations that the variants discussed above, in particular Interaction Forests and RSRF, improve upon Random Forests in these situations. Furthermore, we also observe on real data examples that the interaction-specific methods can outperform Random Forests.\\
We emphasize that our intention is not to promote any of the algorithms for general use in statistical data analysis. Nevertheless, we intend to convey about shortcomings of Random Forests by illustrating that the considered interaction-specific tree algorithms, particularly Interaction Forests, improve upon conventional Random Forests in pure interaction scenarios.\\
In the literature, there exist different algorithms that are both related to Random Forests and designed for models with interactions. Apart from Interaction Forests and RSRF, related algorithms include Bayesian Additive Regression Trees \citep{bart}, Iterative Random Forests \citep{basu2018iterative} and certain non-binary trees used in Random Planted Forests \citep{rpf}. The celebrated Bayesian Additive Regression Trees \citep{bart} algorithm fits a sum of parameterized regression trees by updating trees using a Bayesian backfitting procedure. In a classification setting, Iterative Random Forests \citep{basu2018iterative} identify interactions by reweighting the probability vector for choosing an allowed split coordinate in CART (after each tree was built), using a variable importance measure.\\
    Various variants of Random Forests have been designed for specific purposes, e.g. in survival analysis \citep{surival-forests,dandl2024heterogeneous,moradian2019survival}, quantile estimation \citep{quantile-regression-forests}, ranking problems \citep{ranking-forests}, or estimation of heterogenous treatment effects, see for example \cite{wager2018estimation,dandl2024heterogeneous}. In \cite{biau-guided-tour}, a general review over Random Forests and its variants is provided, including stylized algorithms used in theoretical analyses. For recent theoretical results on consistency for regression trees that use the CART splitting criterion, we refer to \citet{Chi,klu-large,syr,mazumder_wang,tpaper-arxiv,tan2024statistical-arxiv}.
\subsection{Organisation of the paper}
    The paper is structured as follows. In \Cref{sec:pure_interactions}, we introduce the notion of pure interactions and formally introduce the classical CART splitting criterion (for regression) used in CART trees. Then, we discuss why the classical CART criterion is not an appropriate splitting criterion in case of pure interactions. \Cref{sec:algorithms} describes Interaction Forests and \rsrf, while \Cref{sec:all_algorithms} provides an overview over all algorithms considered in our simulations study. The results of our simulation study are presented and discussed in \Cref{sec:simulations}. \Cref{sec:realdata} contains real data examples. A summary and conclusion can be found in \Cref{sec:summary_conclusion}.

\section{Hidden variables unseen by Random Forests}\label{sec:pure_interactions}
We introduce the notion of pure interactions and discuss why the CART algorithm has problems dealing with them. To this end, we need the functional ANOVA decomposition, see \Cref{def:anova} below. Let us denote by $(Y,X)$ a random element distributed as $(Y_i,X_i)$ from \eqref{eq:regmodel}, being independent of $(Y_1,X_1),\dots,(Y_n,X_n)$. Note that $X_k$ may ambiguously refer to the $k$-th component in $X=(X_1,\dots,X_d)$ and the $k$-th observation of the feature vector in the sample, though the meaning becomes clear from the context. For better reading, we use $X_j$ with $j$ in the index for the former ($j$-th component of the vector), and $X_i$ for the latter ($i$-th observation).
\begin{definition}[Functional ANOVA decomposition, see \citet{stone1994use,hooker2007generalized}]\label{def:anova}
We say that the regression function $m$ is decomposed via a functional ANOVA decomposition if
\begin{align*}
m(x)=\sum_{u\subseteq\{1,\dots,d\}} m_u(x_u),
\end{align*} 
with identification constraint that for every $u\subseteq \{1,\dots,d\}$ and $k\in u$,
\begin{align*}
\int m_{u}\left(x_{u}\right) \int p(x) \mathrm dx_{-u} \ \mathrm d x_{k}=0,
\end{align*}
where $p(x)$ is the density of $X = (X_1,\dots, X_d)$, $m_u:\R^{|u|} \to \R$, $x_{u}=(x_j:j\in u)$, $x_{-u}=(x_j:j\notin u)$ and $|u|$ denotes the cardinality of the set $u$.
\end{definition}
We shall discuss the issue by means of the following notion of \emph{simple pure interaction} between two variables. The discussion can be expanded to more general cases.

\begin{figure}
\begin{subfigure}{0.95\textwidth}
        \centering
         \includegraphics[width=0.76\linewidth]{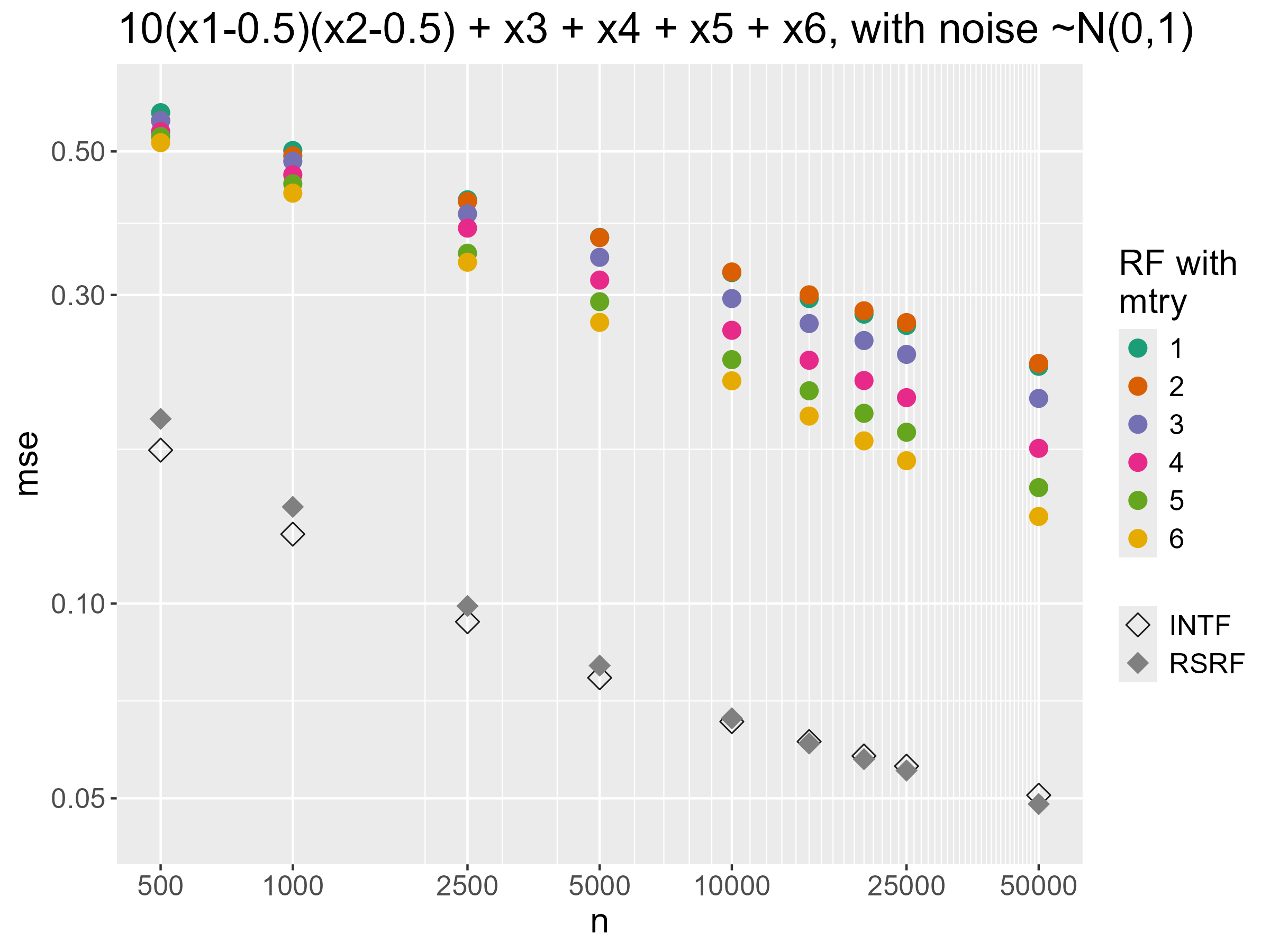} 
        \caption{}\label{fig:plotF_new}
    \end{subfigure}
   \begin{subfigure}{0.95\textwidth}
    \centering
       \includegraphics[width= 0.72\linewidth]{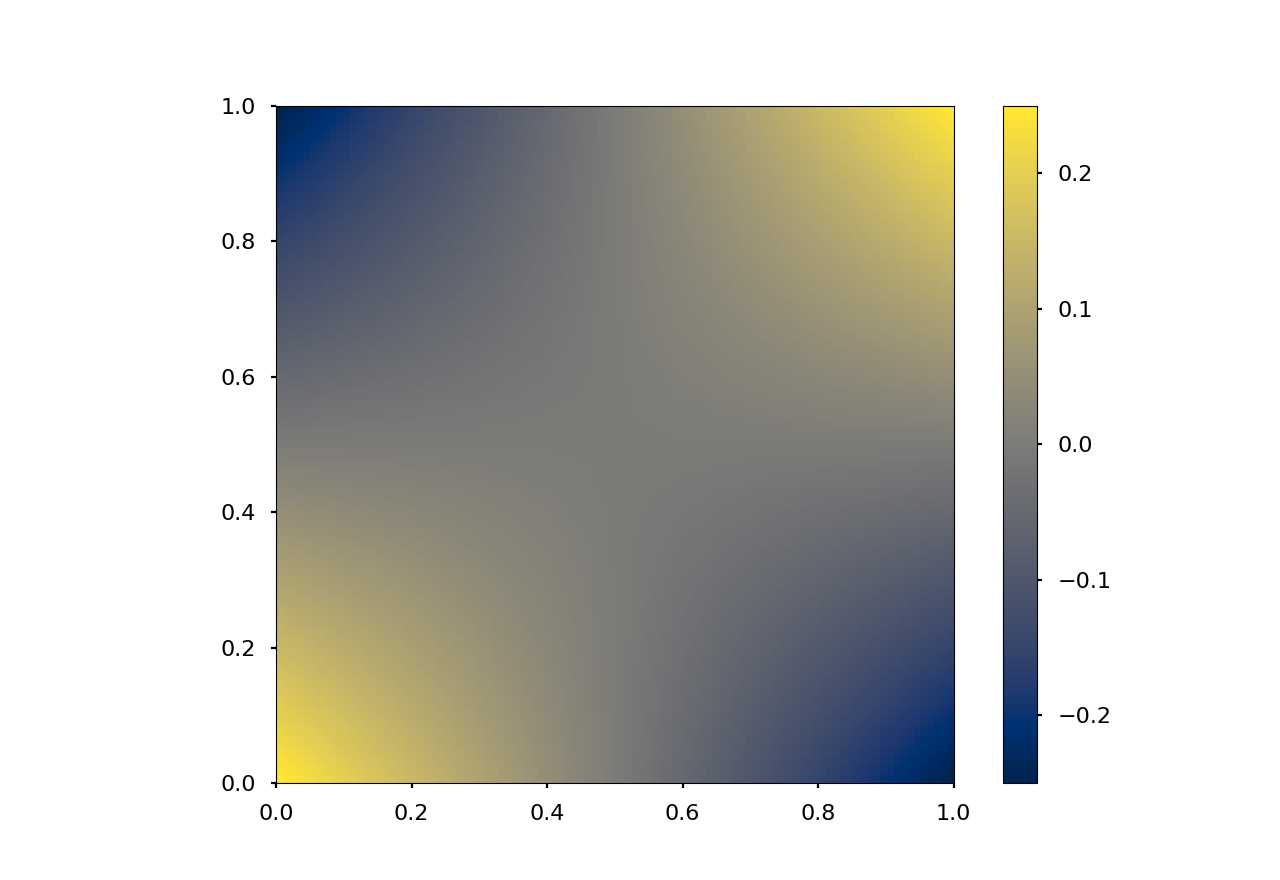}
       \caption{}\label{fig:example_function}
    \end{subfigure}
    \caption{Comparison of Random Forests with Interaction Forests and RSRF at different sample sizes in a pure interaction scenario.\\[0.1cm]
    (\subref{fig:plotF_new}) Estimated mean squared error using Random Forests (for different values of $\texttt{mtry}$), Interaction Forests (INTF) and Random Split Random Forests (RSRF), plotted at $\log$-scales for different sample sizes $n$. The regression model is $Y=10(X_1-0.5)(X_2-0.5) + X_3+X_4+X_5+X_6 + \varepsilon$ with regressors $(X_1,\dots, X_6)$ uniform on $[0,1]^6$ and noise $\varepsilon \sim \mathcal{N}(0,1)$, also see \Cref{fig:example_function}. The number of trees was set to $100$ for each of the methods. For each sample size, $100$ simulations were carried out and the reported mse is the average over mean squared errors calculated on independent test data of size $n$, analogously to the description in \Cref{sec:simulations}.\\[0.1cm]
    (\subref{fig:example_function}) Plot of the function $g: [0,1]^2 \to \R$, $g(x_1,x_2) = (x_1-0.5)(x_2-0.5)$. Taken from \citet{tpaper-arxiv}.}
\end{figure}

\begin{definition}[Simple pure interaction effect]\label{def:simple-pure-interaction}
	Let $m_u$, $u\subseteq \{1,\dots,d\}$ be the components of the functional ANOVA decomposition of $m$ and $j_1,j_2 \in \{1,\dots,d\}$ with $j_1\neq j_2$. The regression function $m$ has a \emph{simple pure interaction effect} in $J = \{j_1,j_2\}$ if 
	\begin{itemize}
 \item	$(X_{j_1},X_{j_2})$ is independent of $(X_j : j  \notin \{j_1,j_2\})$,
 \item $m_{J}\neq 0$,
 \item $m_u = 0$ for any $u$ with $j_1 \in u, j_2\notin u$, or with $j_1 \notin u, j_2\in u$.
	\end{itemize} 
\end{definition}
For a discussion of a notion of higher-order pure-interaction effects, see also \cite{abbe2022merged,tan2024statistical-arxiv}. For the remainder of this section, let us assume for simplicity that $X,X_1,\dots,X_n$ take values in $[0,1]^d$. We need the following property.
\begin{proposition}\label{prop:simple_pure_interaction}
	Assume $m$ has a simple pure interaction effect in $\{1,2\}$. Let $I, I_3,\dots, I_d \subseteq [0,1]$ be measurable subsets and suppose 
	\begin{equation*}
		t = I \times [0,1] \times I_3 \times \dots \times I_d \text{  or  } t= [0,1] \times I \times I_3 \times \dots \times I_d
	\end{equation*}
 with $\PP(X\in t) > 0$. Then, 
 \begin{align}\label{eq:prop}
 \E[m(X) | X\in t] = \E[m(X) | X_3 \in I_3, \dots, X_d \in I_d]. \end{align}
\end{proposition}
For the proof, see Appendix A.\\
The right hand side of \eqref{eq:prop} is the expected mean of a node where no split in $\{1,2\}$ has occurred so far.
The left hand side considers the conditional mean if that node would next be split in coordinate 1 or 2.
We now discuss why algorithms using CART face problems when pure interactions are present. The difficulty lies in the absence of one-dimensional marginal effects guiding to the pure interaction effect, see also \citet{wright2016}. To make this point more concrete for regression trees, let us recall the CART criterion used in regression trees. Suppose $t$ is a rectangular subset of $[0,1]^d$. Write
\begin{align*}
                t_L = t_L(j,s) = \{ x\in t : x_{{j} } \leq {s} \}, \quad t_R = t_R(j,s) = \{x \in t : x_{{j}} > {s}\}.
\end{align*}
for $j\in\{1,\dots,d\}$ and $s\in [0,1]$. One says that $t$ is split at $(j,s)$ into daughter cells $t_L$ and $t_R$. The next definition is the well known CART criterion, given the sample $(Y_i,X_i)$, $i=1,\dots, n$.
\begin{definition}[CART criterion, see \citet{cartbook}]\label{def:sample-cart}
            Let $t\subseteq [0,1]^d$ and let $J \subseteq \{1,\dots,d\}$. The Sample-CART-split of $t$ is defined as splitting $t$ at coordinate $\hat{j} \in J$ and $\hat{s}\in [0,1]$ into daughter cells
            where the split point $(\hat{j},\hat{s})$ is chosen from the CART criterion, that is,
        \begin{align}\label{eq:cart-split-criterion1}
			 (\hat{j},\hat{s}) \in \ \underset{j \in J, s\in t^{(j)} } {\arg\min}\ \mathcal{V}(j,s)
        \end{align}
        with $t^{(j)}=\{x_j: x\in t\}$ and
        \begin{align*}
            \mathcal{V}(j,s) := \sum_{i:X_i \in t_L(j,s)}(Y_i - \hat{\mu}_L)^2 + \sum_{i:X_i \in t_R(j,s)}(Y_i - \hat{\mu}_R)^2,
        \end{align*}
        where  \(\hat{\mu}_k = \{\card{ t_k(j,s)}\}^{-1}\sum_{i : X_i \in  t_k(j,s) } Y_i, \ k=L,R\) and \(\card{t} := \# \big\lbrace i \in \{1,\dots,n\}: X_i \in t \big\rbrace\).
    \end{definition}
    Coming back to the detection of pure interactions, observe that for large samples
 \begin{align}
 \label{eq:CART-approx}
 \begin{split}       \mathcal{V}(j,c) \times( &\#t)^{-1}\\ \approx \qquad &\PP(X \in t_L(j,c)|X\in t)\E[(Y-\E[Y|X\in t_L(j,c)])^2|X\in t_L(j,c)]\\
        + \ &\PP(X \in t_R(j,c)|X\in t)\E[(Y-\E[Y|X\in t_R(j,c)])^2|X\in t_R(j,c)],
        \end{split}
\end{align}
recalling that $(Y,X)$ is distributed as $(Y_1,X_1)$.\\Now assume that the regression function $m$ has a simple pure interaction effect in features $\{1,2\}$. Then, in view of \Cref{prop:simple_pure_interaction}, for any set of the form 
\begin{equation*}
    t=[0,1]^2\times I_3\times \dots \times I_d\subseteq [0,1]^d,
\end{equation*}
and any $j=1,2$ and $s\in[0,1]$,
the right hand side of \eqref{eq:CART-approx} is equal to 
$\E[(Y-\E[Y|X\in t])^2|X\in t]$. This is the maximal possible value attainable. Hence, in the presence of other features $k=3,\dots,d,$ features $j=1,2$ will probably not be chosen to be split leaving the pure interaction effect undetected.
One example is the function $m(x)=A(x_1-0.5)(x_2-0.5) + B\sum_{j=3}^dx_j$ for $A,B\neq 0$, and $X_i$ uniformly distributed on $[0,1]^d$. In this setup, a Sample-CART-split will rarely take on values $\hat{j}=1,2$ if $J\cap \{3,\dots,d\}\neq \emptyset$ and thus the term $A(x_1 - 0.5)(x_2 - 0.5)$ may not be taken into account well.\\
\begin{table}[t]
    \centering
     \begin{tabular}{l C} 
        Algorithm & \text{MSE}  \\
        \hline
 \rowcolor{gray!15}     Interaction Forests (\simif)  &0.154\ (0.029)\\
                        Random Split Random Forests (\simrsrfnf) & 0.190\ (0.030)\\
  \rowcolor{gray!15}    Random Forests (\simrf) & 0.510\ (0.062)\\
                        Extremely Randomized Trees (\simet) &  0.418\ (0.049)   \\
        \hline \hline 
  \end{tabular}   
    
    \caption{Excerpt from our simulation study: Reported mean squared error estimates for different simulations in the regression model $Y=10(X_1-0.5)(X_2-0.5) + X_3+X_4+X_5+X_6 + \varepsilon$ with $\varepsilon\sim \mathcal{N}(0,1)$ and sample size $n=500$. Standard deviations are provided in brackets. Hyper-parameters for each method are optimally tuned. }
    \label{table:results_excerpt}
\end{table}
We illustrate an empirical example in \Cref{fig:plotF_new}. We find that Random Forests at various sample sizes performs poorly compared to the Random Forest-type algorithms Interaction Forests and RSRF that aim for better performance in presence of interactions. The algorithms are introduced in the next \Cref{sec:algorithms}. We emphasize that the classical \texttt{mtry} parameter of Random Forests does not seem to help much with pure interactions. The \texttt{mtry} parameter of Random Forests, for every split, restricts  possible split coordinates to randomly chosen subsets of size \texttt{mtry} of the feature coordinates $\{1,\dots,d\}$. If \texttt{mtry} is small enough (for example $\texttt{mtry}=1$), one can guarantee that splits occur in any coordinate. We note that this may help as can be observed in \Cref{fig:example_3interact} in the appendix. However, as Figure \ref{fig:plotF_new} reveals, this does not solve the problem in general and in the setting considered in \Cref{fig:plotF_new}, $\texttt{mtry}=d$, i.e. no randomization, seems to perform best independent of sample size.\\
A detailed simulation study is presented in \Cref{sec:simulations}, an excerpt of the results is shown in \Cref{table:results_excerpt}.

\begin{remark}
    The independence assumption in \Cref{def:simple-pure-interaction} is an extreme scenario. In settings with correlated variables, the variables $1$ and $2$ may not be completely hidden, but our simulations indicate that the CART algorithm still suffers in such scenarios. We believe that interactions with low corresponding main effects can be problematic, particularly under high noise. Such a viewpoint is different from the one taken in \citet[Section 1.2]{bien2013lasso}. Therein, it is argued that it is natural to assume that, with an interaction effect, also its parental main effect occurs.\\
    In \Cref{sec:realdata} we will see that we also find differences on real data examples which, possibly, may stem from this fact.
\end{remark}
\subsection{Handling Interactions with Random Forest-type algorithms}\label{sec:algorithms}
In this section we introduce two approaches related to Random Forests, which are designed for settings where (pure) interactions are present. As with Random Forests, both methods are based on aggregation of individual (greedily-built) tree-based estimators. First, we describe the \emph{Interaction Forest} algorithm from \citet{interactionforests}. In a large real data study, the authors demonstrated that Interaction Forests improve upon Random Forests in terms of predictive performance.\\
Secondly, Random Split Random Forest (\rsrf) is introduced. \rsrf\ extends the main principles in Random Forests in order to better handle pure interaction scenarios. We emphasize that \rsrf\ appears as an example in \citet{tpaper-arxiv}, where consistency for a general class of regression tree estimators is established. The \rsrf\ algorithm is used to demonstrate that the class of regression functions covered by the theory can differ depending on the specific choice of the partitioning scheme, cf. Section 3 in \citet{tpaper-arxiv}. In particular, the consistency result for \rsrf\ is valid for a strictly larger function class than the corresponding result for Random Forests. Thus, it appears natural to investigate if a difference in performance in the presence of pure interactions can be observed empirically.
\subsubsection{Interaction Forests}\label{subsec:interactionforests}
\begin{figure}
        \centering
        \includegraphics[scale=0.4]{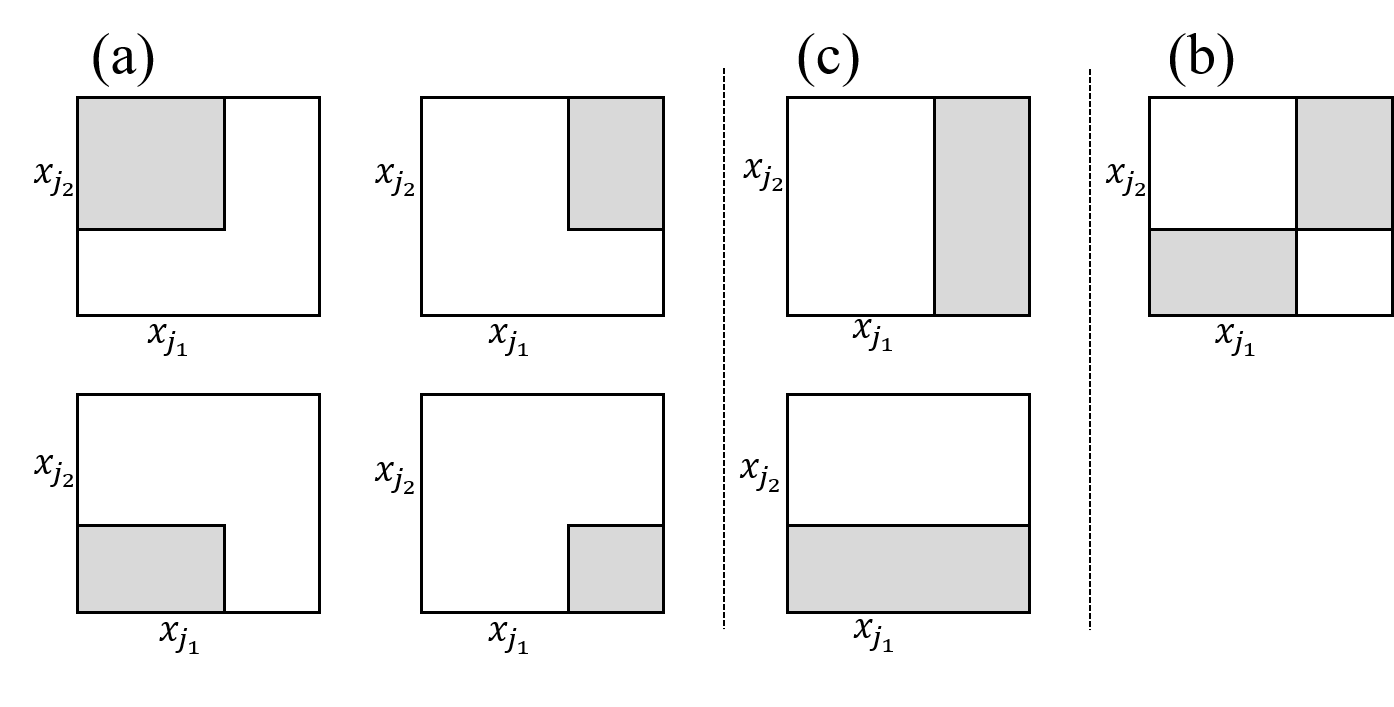}
        \caption{Illustration of possible splits in Interaction Forests. In each of the seven squares, the gray area corresponds to one daughter cell, and the white area to the other. The grouping refers to the three cases $(a)$, $(b)$ and $(c)$ described in \Cref{subsec:interactionforests}. Adapted from \citet[Fig. 2]{interactionforests}.}
        \label{fig:intf}
    \end{figure}

    Let us describe the individual tree estimators. In each iteration step, cells are split into two daughter cells (that are not necessarily rectangles). Let $t \subseteq \R^d$ and two split pairs $(j_1,c_1) \in \{1,\dots, d\} \times t^{(j_1)}$, $(j_2,c_2) \in \{1,\dots, d\} \times t^{(j_2)}$ with $j_1\neq j_2$ be given, where $t^{(j)} = \{x_j : x \in t \}$ is the $j$-th component. Consider the following seven partitions of $t$ into $t_1$ and $t_2 = t\setminus t_1$.
    \begin{align*}
        \begin{minipage}{0.97\linewidth}
            \begin{enumerate}
        \item $t_1 = \{ x \in t : x_{j_1} \blacklozenge_1 c_1 \text{ and }x_{j_2} \blacklozenge_2 c_2 \}$ with $\blacklozenge_1, \blacklozenge_2 \in \{ \leq, \geq \}$,
        \item $t_1 = \{ x \in t : x_{j_1} \leq c_1, x_{j_2} \leq c_2 \} \cup  \{ x \in t : x_{j_1} \geq c_1, x_{j_2} \geq c_2 \}$ and
        \item $t_1 = \{ x \in t : x_{j_l} \leq c_l\}$, where $l=1,2$.
    \end{enumerate}
        \end{minipage}
    \end{align*}
    In \Cref{fig:intf} these seven partitions are illustrated. 

    A current cell $t$ is split by first drawing \texttt{npairs} such variable pairs $(j_1, j_2)$. For each such pair, seven partitions of the forms above are constructed: First, two split points $c_1$ and $c_2$ are randomly drawn and used for the two partitions in case (c). Furthermore, another two split points are chosen at random and these are used to construct the five partitions from (a) and (b). We refer to \citet[Section 4.3]{interactionforests} for the details on how valid split points are chosen. In total, one ends up with $7\times \texttt{npairs}$ partitions of $t$ into two sets among which the one with highest empirical decrease in impurity is chosen. That is, the quantity $\widehat{\zwei}$ is used as score, given by
    \begin{align*}
    \widehat{\zwei}(t; t_1,t_2) =     \frac{ \card{t_{1}}}{\card{t} } \big[  \hat{\mu}(t_{1}) - \hat{\mu}(t) \big]^2 +    \frac{ \card{t_{2}}}{\card{t} } \big[  \hat{\mu}(t_{2}) - \hat{\mu}(t) \big]^2,
\end{align*}
where $\hat{\mu}(t) := \lbrace \# t \rbrace^{-1}\sum_{i:X_i\in t} Y_i$, $\#t = \{i: X_i \in t \}$.\\
    This optimization criterion is equivalent to minimizing variance as in \citet{interactionforests}. For the sake of completeness, this equivalence is shown in the appendix, see \Cref{prop:equivalence_criteria}. We note that Interaction Forests is also available for classification and survival analysis where Gini impurity and the $\log$-rank test statistic, respectively, are used in the corresponding splitting criteria.
\subsubsection{\rsrf: Random Split Random Forests}\label{subsec:rsrf}

The algorithm \rsrf\ is another variant of Random Forests. In contrast to Interaction Forests, the cells remain rectangular. The individual predictors are regression trees built using the \emph{Random-CART procedure}: First, all cells at the current tree depth are split at random, i.e. for each cell, a coordinate is chosen uniformly at random and then, the cell is split at a point chosen uniformly at random along this dimension. Secondly, each of the two resulting cells is split according to the Sample-CART-criterion in \eqref{eq:cart-split-criterion1}. We refer to this combination as a ``Random-CART-step''. Thus, applying such a Random-CART-step, a cell in the tree is split into four cells. In order to enhance the approach, for a given cell $t$, we shall try several Random-CART-steps as candidates for splitting $t$ into four cells $t_{1,1},t_{1,2},t_{2,1},t_{2,2}$, and then choose the one which is ``best'' in terms of empirical (2-step) impurity decrease $\widehat{\zwei}$,
    \begin{align}\label{eq:def_zwei_hut}
        \widehat\zwei(t;t_{1,1},t_{1,2},t_{2,1}, t_{2,2}) = \sum_{ \substack{j=1,2\\k=1,2} } \frac{ \card{t_{j,k}}}{\card{t} } \big[  \hat{\mu}(t_{j,k}) - \hat{\mu}(t) \big]^2.
    \end{align}   
The number $W$ of candidate Random-CART-steps to try is called the ``width parameter''. Furthermore, one can add another candidate split, the ``CART-CART-step'', into this comparison: split the cell $t$ using the Sample-CART criterion (instead of splitting at random) and then split the daughter cells according to the Sample-CART-criterion, again.

\begin{figure}[t]
\centering
\begin{minipage}{0.35\linewidth}
   \includegraphics[width=0.95\linewidth]{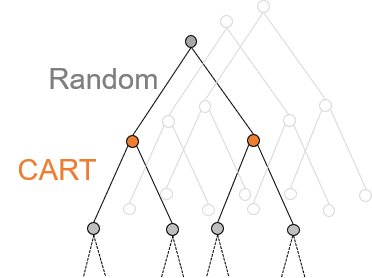}
  \captionof{figure}{Illustration of \rsrf. The background trees (light gray) illustrate other possible candidate partitions. Adapted from \citet{tpaper-arxiv}.}
  \label{fig:RSRF}
\end{minipage}%
\hspace{0.2cm}
\begin{minipage}{.6\linewidth}\centering
        \resizebox{0.9 \linewidth}{0.3 \linewidth}{%
  \begin{tikzpicture}[thick,centered]
            	\node[rectangle,draw, minimum height=3cm, minimum width=4cm] (r) at (0,0) {};
			\coordinate (splitpoint)      at ($(r.north west) + 0.5*(r.south)$);
                \coordinate (splitpointend)   at ($(r.north east) + 0.5*(r.south)$) ;
                \coordinate (splitpoint02)    at ($(r.north) + 0.25*(r.east)$);
                \coordinate (splitpoint02end) at ($(splitpoint02) + 0.5*(r.south)$);
                \coordinate (splitpoint03)    at ($(r.south) + 0.2*(r.west)$);
                 \coordinate (splitpoint03end)    at ($(r.north) + 0.2*(r.west) + 0.5*(r.south)$);

				\draw (splitpoint02) -- (splitpoint02end);
			\draw (splitpoint) -- (splitpointend);
   			\draw (splitpoint03) -- (splitpoint03end);

			\node at ( $(r.north) - (0,0.4)$ ) {};
			\node at ( $(r.center) -(0,0.3)$ ) {};
  
        	\node[rectangle,draw, minimum height=3cm, minimum width=4cm] (t) at (-10,0) {};

                 \node[rectangle,draw, minimum height=3cm, minimum width=4cm] (t1) at (-5,0) {};
			\coordinate (split01)      at ($(t1.north west) + 0.5*(r.south)$);
                \coordinate (split01end)   at ($(t1.north east) + 0.5*(r.south)$) ;
				\draw (split01) -- (split01end);

                \draw[dashed] ([yshift=0.2cm, xshift=+0.3cm]t1.north) to[bend left = 40] node[midway, above]{\huge CART splits} ([yshift=0.2cm]r.north);
                \draw[dashed] ([yshift=0.2cm]t.north) to[bend left = 40]  node[midway, above]{\huge Random split}([yshift=0.2cm,  xshift=-0.3cm]t1.north);
                \node at (t.center) {\huge $t$};
                \node at ($0.5*(splitpoint) + 0.5*(splitpoint02)$) {\huge $t_{11}$};
                \node at ($0.5*(splitpointend) + 0.5*(splitpoint02)$) {\huge $t_{12}$};
                \node at ($0.5*(splitpoint) + 0.5*(splitpoint03)$) {\huge $t_{21}$};
                \node at ($0.5*(splitpointend) + 0.5*(splitpoint03)$) {\huge $t_{22}$};

		\end{tikzpicture}
 }

	\captionof{figure}{Illustration of the procedure used by RSRF for splitting a cell $t$ into $t_{11},t_{12},t_{21}, t_{22}$. Adapted from \citet{tpaper-arxiv}.}
 \label{figure:random-cart}
\end{minipage}
\end{figure}

We refer to Figures \ref{fig:RSRF} and \ref{figure:random-cart} for illustrations of \rsrf. For a detailed description of the algorithm and its implementation, see \Cref{sec:RSRF_detailed}.
In \Cref{sec:extension_depth}, we shortly describe a possible extension of the idea tailored for dealing with higher order interaction terms.

\subsection{Overview: Algorithms considered in our simulation study}\label{sec:all_algorithms}
We compare the following four algorithms in our simulation study.
\begin{align*}
    \simrf\quad & \text{Random Forests \citep{breiman},} \\
     \simet\quad & \text{Extremely Randomized Trees \citep{extra},}\\
     \simif\quad & \text{Interaction Forests \citep{interactionforests}, and} \\
     \simrsrfnf\quad & \text{Random Split Random Forests \citep{tpaper-arxiv}. }
\end{align*}
The individual tree estimators used in the four algorithms have in common that the feature space is partitioned iteratively. Another common feature is that, for each cell $t$ which is about to be partitioned in a single iteration step, the impurity decrease $\widehat{\zwei}$ is used as a score for choosing a partition from a certain set of candidate partitions $P = \{t_1,\dots,t_L\}$ of $t$, where
\begin{align*}
    \widehat{\zwei}(t;P) = \sum_{l=1}^L  \frac{\# t_l}{\#t}[\hat{\mu}(t)-\hat{\mu}(t_l)]^2.
\end{align*}
For example, the CART splitting criterion in \Cref{def:sample-cart} is equivalent to maximizing $\widehat\zwei$ when $L=2$ and $P$ ranges over are all rectangular partitions of $t$, see \Cref{prop:equivalence_criteria}. Thus, the algorithms are of similar structure, however, they differ through the value of $L$ and the specific form of candidate partitions $P$.\\
\subsubsection{Random Forests}\label{sec:rf}
The trees within Random Forests are grown using the CART criterion from \Cref{def:sample-cart} where, in each iteration step, the set $J \subseteq \{1,\dots,d\}$ is chosen uniformly at random and of size $\#J = \texttt{mtry}$. The parameter \texttt{mtry} is the main hyper-parameter in Random Forests.\\
\subsubsection{Extremely Randomized Trees}
Extremely Randomized Trees originate from \citet{extra}, and we use the implementation from \citet{ranger}. Here, in each step \texttt{mtry} coordinates are chosen at random, and for each of these, \texttt{num.random.splits} split points are randomly chosen within this coordinate. Then, the best split is taken using $\widehat{\zwei}$ as criterion. In the extreme case $\texttt{mtry} = \texttt{num.random.splits} = 1$, only a single split is randomly chosen in each iteration step, and no criterion is used.
\subsection{From trees to a forest}
For each of the four algorithms, the final estimator is given by aggregating individual estimators which are of the form
\begin{align*}
    \hat{m}_T(x) = \sum_{t\in T} \1_{(x \in t)} \frac{\sum_{i: X_i \in t} Y_i }{\# \{i :X_i \in t \} },
\end{align*}
where $x$ is an element of the feature space and $T$ denotes the leaf nodes obtained from one of the algorithms. In order to aggregate trees to a forest, $B$ trees are grown each based on a bootstrap sample $(X_i^*,Y_i^*)$, $i=1,\dots,n$ drawn with replacement from the data. Similarly, subsamples of size smaller than $n$ may be used. In any case, this yields $B$ predictors $\hat{m}_{ T^{b} }$, $b=1,\dots,B$, and the final ensemble estimator is obtained by averaging individual tree predictions, i.e.
     \begin{align*}
         \hat{m}_{\text{Forest}}(x) = \frac{1}{B}\sum_{b=1}^B  \hat{m}_{ T^b } (x), \quad x\in [0,1]^d.
     \end{align*}
\begin{remark}
We shortly discuss the relative computation times. This may be particularly important, if one wants to run the different methods to select the best among them, based on a validation set.\\
The computational time of \rsrf\ compared to Random Forests is larger by a factor $W$, respectively $W+1$, depending on inclusion of the CART-CART step. Given that the data is already sorted (for each covariate), the number of calculations of the criterion function required when computing a CART split in Random Forests for a node with $N$ data points is of order $N\times \texttt{mtry}$. The factor $N\times \texttt{mtry}$ reduces to $\texttt{num.random.splits} \times \texttt{mtry}$ for Extremely Randomized Trees. Similarly, this factor is to be replaced by $7\times \texttt{npairs}$, the number of (randomly chosen) partitions considered for a split in Interaction Forests. We note that the extension of \rsrf\ to higher-order interactions described in \Cref{sec:extension_depth} would possibly require much larger $W$ and may only be of practical use for data sets with small number of covariates and sample size. Finally, we did not compare runtimes because \rsrf\ is implemented in R only, and the code is not optimized for computational speed, while the other algorithms use fast implementations available in or based upon the \texttt{ranger} package \citep{ranger}.
\end{remark}

\section{Simulation results}\label{sec:simulations}
We investigate the performance of the algorithms from \Cref{sec:algorithms,sec:all_algorithms} in a simulation study. We consider $\nsimsvar = \nsims$ Monte-Carlo simulations using the underlying regression model
\begin{align*}
 Y^{s}_i = m(X_i^{s}) + \varepsilon_{i}^s, \quad i=1,\dots, n=500 \text{ and } s=1,\dots, \nsimsvar = \nsims.
\end{align*}
In total, we investigate five different models \modelPureType, \modelHierarchical, \modelAdditive, \modelPureZwei, \modelPureDrei\ which are summarized in \Cref{table:sim_reg_models}. The model \modelPureType\ is not pure in the sense of \Cref{def:simple-pure-interaction}, but ignoring correlation, its structure is only slightly different and a similar property as the one in \Cref{prop:simple_pure_interaction} can be derived. The models thus include three settings with pure or pure-type interactions. In the first model the sine function is used to include a nonlinear relationship. \\
We emphasize that, as usual, it is unclear to which extent our simulation studies and our real data examples are representative. Nevertheless, the chosen pure interaction models considered in our simulations may provide a first impression of possible gains of alternative Random Forest type algorithms.\\
For $\modelPureDrei$, the number of covariates was set to $d=6$. For all other models we chose $\dsimvar = 4, 10, 30$. The following distributional assumptions were made. For models $\modelPureZwei$ and $\modelPureDrei$, we assume that $X_i$ is uniformly distributed on $[0,1]^d$. For models \modelPureType,\ \modelHierarchical\ and \modelAdditive, we follow \citet{nisp}, see also \citet{rpf} and set
\begin{align*}
    X_{i,k}^s= 2.5\pi^{-1}\arctan( \widetilde{X}_{i,k}^s ), \quad k=1,\dots, \dsimvar,
\end{align*}
where $\tilde{X}_i^{s} = (\widetilde{X}_{i,1}^s \dots, \widetilde{X}_{i,d}^s)$ follows a $d$-dimensional normal distribution with mean zero and $\cov(\tilde{X}_{i,k}^s) =\corr( \tilde{X}_{i,k}^s ) = 0.3$. We note that $X_i^s=(X^s_{i,1},\dots,X^s_{i,\dsimvar})$ has correlated marginals and bounded support given by $(-1.25,1.25)^{\dsimvar}$.
The $\varepsilon_i^s$'s are i.i.d. standard normal.\\
\begin{table}[b]
    \centering
    \begin{tabular}{c l}
        Abbreviation & Regression function \\
        \hline
        \modelPureType & $m(x)= -2\sin(x_1x_2\pi)+2\sin(x_2x_3\pi)$\\
        \modelHierarchical & $m(x)= -2\sin(x_1 \pi) + 2\sin(x_2 \pi) - 2\sin(x_3 \pi)$ \\
        & \qquad\qquad\qquad $-2\sin(x_1 x_2 \pi ) + 2 \sin( x_2 x_3\pi )$  \\
        \modelAdditive & $m(x)=  -2\sin(x_1\pi) + 2\sin(x_2\pi) - 2\sin(x_3\pi)$\\
        \modelPureZwei & $m(x)=5(x_1 - 0.5)(x_2 - 0.5) + 5x_3$ \\
        \modelPureDrei & $m(x)=10(x_1-0.5)(x_2-0.5)+x_3+x_4+x_5+x_6$
  
    \end{tabular}
    \caption{Overview of models}
    \label{table:sim_reg_models}
    \end{table}
    Denoting by $\widehat m^s$ an estimator of $m$ given data $(X_i^s,Y_i^s)_{i=1,\dots, n}$ we measure its accuracy by the mean squared error on an independently generated test set $X^{\text{(test)},s}_i$ ($i=1,\dots,\simsamplesize$), i.e.
    \begin{align}\label{eq:mse_est}
        \frac{1}{\nsims}\sum_{s=1}^{\nsims} \left( \frac{1}{\simsamplesize}\sum_{i=1}^{ \simsamplesize} \left( \widehat{m}^s\big( X^{\text{(test)},s}_i \big)- m\big( X^{\text{(test)},s}_i \big) \right)^2 \right).
    \end{align}

\begin{table}[h!]
    \centering
    \begin{tabular}{c l C C C} 
        Model &Algorithm & d=4 & d=10 & d=30 \\
        \hline
 \rowcolor{gray!15}\cellcolor{white}      
                        &\simrsrfnf\ (CV)  & 0.208\ (0.030) & 0.263\ (0.042) &  0.389 \ (0.077) \\
  \modelPureType     &\simif\ (CV)   &0.170\ (0.026)& 0.222\ (0.030) &  0.325\ (0.056)\\    
  
  \rowcolor{gray!15}\cellcolor{white}                      &\simrf\ (CV)  & 0.311 \ (0.068) & 0.697 \ (0.191) & 1.338 \ (0.278)\\                        
                        &\simet\ (CV)  & 0.205\ (0.037) & 0.395 \ (0.158) & 0.862\ (0.382) \\                         
  \rowcolor{gray!15}\cellcolor{white}                 & \simmeanY         & 2.137\ (0.150) &  2.170\ (0.159) &  2.163\ (0.141) \\
                 & \simeinsnn        & 1.281\ (0.115) &  2.480\ (0.197) &  3.923\ (0.328) \\
        \hline \hline 
     \rowcolor{gray!15}\cellcolor{white}                     &\simrsrfnf\ (CV)  & 0.425 \ (0.048) & 0.552 \ (0.067) & 0.682 \ (0.072) \\
      \modelHierarchical             &\simif\ (CV) & 0.394\ (0.047)& 0.515\ (0.058) & 0.623\ (0.059) \\
                        \rowcolor{gray!15}\cellcolor{white}   

 &\simrf\ (CV)& 0.418 \ (0.050) & 0.555 \ (0.067)  & 0.677 \ (0.071)\\                     
                        &\simet\ (CV)  & 0.361\ (0.044) & 0.452\ (0.054) &0.538\ (0.059) \\
 \rowcolor{gray!15}\cellcolor{white}                    
              & \simmeanY          & 8.070 \ (0.395) &  8.116 \ (0.446) &  8.062 \ (0.455) \\ 
            &  \simeinsnn        & 2.056 \ (0.175) &  5.794 \ (0.401) & 10.505 \ (0.803) \\                     
        \hline \hline
          \rowcolor{gray!15}\cellcolor{white}                       
                        &\simrsrfnf\ (CV)   & 0.371 \ (0.041) &0.472 \ (0.056) & 0.571 \ (0.050) \\                    
                                             \modelAdditive             &\simif\ (CV) & 0.343\ (0.041) & 0.431\ (0.047) & 0.512\ (0.050) \\             
               \rowcolor{gray!15}\cellcolor{white}            &\simrf\ (CV)  & 0.350 \ (0.040) & 0.460 \ (0.051) & 0.554\ (0.054)\\
                                                 
            &\simet\ (CV)  & 0.299\ (0.035)& 0.372\ (0.046) & 0.430\ (0.048)\\
             \rowcolor{gray!15}\cellcolor{white}    & \simmeanY        & 5.992 \ (0.350) &  5.933 \ (0.310) &  5.958 \ (0.338) \\ 
            & \simeinsnn       & 1.778 \ (0.138) &  4.252 \ (0.289) &  7.713 \ (0.580) \\ 

        \hline \hline
                     \rowcolor{gray!15}\cellcolor{white}      &\simrsrfnf\ (CV)  & 0.155\ (0.027)  & 0.197\ (0.021) & 0.226\ (0.024)\\                     
            \modelPureZwei               &\simif\ (CV) & 0.127\ (0.024) & 0.172\ (0.025)& 0.214 \ (0.023) \\
         \rowcolor{gray!15}\cellcolor{white} 
                        &\simrf\ (CV)  & 0.188\ (0.029) & 0.235\ (0.023)& 0.251\ (0.023) \\
        &\simet\ (CV)  & 0.128\ (0.024)& 0.190\ (0.020) & 0.209\ (0.021)\\
 \rowcolor{gray!15}\cellcolor{white} 
   & \simmeanY  & 2.264 \ (0.105) &  2.269 \ (0.111) &  2.290 \ (0.086) \\
   & \simeinsnn      & 1.156 \ (0.083) &  1.872 \ (0.136) &  3.132 \ (0.231) \\ 
   \hline \hline
    \end{tabular}
    \newline
\vspace*{1cm}
\newline
 \begin{tabular}{c l C}
        &Algorithm & d=6  \\
        \hline
 \rowcolor{gray!15}\cellcolor{white}      
                        &\simrsrfnf\ (CV)  &0.190\ (0.030)\\
            \modelPureDrei             &\simif\ (CV)  & 0.154\ (0.029)\\
      \rowcolor{gray!15}\cellcolor{white}                   &\simrf\ (CV)  & 0.510\ (0.062)\\
                           
                        &\simet\ (CV)  &  0.418\ (0.049) \\
    \rowcolor{gray!15}\cellcolor{white}                      &  \simmeanY & 1.027 \ (0.068) \\                        
                    & \simeinsnn & 1.289 \ (0.103) \\ 
        \hline \hline 
  \end{tabular}

    \caption{Reported mean squared error estimates for different simulations with parameter choice using (CV). $\simmeanY$ uses the mean of the responses as estimator and $\simeinsnn$ is the $1-$nearest neighbor estimator. Standard deviations are provided in brackets.}
    \label{table:results_simulation_cv}
\end{table}

\begin{remark}
    For the sake of clarity, we shortly elaborate on why we report on \eqref{eq:mse_est}, and not on the same expression but with $m(X_i^{\operatorname{(test)},s } )$ replaced by the corresponding response $Y_i^{\operatorname{(test)},s } $, see \eqref{eq:estimator_using_response} below. Suppose that $(Y,X)$ is distributed as $(Y_i,X_i)$ from model \eqref{eq:regmodel} and assume that $(Y,X)$ is independent of the $(Y_i,X_i)'s$. Then, 
    \begin{align}\label{eq:mses_sum}
       \operatorname{MSE}_Y:= \E[ (Y - \widehat{m}(X))^2 ]= \underbrace{\E[ (m(X) - \widehat{m}(X))^2] }_{=:\operatorname{MSE}} + \sigma^2,
    \end{align}
    with $\sigma^2 = \E[\varepsilon_i^2]$. In the simulations, we have $\sigma^2=1$.
    $\operatorname{MSE}_Y$ is the expected squared difference of the prediction from the fitted model and the outcome variable. $\operatorname{MSE}$ is the expected squared difference of the prediction from the fitted model and the true model. Thus, these quantities differ by the known constant $\sigma^2$. One could also estimate $\operatorname{MSE}_Y$ using		\begin{align}\label{eq:estimator_using_response}
			\frac{1}{100}\sum_{s=1}^{100} \left( \frac{1}{500}\sum_{i=1}^{500} \left(Y_i^{\operatorname{(test)},s } - \hat{m}^s(X_{i}^{\operatorname{(test)},s } ) \right)^2 \right).
		\end{align}
        However, instead of \eqref{eq:estimator_using_response}, a better estimate for $\operatorname{MSE}_Y$ is given by
		\begin{align}\label{eq2:estimator_using_truth}
			\frac{1}{100}\sum_{s=1}^{100} \left( \frac{1}{500}\sum_{i=1}^{500} \left(m(X_i^{\operatorname{(test)},s } ) - \hat{m}^s(X_{i}^{\operatorname{(test)},s } ) \right)^2 \right) + \sigma^2,
		\end{align}
        because knowledge of $m$ and $\sigma^2$ is available in the simulation. In the simulation study, we report on the first summand of \eqref{eq2:estimator_using_truth} (i.e. an estimator for $\operatorname{MSE}$).
\end{remark}

We used the $R$-package \texttt{ranger} \citep{ranger} for Random Forests and Extremely Randomized Trees. 
For the latter, the option \texttt{splitrule} in \texttt{ranger} is set to ``extratrees''. Interaction Forests are implemented in the $R$-package \texttt{diversityForest} \citep{diversity}. The RSRF algorithm is implemented using $R$ and code is provided in the supplementary material. See also \Cref{sec:RSRF_detailed} for details on RSRF.\\
In order to determine a suitable choice for the hyper-parameters from a set of parameter combinations, we use $10$-fold cross-validation (CV). Additionally, we determine ``optimal'' parameters (opt) chosen in another simulation beforehand. In both cases, $200$ sets of parameter combinations are chosen at random and we refer to \Cref{tab:parameters} for the parameter ranges. 
\begin{table}[t]
    \centering
    \begin{tabular}{c l l}
        Algorithm & Parameter & Value / Range \\
        \hline  
        \simrsrfnf & \simcartcart   & True, False \\
                   & \simreplace   & True, False \\
                   & \simwidth        & $1,2,\dots, 15 \ (d=4,6,10)$ resp.  $1,2,\dots,30 \ (d=30)$ \\
                   & \simmtrycartcart & $1,2,\dots, d$ \\
                   & \simmtryrc       & $1,2,\dots, d$ \\
                   & \simminnodesize  & $5,6,\dots, 30$ \\
                   & \simntrees       & $100$ \\
                   & \mtrymode        & \mtrymodenf \\ 
            \hline 
            \simrf & \texttt{num.trees} & $500$ \\
                   & \texttt{min.node.size} & $5,6,\dots 30$ \\
                   & \texttt{replace} & True, False \\
                   & \texttt{mtry}    & $1,2,\dots ,d$ \\
                   \hline
           \simif &  \texttt{num.trees} & $500$ \\     
                  & \texttt{min.node.size} & $5,6,\dots 30$ \\
                   & \texttt{replace} & True, False \\
                   & \texttt{npairs} & $1,2,\dots,100 \ (d=4)$, resp. $1,2,\dots, 150 \ (d=6)$,\\& & \quad resp. $1,2,\dots, 250 \ (d=10)$,  \\
                   & & \quad resp. $1,2,\dots, 750 \ (d=30)$ \\
                    \hline
            \simet &  \texttt{num.trees} & $500$ \\     
                  & \texttt{min.node.size} & $5,6,\dots 30$ \\
                   & \texttt{replace} & True, False \\
                    & \texttt{num.random.splits}    & $1,2,\dots ,10$ \\
                    & \texttt{mtry}    & $1,2,\dots ,d$ \\
                   & \texttt{sample.fraction}  & $1$ \\
                   & \texttt{splitrule} & \texttt{extratrees}
            
    \end{tabular}
    \caption{Parameter settings used in the simulation study for the different algorithms.}
    \label{tab:parameters}
\end{table}
To determine ``optimal'' parameters we ran $30$ independent simulations on new data $\bar{X}^s_{i}$ and test points $\bar{X}^{\text{(test)},s}_{i}$ ($i=1,\dots, 500$; $s=1,\dots,30$) and chose the parameter settings for which lowest mean squared error was reported, averaged over $30$ simulations, that is
\begin{align*}
      \frac{1}{30}\sum_{s=1}^{30} \left( \frac{1}{\simsamplesize}\sum_{i=1}^{ \simsamplesize} \left( \widehat{m}^s\big( \bar{X}^{\text{(test)},s}_i \big)- m\big( \bar{X}^{\text{(test)},s}_i \big) \right)^2 \right).
\end{align*}
The parameter settings obtained from this search can be found in \Cref{subsec:parameter_settings}. Apart from the four algorithms under investigation, we also include two naive base learner for comparison. First, by \simmeanY\ we denote taking the mean of the response as an estimator. This is the best estimator in terms of squared error if the regression model did not contain any covariates. Second, we add the $1$-nearest neighbor estimator (\simeinsnn) in order to include an estimator that overfits to the data.\\
The results from our simulation can be found in \Cref{table:results_simulation_cv} (CV) where parameters were chosen through cross-validation. Furthermore, \Cref{table:results_simulation_opt} (opt) in the appendix shows the results using optimally chosen parameters. In general, the results from (CV) and (opt) differ in most cases by less than $5\%$. This indicates that the performance of the procedures with parameters chosen via cross-validation is not strongly affected by the dimensions of the tuning parameters. In \Cref{tab:results_ranked} the algorithms are ranked from lowest to largest MSE (for the version with optimal parameters).
The details on the RSRF algorithm are described in \Cref{sec:RSRF_detailed} and we note that in the appendix, we distinguish two slightly different setups of \rsrf\ one of which is the one reported on here, and the interested readers can find additional simulation results for the other setup in \Cref{sec:simulations_rsrf_af}.
\subsection{Discussion of the simulation results}
From \Cref{table:results_simulation_cv} we observe that in the models \modelPureType\ and \modelPureDrei\ where strong pure interactions are present, \rsrf\ and Interaction Forests clearly outperformed Random Forests. \\ 
Comparing $\modelPureZwei$ and $\modelPureDrei$, we see that the gap between the algorithms \simif /\simrsrfnf\ and $\simrf$ is much larger in $\modelPureDrei$ where more additive components are present and contribution by the two interacting variables is stronger. However, we see that in \modelPureZwei, \simrsrfnf\ and \simif\ performed slightly better than \simrf. The model \modelAdditive\ is treated equally well by $\simrf$ and $\simrsrfnf$. The same holds true for the hierarchical interaction model.
In every simulation, $\simet$ was better than \simrf. Similarly, $\simif$ was generally better than \simrsrfnf. When the parameter \texttt{npairs} is not extremely large, the number of partitions considered in any step for $\simif$ is smaller than the number of partitions for \simrsrfnf. This suggests that the Interaction Forests algorithm and the Extremely Randomized Trees algorithm benefit from additional randomization in similar ways.\\ \ \\
We note that, in \modelPureZwei, Extremely Randomized Trees was slightly better than $\simrsrfnf$. An inspection of \Cref{tab:et_opt} in the appendix reveals that, for each of the \texttt{mtry} coordinates, only a single random splitpoint is drawn. In \modelPureDrei, however, we find that Extremely Randomized Trees performed much worse than \simrsrfnf\ or \simif, though still being better than \simrf.\\
To sum this up, solely imposing additional randomness to the Sample-CART criterion \eqref{eq:cart-split-criterion1} as is done via \texttt{mtry} in $\simrf$, and even more strongly in $\simet$, appears not to be always sufficient to obtain good predictive performance in pure interaction models. Indeed, the algorithms \simif/\simrsrfnf\ which use both random splits and different cell partitioning schemes in any step, performed best in the presence of pure interactions.\\ \ \\
The base learners \simmeanY\ and \simeinsnn\ were clearly outperformed by the other algorithms, in any setting, as can be seen in \Cref{table:results_simulation_cv}. In addition, we see that \simeinsnn\ suffers from the curse of dimensionality. We note that, in model \modelPureType, the mean squared error of \simrf\ relative to that of \simeinsnn\ (i.e. the ratio of the two errors) even increases with growing dimension $d$ (see \Cref{table:results_cv_normed_einsnn} in the appendix). Thus, in this example the Random Forest responds poorly to increasing sparsity in the model. In \Cref{table:results_cv_normed_meanY} in the appendix, the interested reader can find mean squared errors, relative to the mean squared error for \simmeanY\ (i.e. the best estimator when the model did not contain any covariates). For example, for all the models with $d=10$, we see that these values are approximately in the range from $6\%$ to $12\%$ for \simrsrfnf\ and \simif. This is valid for \simrf, too, except for the model \modelPureType\ where the ratio is considerably larger ($32\%$). In general, smallest values were obtained for the hierarchical model.\\
In \Cref{subsec:parameter_settings}, tables containing the parameters used for (opt) can be found. Furthermore, for the new \rsrf\ algorithm, we additionally include the top $30$ parameter settings from the parameter search (opt) in the supplementary material and some remarks are included in the appendix, see \Cref{subsec:remark_parameters}. However, we point out that our aim was not to provide an in-depth analysis on the hyper-parameter choices which would be beyond the scope of the paper. 
\section{Illustrations on real data}\label{sec:realdata}
\begin{figure}[t]
     \centering
     \begin{subfigure}[b]{0.32\textwidth}
            \centering
            \caption{\concrete}
         \includegraphics[width=\textwidth]{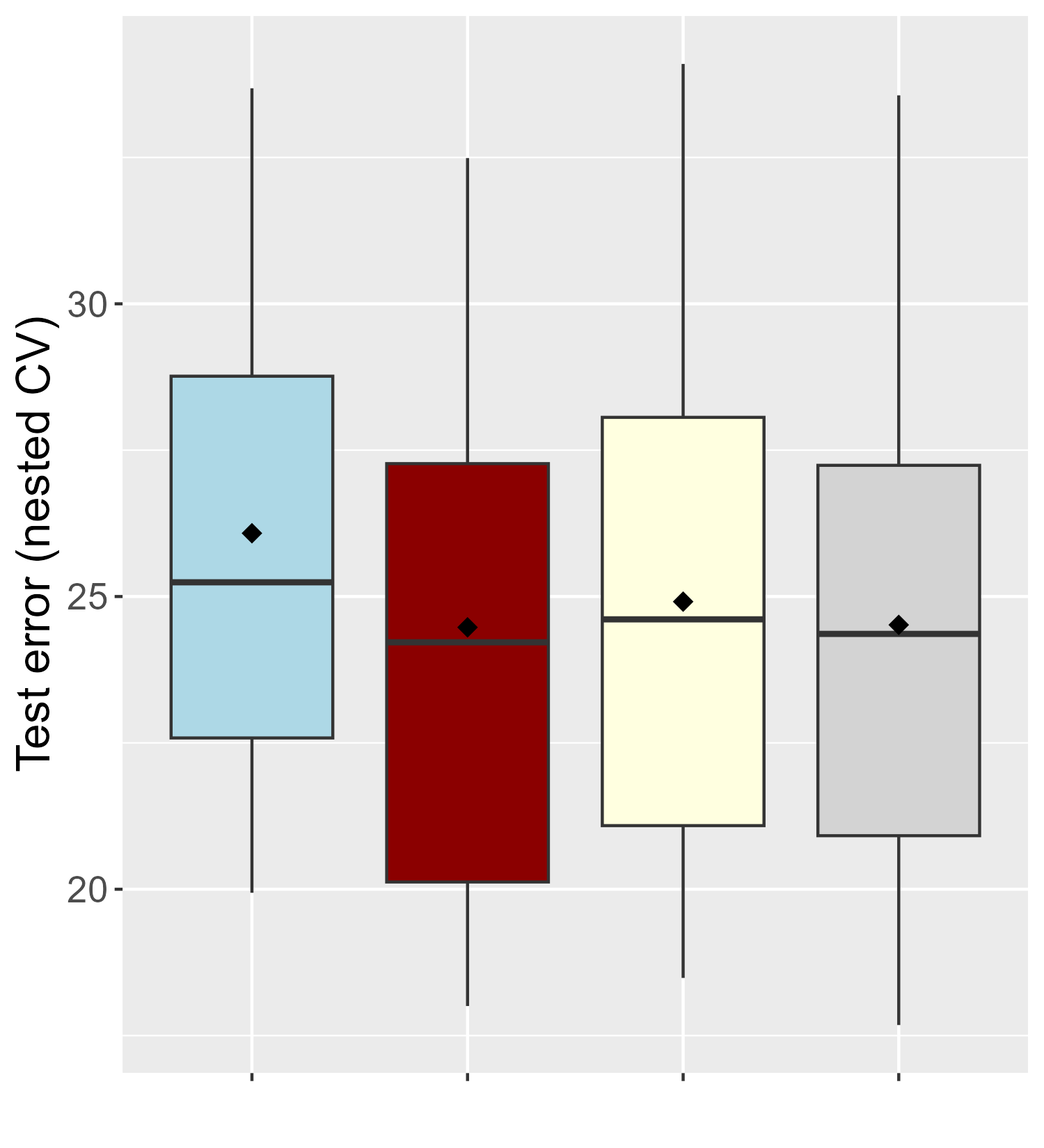}
     \end{subfigure}
     \hfill
     \begin{subfigure}[b]{0.32\textwidth}
         \centering
         \caption{\airfoil}
         \includegraphics[width=\textwidth]{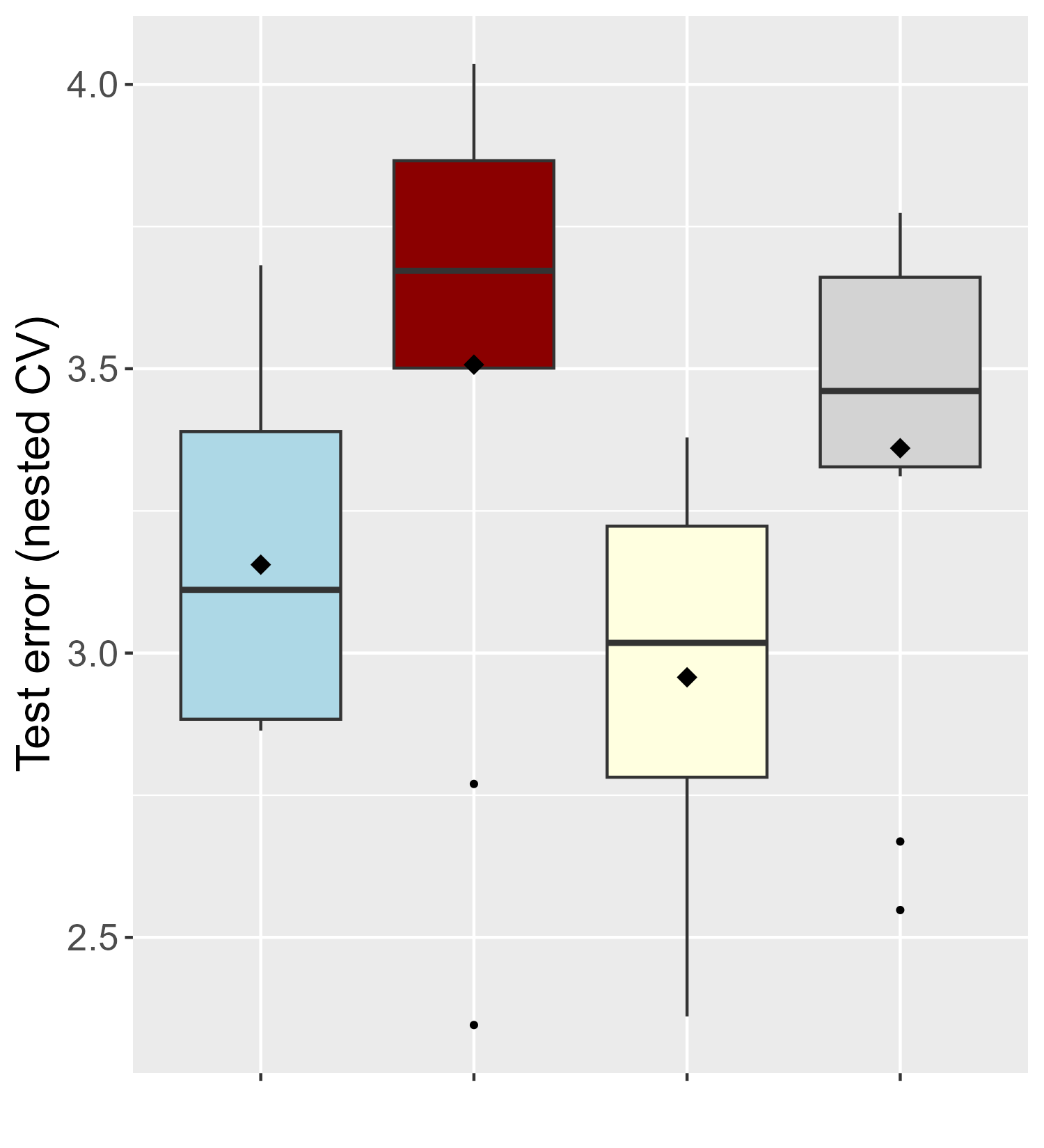}
     \end{subfigure}
     \begin{subfigure}[b]{0.32\textwidth}
         \centering
          \caption{\abalone}
         \includegraphics[width=\textwidth]{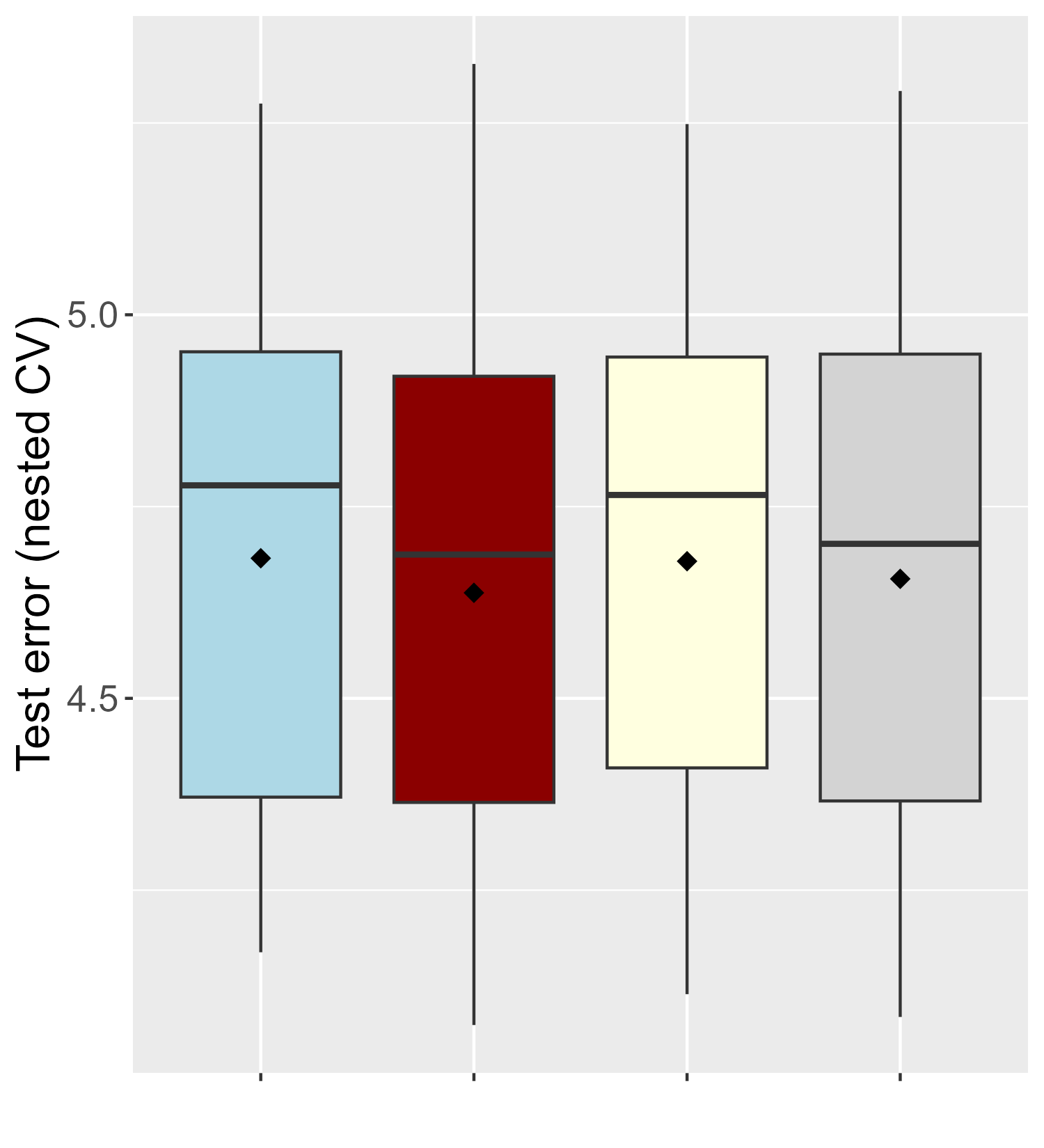}
     \end{subfigure}
       \hfill
    \begin{subfigure}[b]{0.32\textwidth}
         \centering
           \caption{\robot}
         \includegraphics[width=\textwidth]{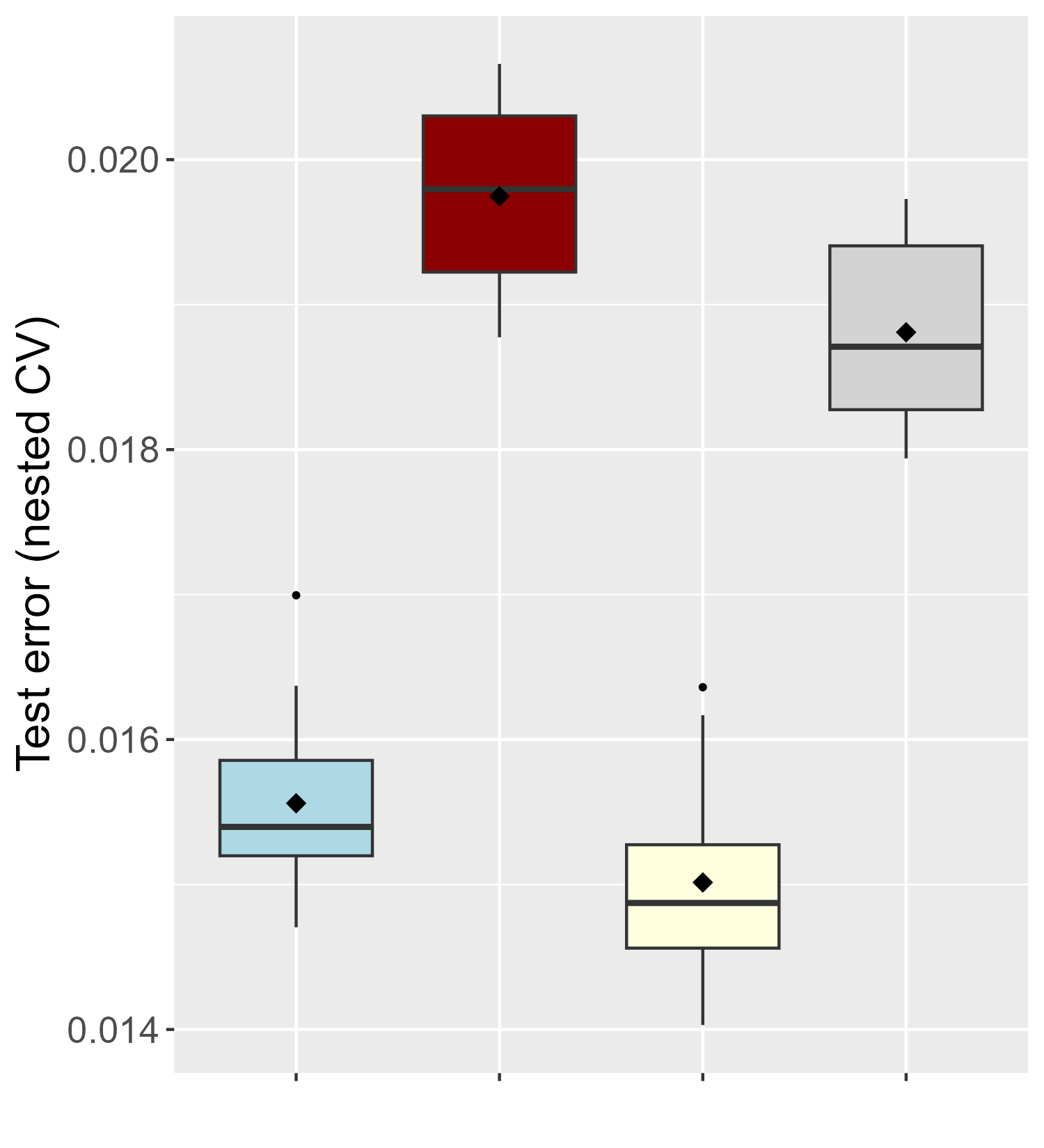}
     \end{subfigure} 
     \begin{subfigure}[b]{0.32\textwidth}
         \centering
           \caption{\calhousing}
         \includegraphics[width=\textwidth]{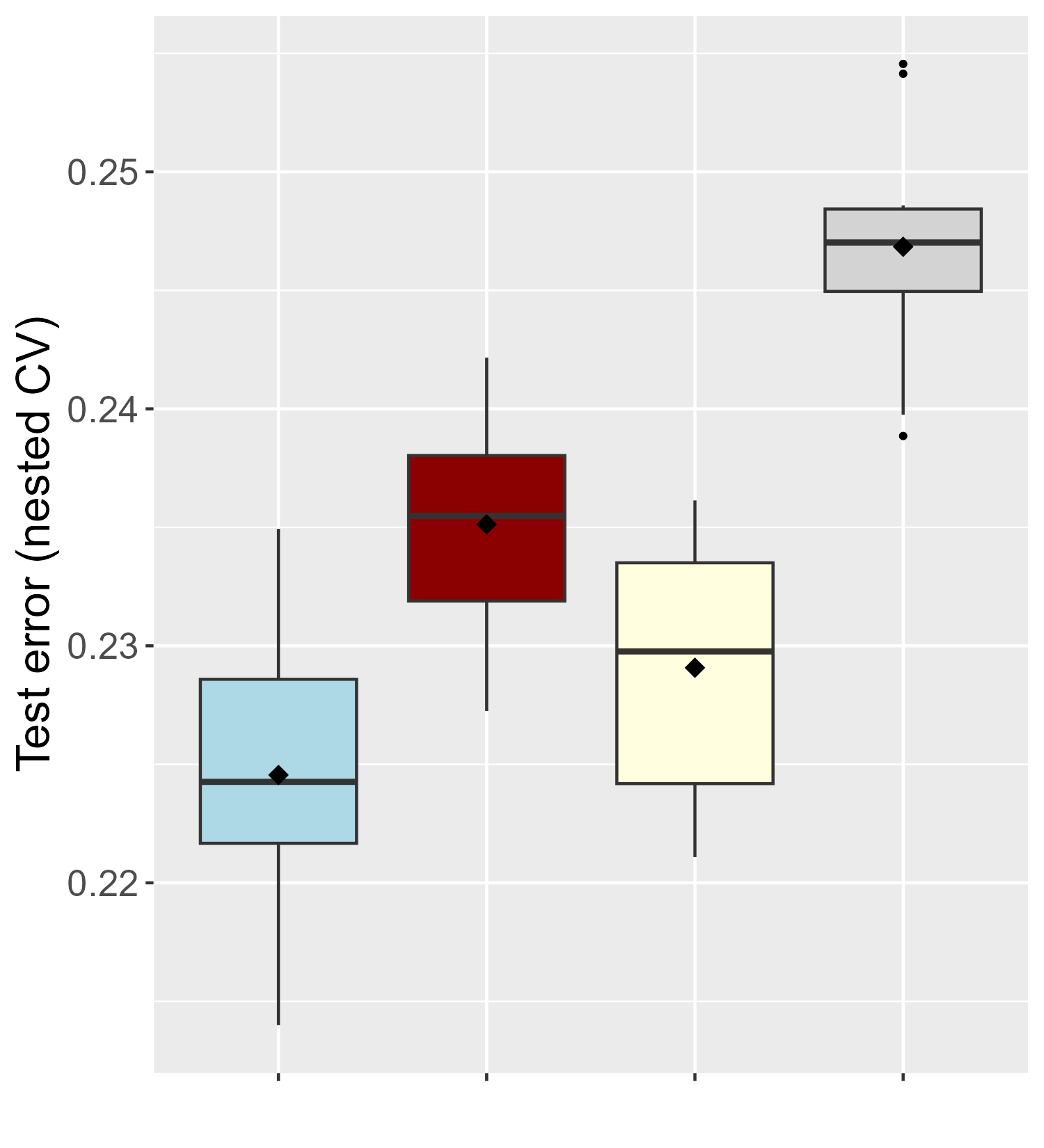}
     \end{subfigure} \hfill
       \begin{subfigure}[b]{0.32\textwidth}
         \centering
           \caption{Plot Legend}
         \includegraphics[width=\textwidth]{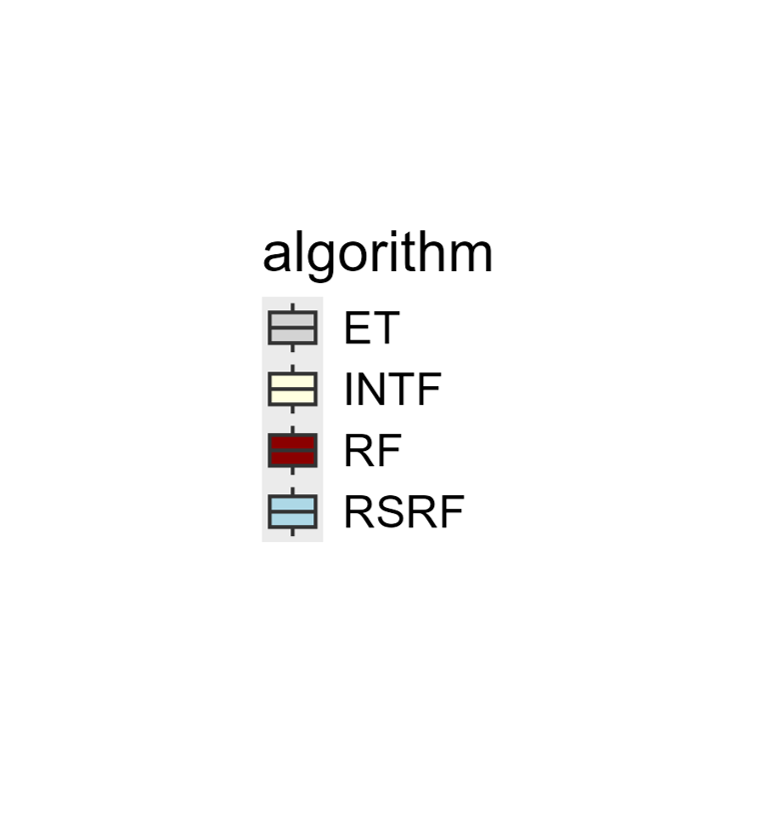}
     \end{subfigure}
        \caption{Boxplots showing squared errors from nested CV for the real datasets. The black diamonds represent the mean of the squared errors. The parameters considered are shown in \Cref{tab:parameters_realdata} in the appendix.}
        \label{fig:boxplots}
\end{figure}

\begin{figure}[t]
     \centering
     \begin{subfigure}[b]{0.32\textwidth}
            \centering
            \caption{\concrete\ (HD)}
         \label{fig:concrete_hd}
         \includegraphics[width=\textwidth]{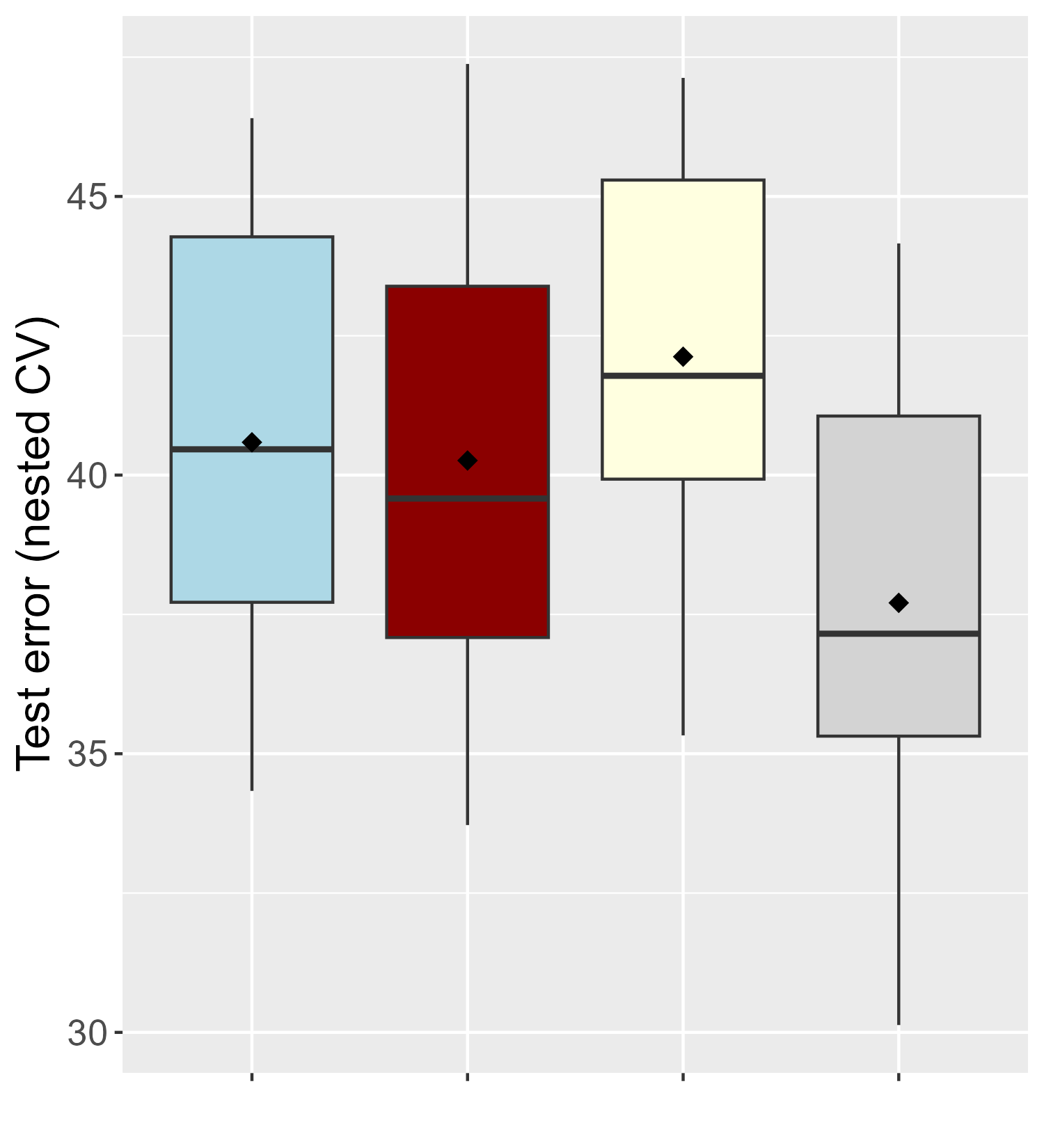}
     \end{subfigure}
     \hfill
     \begin{subfigure}[b]{0.32\textwidth}
         \centering
         \caption{\airfoil\ (HD)}
         \label{fig:airfoil_hd}
         \includegraphics[width=\textwidth]{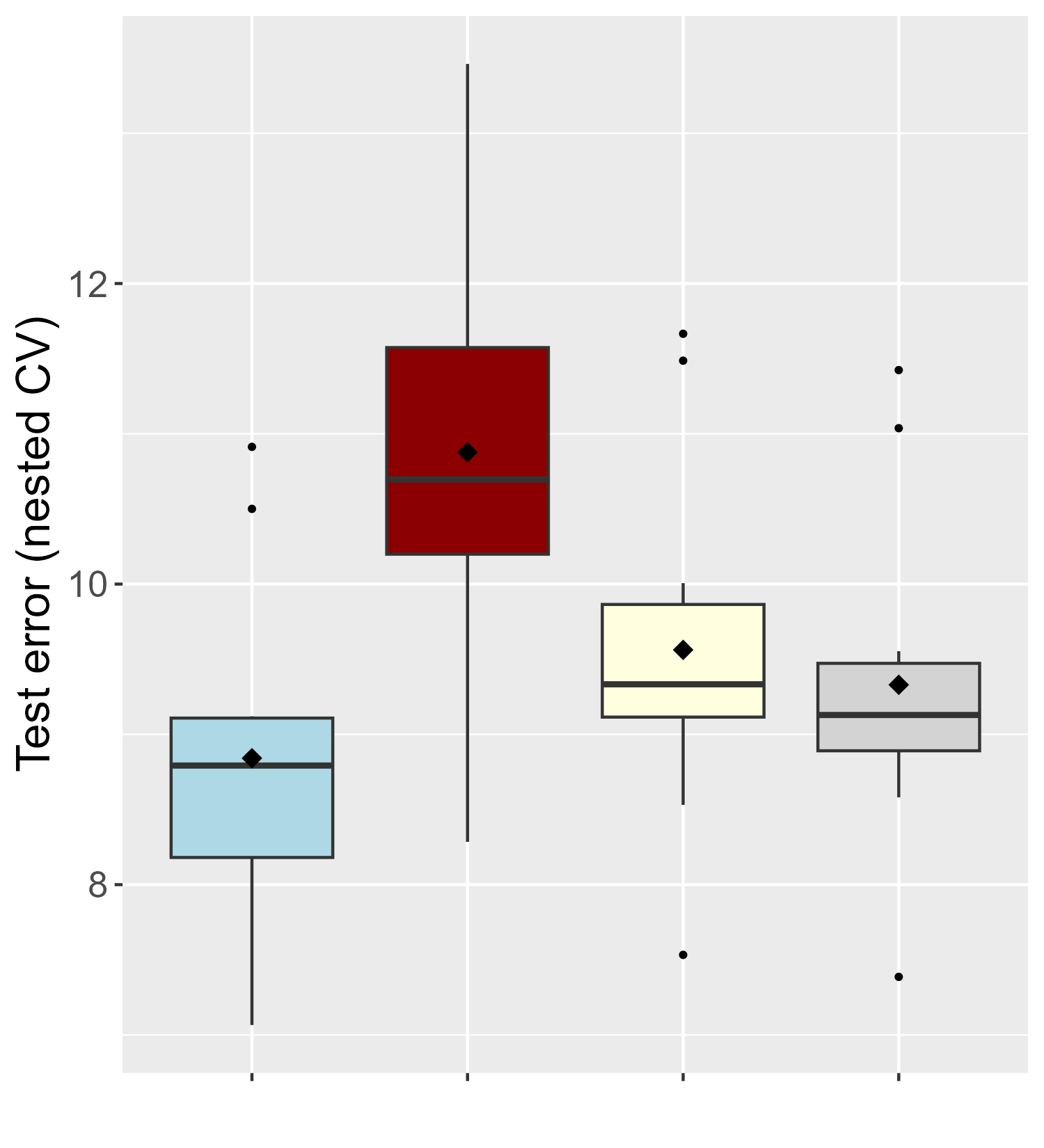}
     \end{subfigure}
     \begin{subfigure}[b]{0.32\textwidth}
         \centering
          \caption{\abalone\ (HD)}
         \label{fig:abalone_hd}
         \includegraphics[width=\textwidth]{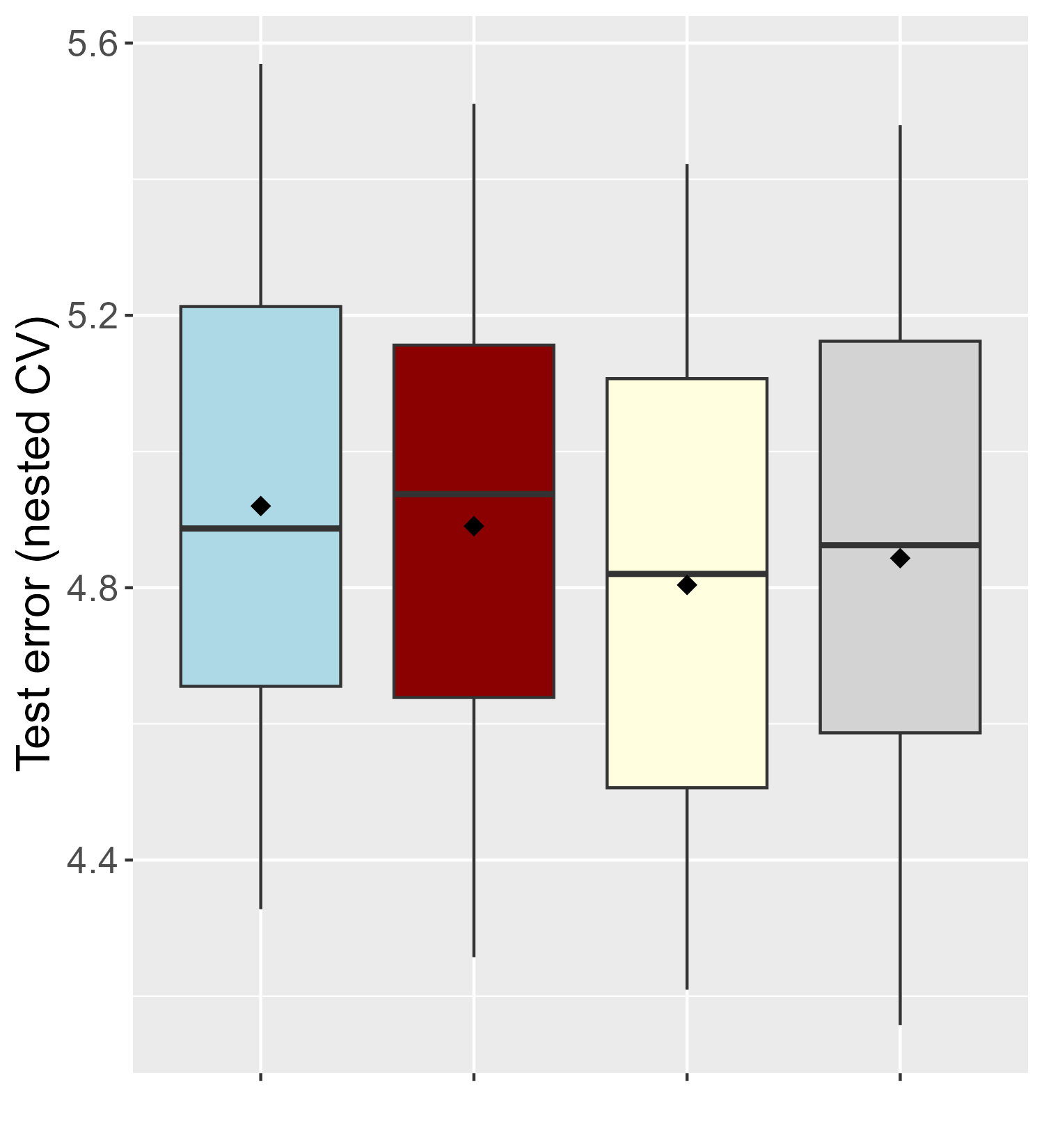}
     \end{subfigure}
       \hfill
    \begin{subfigure}[b]{0.32\textwidth}
         \centering
           \caption{\robot\ (HD)}
         \label{fig:robot_hd}
         \includegraphics[width=\textwidth]{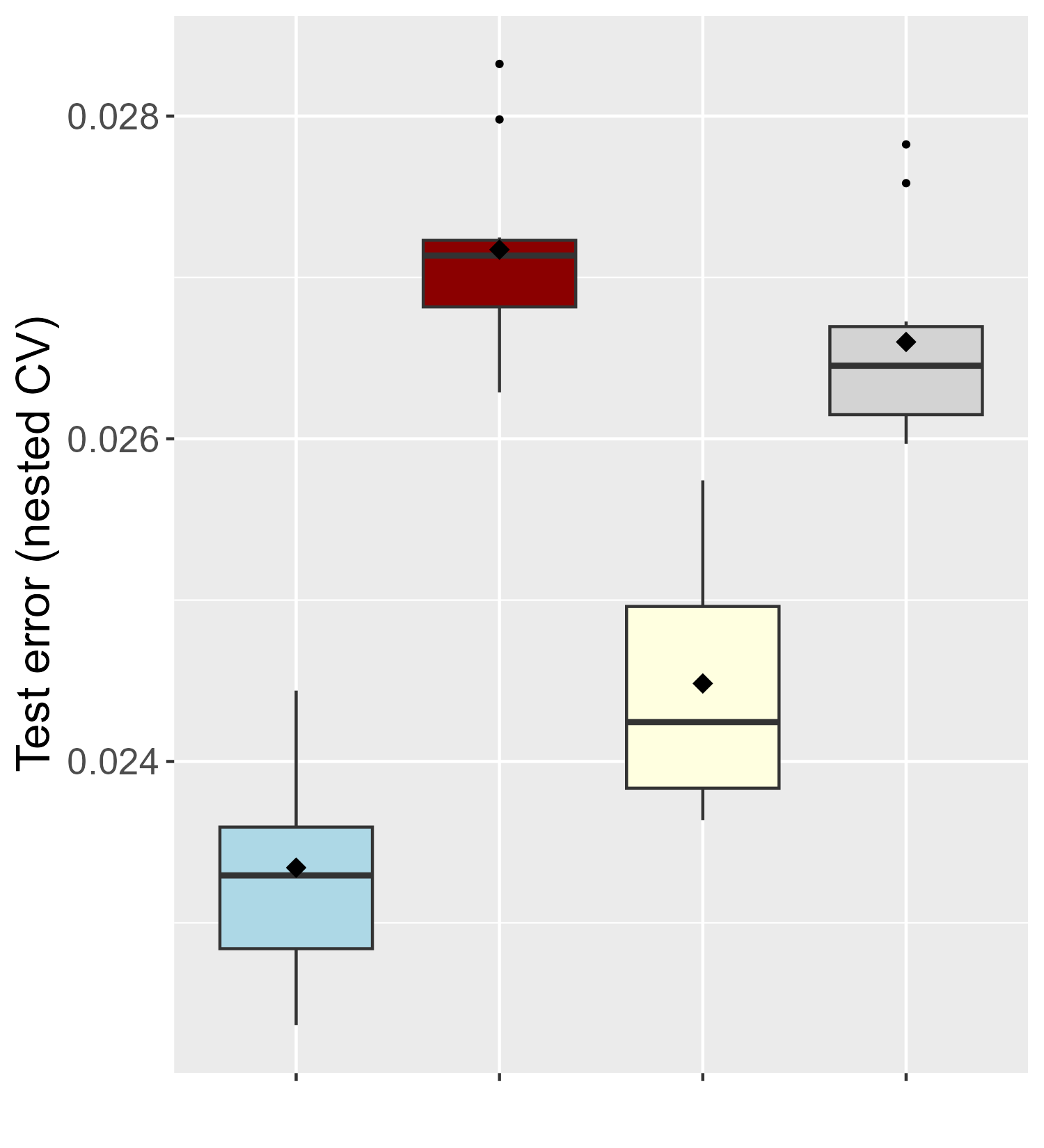}
     \end{subfigure} 
     \begin{subfigure}[b]{0.32\textwidth}
         \centering
           \caption{\calhousing\ (HD)}
         \label{fig:cal_hd}
         \includegraphics[width=\textwidth]{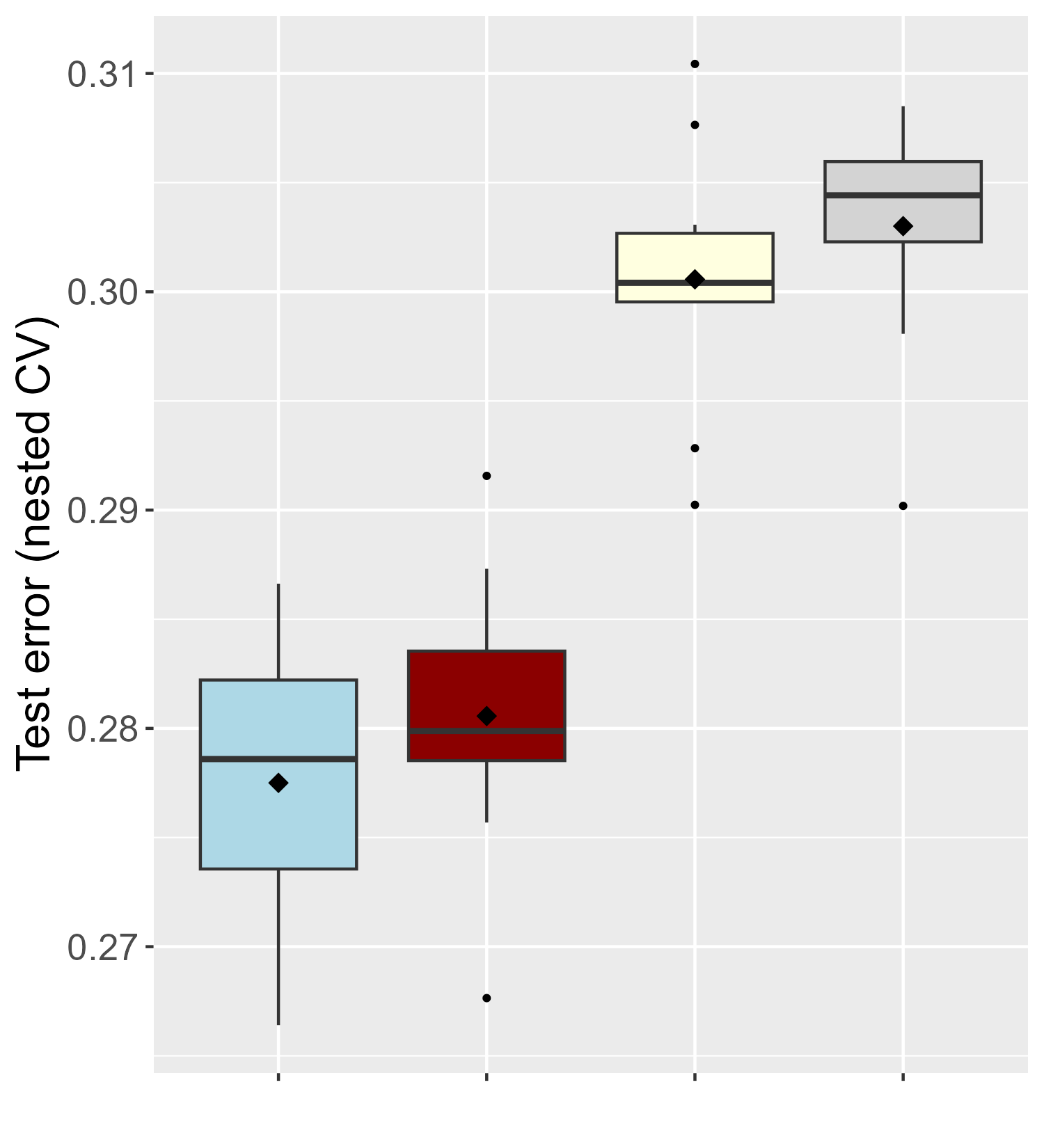}
     \end{subfigure} \hfill
       \begin{subfigure}[b]{0.32\textwidth}
         \centering
           \caption{Plot Legend}
         \label{fig:legend_hd}
         \includegraphics[width=\textwidth]{figures_dataexamples/legend_main_print.png}
     \end{subfigure}
        \caption{Boxplots showing squared errors from nested CV for the real datasets with artificially added covariables (HD). The black diamonds represent the mean of the squared errors. The parameters considered are shown in \Cref{tab:parameters_realdata_noisecov} in the appendix.}
        \label{fig:boxplots_hd}
\end{figure}

In this section, we present results from applying the algorithms from \Cref{sec:algorithms} to real datasets. The datasets were chosen to highlight some differences in the performance of the algorithms that may occur depending on the data available. We do not claim that the selection of datasets is representative. The following datasets were considered.
\begin{itemize}
\item \concrete\ (sample size $n=1030$, number of covariates $d=8$) The response variable in this dataset\footnote[1]{OpenML \citep{OpenML2013} data-ids: $43919$ (\airfoil), $44956$ (\abalone), $44959$ (\concrete)} is the compressive strength of concrete. Covariates are given by age and various ingredients of the concrete. For details, see \citet{concrete}.
\item \airfoil\ ($n=1503$, $d=5$). The origin of this dataset\footnotemark[1] is \citet{airfoil}. The loudness (response) of airfoils spanned in a wind tunnel is reported in this dataset, together with e.g. size of the airfoil, different wind speeds and angles of attack of the wind.
\item \abalone\ ($n=4177$, $d=8$) This dataset\footnotemark[1] comprises the number of rings of abalones (target) which determines the age, and different biological characteristics such as the diameter of the shell or the abalone's weight. The sex is available as categorical variable (with values male, female, infant) and we used binary encoding for this variable (for any of the algorithms).
\item \robot \ ($n=8192$, $d=8$). This dataset\footnote[2]{\url{https://www.dcc.fc.up.pt/~ltorgo/Regression/DataSets.html}} is concerned with the position (target) of the end effector of an robot arm relative to a fixed point, given different angles associated with the robot arm.
\item \calhousing\ ($n=20640$, $d=8$). This dataset\footnotemark[2] originates from \citet{pace1997sparse} and contains data from the 1990 census on housing prices in California. The covariates include e.g. longitude, latitude and median house income. We use the median house value divided by $10^4$ as response.
\end{itemize}
We apply the algorithms \simrsrfnfRealdata, \simrfRealdata, \simifRealdata\ and \simetRealdata\ for any combination of parameter settings shown in \Cref{tab:parameters_realdata} in the appendix. We note that we restricted ourselves to using bootstrap samples and to a fixed minimum node size of five for any of the algorithms. Additional results for different settings are also available in the appendix, see \Cref{subsec:realdata_additional_simulations}.\\
To evaluate the performance, we use nested cross-validation to estimate the mean squared errors. For any of the datasets, we have used five inner folds and five outer folds, and the whole procedure has been iterated two times. In total, we thus have ten estimates per dataset. \Cref{fig:boxplots} shows boxplots of the squared errors and the black diamonds indicate the estimated mean squared errors.\\
In addition, we studied expanded versions of the datasets where $50$ noisy covariates are added artificially. These were chosen such that each of these has nonzero correlation with some of the other (noisy) covariates, but not with all of them. More precisely, we chose $Z = (Z_1,\dots,Z_{50})$ normally distributed with mean zero and the property that for any $j=1,\dots,50$ and possible $m \in \N$,
\begin{align*}
    \cov( Z_{j}, Z_{j+m} ) = \begin{cases}
        1, & m = 0,\\
        \sqrt{\frac{3}{8}} \approx 0.6124, &m=-1,1 \\
        0.375, & m=-2,2\\
0 , & \text{otherwise.}
    \end{cases} 
\end{align*}
We denote these newly created datasets by adding the token ``(HD)''. The parameter settings for the algorithms under this setup are summarized in \Cref{tab:parameters_realdata_noisecov} in the appendix. The results on the real data examples with artificially added noisy covariables can be found in \Cref{fig:boxplots_hd}.\\
In both settings with and without additional feature vectors, \simrsrfnf\ and \simif\ showed a better overall performance compared to \simrf. They showed better or even strongly better results in three models in \Cref{fig:boxplots}, and in two models in \Cref{fig:boxplots_hd}. \simrf\ only outperformed \simif\ in \Cref{fig:boxplots_hd} for the \calhousing\ (HD) data and it showed slightly better CV errors than \simrsrfnfRealdata\ in \Cref{fig:boxplots} for the \concrete\ dataset.\\
In \Cref{subsec:realdata_additional_simulations}, we also show results for Extremely Randomized Trees when using the original sample in the trees instead of using bootstrap samples. In case of the (HD)-settings this showed a much better performance than the bootstrap version and also outperformed the other algorithms in two cases.\\
Finally, we want to note that Interaction Forests with subsampling instead of using bootstrap samples (\texttt{replace} = false) led to an overall better performance than its bootstrap counterpart, as can be seen in the additional numerical results presented in \Cref{subsec:realdata_additional_simulations}.

\section{Summary and conclusion}\label{sec:summary_conclusion}
We have studied limitations of Random Forests for regression in pure interaction scenarios, i.e. when an interaction effect of two (or more) variables is present without main effect. This was motivated from a theoretical investigation of the classical CART splitting criterion used in regression trees. In a large simulation study we found that the performance of Random Forests is indeed poor in such scenarios. The algorithms Interaction Forests and RSRF -- which both use a modified splitting criterion and additional randomness -- have clearly outperformed Random Forests in the pure interactions settings, while not loosing predictive power in the other scenarios considered. In most of the simulations, the Interaction Forests algorithm performed better than RSRF. This is also the case in the real data examples, but not to the same extent. This may suggest to use Interaction Forests for practical applications, rather than RSRF. Solely adding randomness without modifying the splitting criterion as is done by Extremely Randomized Trees led to a general improvement upon Random Forests, but the method was still weaker than Interaction Forests and RSRF in most of the pure interaction settings.\\
In addition to the simulations, we have conducted experiments on real datasets. Applying the algorithms to real datasets is particularly interesting within the present context, because models are often believed to be hierarchical, i.e. interaction effects are accompanied by at least some of the corresponding main effects. For a discussion, see e.g. \citet[Section 1.2]{bien2013lasso}. However, in some of the real data examples, Interaction Forests and RSRF did better than Random Forests / Extremely Randomized Trees, suggesting the existence of cases with hard-to-detect interactions. Possibly, these interaction effects could be ``approximately'' pure (by ``approximately'' we mean an interaction effect with low main effects, which is more realistic than a pure interaction). Though the considered examples may not be representative for real data applications in general, this would be opposed to the common believe that
always hierarchical models should be used in data analysis.\\
Finally, though we restricted ourselves to regression, it is possible that parts of our observations carry over to classification, but this has to be checked in additional research.
\section*{Supplementary Material}
The appendix can be found in Supplementary Material 1 which is available online. Further supplementary material reporting on additional details from the simulation concerning the parameters in RSRF is provided along with code for \rsrf\ and for the simulations online at \url{https://github.com/rblrblrbl/rsrf-code-paper/}. There, we also provide details concerning the software and package versions used in the simulations.  
\section*{Acknowledgements} The authors acknowledge support by the state of Baden-Württemberg through bwHPC.
\section*{Declaration of interest}
None.
\bibliography{library}

\appendix

\renewcommand{\thesection}{\Alph{section}} 
\makeatletter
\renewcommand\@seccntformat[1]{\appendixname\ \csname the#1\endcsname.\hspace{0.5em}}
\makeatother
\newpage

\pagenumbering{arabic}
\renewcommand*{\thepage}{\arabic{page}}

\begin{center}
    \Large\textbf{Supplementary material 1: \\Appendix to ``Pure interaction effects unseen by Random Forests''}
\end{center}

\noindent
\begin{minipage}[t]{0.24\textwidth}
    {\centering Ricardo Blum\footnotemark[1]}
\end{minipage}\hfill
\begin{minipage}[t]{0.24\textwidth}
    {\centering Munir Hiabu\footnotemark[2]}
\end{minipage}\hfill
\begin{minipage}[t]{0.24\textwidth}
    {\centering Enno Mammen\footnotemark[1]}
\end{minipage}\hfill
\begin{minipage}[t]{0.24\textwidth}
    {\centering Joseph Theo Meyer\footnotemark[1]}
\end{minipage}\hfill \\[0.1cm]

\begin{minipage}[t]{0.1\textwidth}
\end{minipage}\hfill
\begin{minipage}[t]{0.7\textwidth}
    \begin{center}  \small \footnotemark[1]University~of~Heidelberg,~Institute~for~Mathematics,\\ Im~Neuenheimer~Feld~205, 69120~Heidelberg, Germany \end{center}
    \begin{center} \small
    \footnotemark[2]University~of~Copenhagen,~Department~of~Mathematical~Sciences,\\
            Universitetsparken~5, 2100~Copenhagen~Ø, Denmark
    \end{center}
\end{minipage}\hfill
\begin{minipage}[t]{0.1\textwidth}
\end{minipage}\hfill
\section*{Structure of the appendix}
The appendix is structured as follows. We give the proof of \Cref{prop:simple_pure_interaction} in \Cref{sec:proofs} and some additional calculations on the splitting criteria mentioned in the main text. \Cref{sec:RSRF_detailed} describes the implementation of RSRF in details, while \Cref{sec:extension_depth} contains a short outline on a possible extension. Lastly, we provide additional material concerning the simulations in \Cref{sec:appendix_simulation}, and the real data applications in \Cref{sec:appendix_realdata}.
\section{Proofs and additional calculations}\label{sec:proofs}
\subsection{Proof of \Cref{prop:simple_pure_interaction}}
\begin{proof}[Proof of \Cref{prop:simple_pure_interaction}]
	Let $t = A \times [0,1] \times A_3 \dots \times A_d$. Then,
	\begin{align*}\label{eq:id_constraint_calc}\begin{split}
		\E&[m_{ \{1,2\}}(X_1,X_2)\1_{(X\in t)}] \\&= \int_t m_{\{1,2\}}(x_1,x_2) p(x_1,\dots, x_d) \dx x_1 \cdots \dx x_{d} \\&= \int_{A} \int_{[0,1]} m_{\{12\}}(x_1,x_2) p(x_1,x_2) \dx x_2 \dx x_1 \int_{A_3\times \cdots \times A_{d}} p(x_3,\dots,x_d) \dx x_{3}\cdots \dx x_{d} \\&= 0,
  \end{split}
	\end{align*}	
	due to the identification constraint. Thus, $\E[m_{ \{1,2\} }(X_1,X_2)| X\in t ] = 0$. More generally,
 for any $J\supseteq \{1,2\}$, 
\begin{equation}\label{eq:id_constraint_calc_general} \E[m_J(X_J)|X\in t] = 0,\end{equation}
 which follows from analogous calculations and the fact that \begin{align*}
     p(x_{1},\dots,x_d) = p(x_{-J} | x_J ) p(x_J) = p(x_{-J}|x_{J\setminus \{1,2\} } ) p(x_J),
 \end{align*}
due to independence. Here, we used the notation $x_J = (x_j:j\in J)$ and $x_{-J} = (x_j: j\notin J)$.
Now, from \eqref{eq:id_constraint_calc_general} and the assumption, we see that
	\begin{align}\label{eq:summation}
		\E[m(X)|X\in t] = \sum_{ \substack{ u \subseteq \{1,\dots, d\}	 \\ 1 \notin u, 2 \notin u } } \E[m_u(X_u) | X \in t].
	\end{align}
For each $u$ as in the sum,
	\begin{align}\label{eq:calc_summands}
		\begin{split}\E[m_u(X_u) | X \in t] &=  \frac{ \E\big[m_u(X_u)\1_{(X_1 \in A, X_2 \in [0,1])} \1_{(X_3\in A_3)}\cdots \1_{(X_d \in A_d)} \big] }{\PP(X_1 \in A, X_2 \in [0,1], X_3 \in A_3,\dots, X_d \in A_d) }\\&= \E[m_u(X_u)|X_3 \in A_3,\dots, X_d \in A_d],
  \end{split}
	\end{align}
	using independence of $(X_1,X_2)$ and $(X_3,\dots,X_d)$ in the last step. Note that from letting $A=[0,1]$, by  \eqref{eq:id_constraint_calc_general}, we also have
  \begin{align}\label{eq:calc_12}
     \E[m_{J}(X_J)|X \in t] = \E[m_{J}(X_J)|X_3 \in A_3,\dots, X_d \in A_d],
 \end{align}
for any $J \supseteq \{1,2\}$. The result then follows from \eqref{eq:summation} in view of \eqref{eq:calc_summands} and \eqref{eq:calc_12}.
\end{proof}
\subsection{Equivalence of splitting criteria}
The following calculations on the equivalence of splitting criteria is added for the sake of completeness.
\begin{proposition}\label{prop:equivalence_criteria}
    Let $t_1,t_2$ be a partition of $t \subseteq \R^d$ and $(Y_1,X_1),\dots (Y_n,X_n) \in \R \times \R^d$. Let us write $\hat\mu (A) = \frac{1}{\#A}\sum_{i:X_i\in A} Y_i$ with $\# A = \#\{i: X_i \in A\}$, and abbreviate $T_1:=\#t_1$, $T_2 := \#t_2$, $T := \# t$. Consider
    \begin{align*}
          \widehat{\zwei}(t; t_1,t_2) &=     \frac{ T_1}{T } \big[  \hat{\mu}(t_{1}) - \hat{\mu}(t) \big]^2 +    \frac{ T_2}{T } \big[  \hat{\mu}(t_{2}) - \hat{\mu}(t) \big]^2,
		\\ \mathcal{M}(t;t_1,t_2)&= \frac{ \left( \sum_{i: X_i \in t_1} Y_i \right)^2 }{T_1} + \frac{ \left( \sum_{i: X_i \in t_2} Y_i \right)^2 }{T_2} \text{ and } \\
    \mathcal{V}(t;t_1,t_2) &= \sum_{i: X_i \in t_1} \big(Y_i - \hat{\mu}(t_1) \big)^2 + \sum_{i:X_i\in t_2}\big(Y_i-\hat\mu(t_2) \big)^2.
\end{align*}
Then,
\begin{align*}
    \max_{t_1,t_2}\ \widehat{\zwei}(t;t_1,t_2) = \max_{t_1,t_2}\ \mathcal{M}(t;t_1,t_2) =\min_{t_1,t_2}\ \mathcal{V}(t;t_1,t_2),
\end{align*}
where the maximum/minimum is over a set of partitions $(t_1,t_2)$ of $t$.
\end{proposition}
\begin{proof}
	\begin{align*}
		     T \widehat{\zwei}(t;t_1,t_2) &=T_1 [\hat\mu(t)- \hat\mu(t_1)]^2 + T_2 [\hat\mu(t)- \hat\mu(t_2)]^2 \\
		     &=T_1 [ \hat\mu(t)^2 - 2\hat \mu(t) \hat\mu(t_1) + \hat\mu(t_1)^2 ] + T_2[ \hat\mu(t)^2 - 2\hat \mu(t) \hat\mu(t_2) + \hat\mu(t_2)^2 ] \\
		     &=T \hat \mu(t)^2 -  2\hat\mu(t) \Big\lbrace T_1 \hat\mu(t_1) +T_2 \hat\mu(t_2) \Big\rbrace +  T_1 \hat\mu(t_1)^2 +T_2 \hat\mu(t_2)^2\\
             &= -T \hat\mu(t)^2  + \mathcal{M}(t;t_1,t_2),
	\end{align*}
	and the first summand $-T\hat\mu(t)^2$ does not depend on $t_1,t_2$. Similarly, we see
    \begin{align*}
        \mathcal{V}(t;t_1,t_2) &= \sum_{i: X_i \in t_1} \big(Y_i - \hat{\mu}(t_1) \big)^2 + \sum_{i:X_i\in t_2}\big(Y_i-\hat\mu(t_2) \big)^2 \\
        &= \sum_{i: X_i \in t_1}Y_i^2 - 2\Big( \sum_{i: X_i \in t_1} Y_i \Big) \hat\mu(t_1) + T_1 \big(\hat{\mu}(t_1) \big)^2 \\
        &\qquad + \sum_{i: X_i \in t_2}Y_i^2 - 2\Big( \sum_{i: X_i \in t_2} Y_i \bigg) \hat\mu(t_2) + T_2 \big(\hat{\mu}(t_2) \big)^2 \\
        &=\sum_{i=1}^n Y_i^2 - \big[ T_1\hat\mu (t_1)^2 + T_2 \hat\mu(t_2)^2 \big] \\
        &=\sum_{i=1}^n Y_i^2 - \mathcal{M}(t;t_1,t_2),
    \end{align*}
    and the first summand does not depend on $t_1$ and $t_2$.
\end{proof}

\section{Detailed description of the \rsrf\ algorithm}\label{sec:RSRF_detailed}
 We provide details on the implementation of RSRF. An overview of the tree growing algorithm is given below in \Cref{def:algo_practise} and an overview over all the parameters is given in \Cref{table:parameters_rsrf}. In \Cref{subsec:details_algorithm} we shall introduce all remaining parts of the algorithm. Recall the definition of $\widehat{\zwei}$ from equation \eqref{eq:def_zwei_hut}.
\begin{remark}[Overview of the \rsrf\ tree growing procedure]\label{def:algo_practise}
        Let $(x_i,y_i)_{i=1,\dots,n}$ be given. Starting with $T_0=\{\rootcell\}$ with $\rootcell = [0,1]^d$, for $m=0,2,4,\dots$ apply the following steps to all current leaf nodes $t \in T_{m}$ which contain at least $\nodesize$ many data points.
        \begin{enumerate}
            \item Draw $\width$ many pairs $(j^w,c^w)$ by choosing $j^w \in \{1,\dots,d\}$ uniformly at random, and by drawing $c^w$ from the uniform distribution on the data points $$ \{x_{i,j^w} \ : x_i \in t\} \setminus  \max\{  x_{i,j^w} : x_i \in t \}.$$
            \item For each $w=1,\dots,W$ split $t$ at $(j^w,c^w)$ into $t^w_{1}$ and $t^w_{2}$ and then split $t^w_{1}$ and $t^w_{2}$ according to the Sample-CART criterion in \Cref{def:sample-cart}. This gives a partition $\{t^{w}_{1,1}, t^{w}_{1,2}, t^{w}_{2,1}, t^{w}_{2,2} \}$, for each $w$.    If \cartcart\ is set to \quot{true}, additionally consider $w=0$ where $t$ is split using the Sample-CART criterion into $t^0_{1}, t^0_{2}$, and then these cells are again split using Sample-CART.       
            \item Choose the splits with index $w_{\text{best}} \in \underset{w}{\arg\max}\ \widehat{\zwei}\left( t;t^{w}_{1,1},t^{w}_{1,2},t^{w}_{2,1},t^{w}_{2,2} \right)$.
             \item Add $t_{k,j}^{w_{\text{best}}}$ to $T_{m+2}$ for $k,j\in\{1,2\}$.
        \end{enumerate}
    \end{remark}
    The parameter $\width$ is called width parameter. If \cartcart\ is \quot{false}, then there are $\width$ candidate partitions generated in each iteration step. Recall that we refer to the procedure to generate one of the candidate partitions as a Random-CART-step. In the other case, when \cartcart\ is \quot{true}, the number of candidates is $\width +1$, due to adding the CART-CART-step.
   
\begin{remark}\label{remark:relation_2D_CART}
    If the width $W$ is infinity, then the algorithm can be seen as a two-dimensional extension of the Sample-CART criterion, where the optimization is jointly over the first split, and the two splits for the two daughter cells.
\end{remark}
\begin{remark}
        We distinguish different approaches for determining allowed split coordinates for the Sample-CART splits used in (b) in the above algorithm. These are called \texttt{mtrymodes} and the details are to be found in \Cref{subsec:mtrymodes}.
    \end{remark}
    \subsection{Further details on \rsrf}\label{subsec:details_algorithm}
    We list the remaining features of the algorithm. An overview over the parameters can be found in \Cref{table:parameters_rsrf}.
    \begin{table}[t!]
    \centering
	\begin{tabular}{l p{4cm} p{4cm}}
			$\nodesize$ & \multicolumn{2}{p{8cm}}{Minimum node size for a cell to be split} \\
			\hline
			\simwidth ($\width$) & \multicolumn{2}{p{8cm}}{Number of candidate partitions} \\
			\hline
                \simntrees ($\ntrees$) & \multicolumn{2}{p{8cm}}{Number of trees} \\
                  \hline
              $\replace$ & \multicolumn{2}{p{8cm}}{If \quot{true} (\quot{false}), bootstrap samples (subsamples) are used.}\\
              \hline
             $\cartcart$ & \multicolumn{2}{p{8cm}}{If \quot{true}, then the CART-CART step is added to the candidate splits.} \\
			\hline
            \hline
            $\mtrymode$ & \multicolumn{2}{p{8cm}}{Determines whether possible split coordinates remain fixed over candidate partitions, or not.} \\
             &\multicolumn{1}{c}{$\mtrymodeaf$: } & \multicolumn{1}{c}{$\mtrymodenf$:} \\
            \hline 
                $\mtrycartcart$ & (not available) & Number of possible coordinates to choose for the CART splits in CART-CART step. \\
                \hline
                $\mtryrandom$ & Number of possible coordinates to choose from for the Random split. & (not available)\\ \hline
                $\mtryrandomcart$ & \multicolumn{2}{p{8cm}}{Number of coordinates to choose from when placing a CART-split in a Random-CART step}  \\
	\end{tabular}
  \caption{Overview over parameters for \rsrf.}
   \label{table:parameters_rsrf}
\end{table}
    \subsubsection*{Placing the random split point}
    Whenever we place a random split at a cell $t = \bigtimes_{j=1}^d t^{(j)}$, we first choose the dimension $j$ uniformly at random from $\{1,\dots,d\}$. Then, a split point is drawn uniformly at random from the data points $\{x_{i,j} \ : x_i \in t\} \setminus  \max\{  x_{i,j} : x_i \in t \}$.
    \subsubsection*{Stopping condition} A current leaf will only be split into new leafs if it contains at least $\nodesize \in \N$ number of data points.
    \subsubsection*{From trees to a forest}
    Similar to Random Forests, we generate an ensemble of trees based on bootstrapping or subsampling. The single tree predictions will then be averaged. Denote by $\hat{m}_{\hat{T}}$ the estimator corresponding to the tree generated through the RSRF tree growing procedure, based on data points $(x_i,y_i)$, $i=1,\dots,n$. That is, the prediction at some $x\in[0,1]^d$ is given by
    \begin{align*}
        \hat{m}_{\hat{T}}(x) = \sum_{ t \in \hat{T} } \Big( \frac{1}{ \card{t} } \sum_{i: x_i \in t } y_i \Big)\1_{(x \in t)},
    \end{align*}
    where the sum is over the leaves in $\hat{T}$. In case $\replace = \quot{\text{true} }$, we draw \ntrees \ bootstrap samples
    \begin{align*}
        (x_{i}^{*b},y_{i}^{*b})_{i=1,\dots,n}, \quad b =1 ,\dots, \ntrees
    \end{align*}
    from $(x_i,y_i)_{i=1,\dots,n}$ with replacement. For each of these bootstrap samples, obtain an estimator $\hat{m}^{b}_{\hat{T}}$. Then, this results in the final \rsrf\ estimator
    \begin{align*}
        \hat{m}^{\text{forest}}(\cdot) = \frac{1}{B}\sum_{b=1}^{\ntrees} \hat{m}_{\hat{T}}^b(\cdot).
    \end{align*}
    In case $\replace = \quot{\text{false} }$, subsamples (without replacement) are used. The subsample size is set to $0.632$ in accordance with the default setting in the Random Forest implementation \texttt{ranger} \citep{ranger}.
   
    \subsubsection{Different mtry-modes and its related mtry parameters.}\label{subsec:mtrymodes}
    \label{pageref:mtrymode}We distinguish two variants for determining which coordinates are allowed to split on. As this is related to the \texttt{mtry} parameter in Random Forests we call it \quot{\mtrymode} and its values are \mtrymodeaf \ and \mtrymodenf. The key difference is whether the possible split coordinates remain fixed among candidate splits or not.\\
    First, when $\mtrymode = \mtrymodenf$, for the current candidate splits, the possible split coordinates are drawn independently of each other. Here, we have two mtry parameters:\\\mtryrandomcart\ determines the number of possible split coordinates for the CART splits within a Random-CART step. \mtrycartcart \ determines the number of possible split coordinates for the splits in a CART-CART step (thus, it only applies if \cartcart \ is set to \quot{true}). \\
    Secondly, if $\mtrymode = \mtrymodeaf$, then we first draw a subset $J \subseteq \{1,\dots, d\}$ of size \mtryrandom \ and two subsets $J_{1}, J_{2} \subseteq \{ 1,\dots, d\}$ of size \mtryrandomcart. These remain fixed for all candidate splits in the current iteration step. $J$ determines the possible coordinates for splitting the current cell $t$, and $J_{1}, J_{2}$ determines the possible coordinates for its daughter cells. \\ 
    \begin{remark}\label{remark:mtrymode}
      Following \Cref{remark:relation_2D_CART}, the procedure for \texttt{mtrymode = fixed} is analogous to \texttt{mtry} in Random Forests. The \texttt{mtrymode = not-fixed} version, however, is more random, as the first split is always a full random split and not restricted to be taken from a particular subset of $\{1,\dots,d\}$.
    \end{remark}
 
 \section{Extensions to arbitrary depth} \label{sec:extension_depth}
    The \rsrf\ algorithm from \Cref{subsec:rsrf} can be extended by introducing the depth parameter $\depth \in \{2,3,\dots\}$.
    Suppose we have a cell $t$. Starting with $t$ we can iteratively split all current end cells evolving from $t$ by placing random splits. This is repeated $\depth - 1$ times. Afterwards, a Sample-CART split is placed for each end cell. Clearly, the cell $t$ is thus partitioned into $2^{\depth}$ cells. When evaluating the candidate splits, one may then use
    \begin{align*}
        \widehat{\zwei}\left( t;t_a, a \in \{1,2\}^D \right) = \sum_{a \in \{1,2\}^D }\frac{\card{t_a}}{\card{t}} \big[ \hat{\mu}(t_a) - \hat{\mu}(t) \big]^2.
    \end{align*}
    Among $W$ candidate partitions of $t$, the one which maximizes $\widehat{\zwei}$ is chosen. Though we restricted ourselves to the case $D = 2$ in this paper, below, we include a short remark on Random Forests applied to an order-$3$ pure interaction. 
    \begin{remark}
        In \Cref{fig:example_3interact}, simulation results using Random Forests on a model with pure interaction of order $3$ are shown. In contrast to the example for $D=2$ in \Cref{fig:plotF_new} in the main text, here, $\texttt{mtry}=1$ (forcing splits in any coordinate) is clearly the best choice, while $\texttt{mtry}=6$ seems to catch up on a large scale.
 \begin{figure}[t]
        \centering
        \includegraphics[scale=0.75]{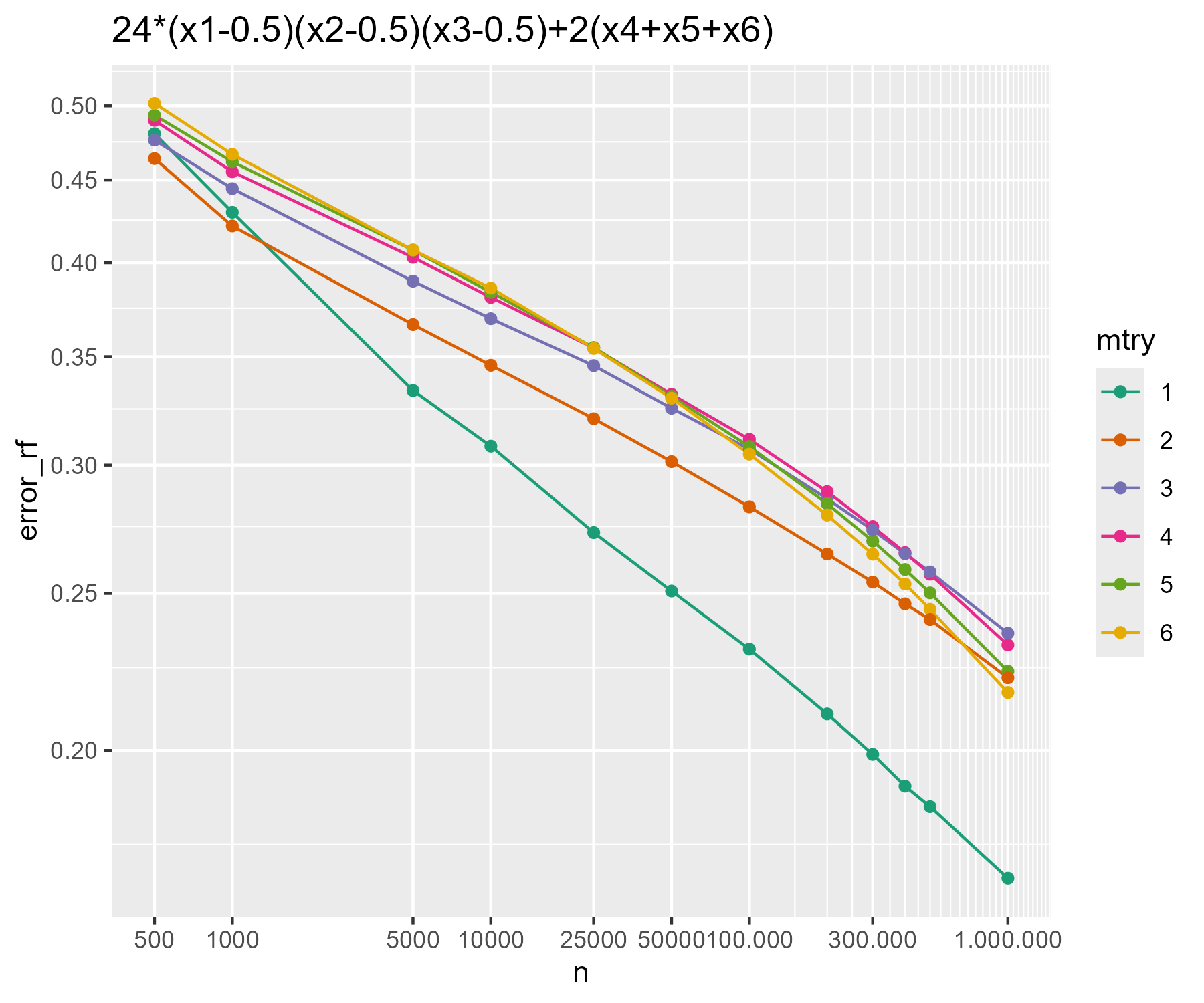}
        \caption{Estimated mean squared error using Random Forest plotted at $\log$-scales for different sample sizes $n$, for the regression model $Y=24(X_1-0.5)(X_2-0.5)(X_3-0.5) + 2(X_4+X_5+X_6) + \varepsilon$ with $(X_1,\dots, X_6)$ uniform on $[0,1]^6$ and $\varepsilon \sim \mathcal{N}(0,1)$. The number of trees was set to $100$ and node size was set to $5$. For each sample size, $100$ simulations were carried out.}
        \label{fig:example_3interact}
    \end{figure}
    \end{remark}

\section{Appendix to the simulation section}\label{sec:appendix_simulation}
In this section, we provide additional simulation results and collect further details on the simulation results presented in the main text.
\subsection{Additional simulation results for different $\mtrymode$}\label{sec:simulations_rsrf_af}
We also performed simulations for \rsrf\ with \mtrymode\ set to \mtrymodeaf\ and refer to this setup as \simrsrfaf. The difference between the two versions of \rsrf\ is that for $\simrsrfaf$, the covariates where splits are allowed to be placed (in a Random-CART-step) are fixed among the candidate partitions, see \Cref{subsec:mtrymodes} for details. The setting and results in this case are collected in tables \ref{tab:parameters_rsrf_af},  \ref{table:results_simulation_opt} and \ref{table:results_simulation_cv_rsrf_af}.
Interestingly, there is not a big difference between the results for \simrsrfaf\ and \simrsrfnf. 
\begin{table}[h]
    \centering
    \begin{tabular}{c l l}
        Algorithm & Parameter & Value / Range \\
        \hline
        \simrsrfaf & \simcartcart     & True, False \\
                   & \simreplace      & True, False \\
                   & \simwidth        & $1,2,\dots, 15 \ (d=4,6,10)$ \\
                   & & \qquad resp.  $1,2,\dots,30\ (d=30)$ \\
                   & \simmtryrandom   & $1,2,\dots, d$ \\
                   & \simmtryrc       & $1,2,\dots, d$ \\
                   & \simminnodesize  & $5,6,\dots, 30$ \\
                   & \simntrees       & $100$ \\
                   & \mtrymode        & \texttt{fixed} \\
            
    \end{tabular}
    \caption{Parameter settings for \simrsrfaf\ used in the simulation study.}
    \label{tab:parameters_rsrf_af}
\end{table}
\begin{table}[h]
    \centering
    \begin{tabular}{c l C C C} 
        Model &Algorithm & d=4 & d=10 & d=30 \\
        \hline
 \rowcolor{gray!15}\cellcolor{white}      
            &\simrsrfnf (opt) & 0.201 \ (0.028) & 0.261 \ (0.037)&  0.369 \ (0.062) \\
 \modelPureType    &\simrsrfaf (opt) & 0.201 \ (0.028) & 0.279 \ (0.040) & 0.379 \ (0.068)  \\
\rowcolor{gray!15}\cellcolor{white}   
        &\simif (opt) & 0.160\ (0.024) & 0.220\ (0.030) & 0.317\ (0.054) \\
     &\simrf (opt) & 0.311\ (0.069) & 0.698\ (0.203) &  1.336\ (0.292) \\
 \rowcolor{gray!15}\cellcolor{white}                           
      &\simet (opt) & 0.207 \ (0.048) & 0.403 \ (0.148) &  0.869 \ (0.407) \\
         \hline \hline 
                        &\simrsrfnf (opt) & 0.422 \ (0.045)  & 0.538\ (0.063) & 0.674 \ (0.067) \\
 \rowcolor{gray!15}\cellcolor{white}\modelHierarchical   
                        &\simrsrfaf (opt) & 0.417 \ (0.044) & 0.548\ (0.064) & 0.690 \ (0.069) \\
   &\simif (opt) & 0.383\ (0.042) & 0.512\ (0.058) & 0.624\ (0.062) \\
 \rowcolor{gray!15}\cellcolor{white}                       
                        &\simrf (opt) & 0.416\ (0.050) &0.554\ (0.066) & 0.675\ (0.069) \\                    
                        &\simet (opt) & 0.354 \ (0.042)  & 0.449 \ (0.053) & 0.531 \ (0.059)  \\

        \hline \hline
          \rowcolor{gray!15}\cellcolor{white}                       
                    &\simrsrfnf (opt) & 0.361 \ (0.039) & 0.460\ (0.053) & 0.571 \ (0.053) \\
         \modelAdditive                        
                        &\simrsrfaf (opt) & 0.358\ (0.041) & 0.469\ (0.053) & 0.576\ (0.055) \\
        \rowcolor{gray!15}\cellcolor{white}       
                        &\simif (opt)  & 0.336\ (0.038) & 0.420\ (0.047) &  0.502\ (0.048)\\

                        &\simrf (opt) & 0.343\ (0.039)& 0.458\ (0.050)&  0.551\ (0.053)\\

        \rowcolor{gray!15}\cellcolor{white}                     
                        &\simet (opt) &  0.293 \ (0.033) & 0.366 \ (0.046) &  0.423 \ (0.045) \\

        \hline \hline
                   &\simrsrfnf (opt) & 0.148\ (0.026) & 0.191\ (0.021)& 0.219 \ (0.023) \\
         \rowcolor{gray!15}\cellcolor{white} \modelPureZwei                               
                        &\simrsrfaf (opt) & 0.145 \ (0.022) & 0.192 \ (0.020)& 0.222\ (0.022)  \\
                        &\simif (opt)  & 0.112\ (0.018) & 0.168\ (0.023) &  0.207\ (0.020) \\
       \rowcolor{gray!15}\cellcolor{white}       
                        &\simrf (opt) & 0.184 \ (0.025) & 0.231 \ (0.021) & 0.245 \ (0.022) \\
                        &\simet (opt) & 0.116 \ (0.019) & 0.187 \ (0.018) &  0.210 \ (0.019)\\

                        \hline \hline
    \end{tabular}    
\newline
\vspace*{1cm}
\newline
    \begin{tabular}{c l C} 
        Model &Algorithm & d=6  \\
        \hline
 \rowcolor{gray!15}\cellcolor{white}      
                        &\simrsrfnf (opt) &  0.195\ (0.032)
\\
 \modelPureDrei
                        &\simrsrfaf (opt) & 0.190\ (0.030)
 \\
\rowcolor{gray!15}\cellcolor{white}   
                        &\simif (opt)  & 0.151\ (0.028)\\
                        &\simrf (opt) & 0.518\ (0.063) \\
 \rowcolor{gray!15}\cellcolor{white}                          
                        &\simet (opt) & 0.429\ (0.041)  \\

        \hline \hline 
  \end{tabular}   
    
    \caption{Reported mean squared error estimates for different simulations with parameter choice using (opt). Standard deviations are provided in brackets.}
    \label{table:results_simulation_opt}
\end{table}

\begin{table}[h]
    \centering
    \begin{tabular}{c l C C C C} 
        Model &Algorithm & d=4 & d=10 & d=30 & d = 6 \\
        \hline
 \rowcolor{gray!15}\cellcolor{white}    
           \modelPureType             &\simrsrfaf (CV) & 0.212\ (0.034) & 0.272\ (0.039) & 0.401\ (0.075) &- \\

\modelHierarchical
      &\simrsrfaf (CV) & 0.428\ (0.050) & 0.561\ (0.067) & 0.689\ (0.074) &-  \\ 
      
         \rowcolor{gray!15}\cellcolor{white}                      
           \modelAdditive           &\simrsrfaf (CV) &   0.375\ (0.043) &  0.492\ (0.061) & 0.585\ (0.053) &-  \\
                                           
            \modelPureZwei      &\simrsrfaf (CV)  & 0.152\ (0.025) & 0.198\ (0.021) & 0.230\ (0.025) & -\\
        \rowcolor{gray!15}\cellcolor{white}     \modelPureDrei&\simrsrfaf (CV) & - &- & - & 0.192\ (0.030)  
    \end{tabular}
   
    \caption{Reported mean squared error estimates for different simulations with parameter choice using (CV). Standard deviations are provided in brackets.}
    \label{table:results_simulation_cv_rsrf_af}
\end{table}

\begin{table}[h!]
    \centering
    \begin{tabular}{c l C C C} 
        Model &Algorithm & d=4 & d=10 & d=30 \\
        \hline

 \rowcolor{gray!15}\cellcolor{white}      
                        &\simrsrfnf\ & 0.097 & 0.121 & 0.180
 \\
                        & \simrsrfaf\ & 0.099 & 0.125 & 0.185
\\
  \rowcolor{gray!15}\cellcolor{white}   \modelPureType      
       &\simif\   & 0.080 & 0.102 & 0.150
 \\    
                &\simrf\ & 0.146 & 0.321 & 0.619
 \\          
   \rowcolor{gray!15}\cellcolor{white}      
                      &\simet\ & 0.096 & 0.182 & 0.399
 \\                         
        \hline \hline 

                   &\simrsrfnf\ & 0.053 & 0.068 & 0.085
 \\
  \rowcolor{gray!15}\cellcolor{white}      
                      &\simrsrfaf\ & 0.053 & 0.069 & 0.085
  \\
  \modelHierarchical        &\simif\ & 0.049 & 0.063 & 0.077
 \\   
  \rowcolor{gray!15}\cellcolor{white} &\simrf & 0.052 & 0.068 & 0.084
 \\          
                        &\simet\ & 0.045 & 0.056 & 0.067
 \\                         
        \hline \hline 

  \rowcolor{gray!15}\cellcolor{white}      
                        &\simrsrfnf\ & 0.062 & 0.080 & 0.096  \\
                        & \simrsrfaf\ & 0.063 & 0.083 & 0.098 \\
  \rowcolor{gray!15}\cellcolor{white}   \modelAdditive
       &\simif\    & 0.057 & 0.073 & 0.086 \\    
                &\simrf\ & 0.058 & 0.078 & 0.093\\          
   \rowcolor{gray!15}\cellcolor{white}      
                      &\simet\  & 0.050 & 0.063 & 0.072 \\                         
        \hline \hline

                   &\simrsrfnf\ & 0.068 & 0.087 & 0.099 
\\
  \rowcolor{gray!15}\cellcolor{white}      
                      &\simrsrfaf\  & 0.067 & 0.087 & 0.100

 \\
  \modelPureZwei        &\simif\   & 0.056 & 0.076 & 0.093

\\    
  \rowcolor{gray!15}\cellcolor{white}  &\simrf\ & 0.083 & 0.104 & 0.110

\\          
                        &\simet\ & 0.057 & 0.084 & 0.091

 \\                         
        \hline \hline 
    \end{tabular}
    \newline
\vspace*{1cm}
\newline
 \begin{tabular}{c l C} 
        &Algorithm & d=6  \\
        \hline

 \rowcolor{gray!15}\cellcolor{white}      
                        &\simrsrfnf\  & 0.185
\\
                       &\simrsrfaf\  & 0.187

\\

     \rowcolor{gray!15}\cellcolor{white}      
        \modelPureDrei   &\simif\  & 0.150

 \\
                   &\simrf\   & 0.497

\\ 
      \rowcolor{gray!15}\cellcolor{white}                        &\simet\ & 0.407 \\
              \hline \hline 

  \end{tabular}

    \caption{Quotients of mean squared error estimates for \simif, \simrsrfnf, \simrsrfaf, \simrf, \simet, divided by the mean squared error estimates when using the mean of the responses as estimator (\simmeanY).}
    \label{table:results_cv_normed_meanY}
\end{table}

\begin{table}[h!]
    \centering
    \begin{tabular}{c l C C C} 
        Model &Algorithm & d=4 & d=10 & d=30 \\
        \hline
 \rowcolor{gray!15}\cellcolor{white}      
                        &\simrsrfnf\ & 0.162 & 0.106 & 0.099 \\
                        & \simrsrfaf\ & 0.165 & 0.110 & 0.102\\
  \rowcolor{gray!15}\cellcolor{white}   \modelPureType      
       &\simif\  & 0.133 & 0.090 & 0.083 \\    
                &\simrf\ & 0.243 & 0.281 & 0.341 \\          
   \rowcolor{gray!15}\cellcolor{white}      
                      &\simet\ & 0.160 & 0.159 & 0.220 \\                         
        \hline \hline 

                   &\simrsrfnf\ & 0.207 & 0.095 & 0.065 \\
  \rowcolor{gray!15}\cellcolor{white}      
                      &\simrsrfaf\ & 0.208 & 0.097 & 0.066  \\
  \modelHierarchical        &\simif\ & 0.192 & 0.089 & 0.059 \\   
  \rowcolor{gray!15}\cellcolor{white} &\simrf & 0.203 & 0.096 & 0.064 \\          
                        &\simet\ 
                        & 0.176 & 0.078 & 0.051\\                         
        \hline \hline 

  \rowcolor{gray!15}\cellcolor{white}      
                        &\simrsrfnf\  & 0.209 & 0.111 & 0.074 \\

                        & \simrsrfaf\   & 0.211 & 0.116 & 0.076 \\
                        \rowcolor{gray!15}\cellcolor{white}   \modelAdditive
       &\simif\   & 0.193 & 0.101 & 0.066  \\    
                &\simrf\ & 0.197 & 0.108 & 0.072 \\          
   \rowcolor{gray!15}\cellcolor{white}      
                      &\simet\ & 0.168 & 0.087 & 0.056  \\                         
        \hline \hline 
                   &\simrsrfnf\ & 0.134 & 0.105 & 0.072\\
  \rowcolor{gray!15}\cellcolor{white}      
                      &\simrsrfaf\  & 0.131 & 0.106 & 0.073 \\
  \modelPureZwei        &\simif\   & 0.110 & 0.092 & 0.068\\    
  \rowcolor{gray!15}\cellcolor{white}  &\simrf\ & 0.163 & 0.126 & 0.080\\          
                        &\simet\ & 0.111 & 0.101 & 0.067 \\                         
        \hline \hline 
    \end{tabular}
    \newline
\vspace*{1cm}
\newline
 \begin{tabular}{c l C}
        &Algorithm & d=6  \\
        \hline
 \rowcolor{gray!15}\cellcolor{white}      
                        &\simrsrfnf\ & 0.147 \\
                       &\simrsrfaf\ & 0.149\\
     \rowcolor{gray!15}\cellcolor{white}     
 
        \modelPureDrei   &\simif\  & 0.119 \\
                   &\simrf\  & 0.396 \\ 
      \rowcolor{gray!15}\cellcolor{white}                        &\simet\ & 0.324  \\
              \hline \hline 

  \end{tabular}

    \caption{Quotients of mean squared error estimates for \simif, \simrsrfnf, \simrsrfaf, \simrf, \simet, divided by the mean squared error estimates when using \simeinsnn\ ($1$-nearest neighbor) as estimator.}
    \label{table:results_cv_normed_einsnn}
\end{table}

\begin{table}[t]
    \centering
    \begin{tabular}{r c c c | c c c}
         Model  & \multicolumn{3}{c |}{\modelPureType} & \multicolumn{3}{c}{\modelHierarchical}\\
         $d$ &  $4$  &  $10$ &  $30$ &  $4$  &  $10$ &  $30$ \\
         \hline
         First     & \simif     & \simif        & \simif      &  \simet        &  \simet    &  \simet \\
        Second     & \simrsrfaf  & \simrsrfnf    &  \simrsrfnf &   \simif       &   \simif   &  \simif \\ 
         Third     & \simrsrfnf  &  \simrsrfaf   &  \simrsrfaf &    \simrsrfaf  &  \simrsrfnf    &  \simrf \\
         Fourth    & \simet      &  \simet       &  \simet     &   \simrf       &  \simrsrfaf    & \simrsrfnf \\
         Fifth     & \simrf      &  \simrf       &  \simrf     &   \simrsrfnf   &  \simrf   &  \simrsrfaf
    \end{tabular}
\\[0.5cm]\centering
      \begin{tabular}{r c c c|c c c}
         Model  & \multicolumn{3}{c|}{\modelAdditive} & \multicolumn{3}{c}{\modelPureZwei} \\
         $d$ &  $4$  &  $10$ &  $30$ &  $4$  &  $10$ &  $30$\\
         \hline
         First     & \simet     & \simet      & \simet     &  \simif     &  \simif   &  \simif  \\
        Second     & \simif & \simif    &  \simif &   \simet       &   \simet   &  \simet \\ 
         Third     & \simrf  &  \simrf   &  \simrf &    \simrsrfnf  &  \simrsrfnf    &  \simrsrfnf  \\
         Fourth    & \simrsrfaf      &  \simrsrfnf       &  \simrsrfnf     &   \simrsrfaf       &  \simrsrfaf    & \simrsrfaf \\
         Fifth     & \simrsrfnf      &  \simrsrfaf       &  \simrsrfaf     &   \simrf   &  \simrf   &  \simrf 
    \end{tabular}\\[0.5cm]
          \begin{tabular}{r||c}
         Model  & \multicolumn{1}{c}{\modelPureDrei}\\
         Dimension $d$ & $d=6$\\
         \hline
         First      & \simif \\
        Second     & \simrsrfaf\\ 
         Third    & \simrsrfnf \\
         Fourth   & \simet\\
         Fifth     & \simrf
    \end{tabular}
    \caption{Rankings for (opt).}
    \label{tab:results_ranked}
\end{table}

\subsection{Optimal parameters chosen}\label{subsec:parameter_settings}
\Cref{tab:rsrf_af_opt,tab:rsrf_nf_opt,tab:if_opt,tab:et_opt,tab:rf_opt} contain the optimal parameters found during the parameter search.
\begin{table}[hbt]
    \centering
    \begin{tabular}{r|c c c|c c c}
        \multirow{2}{*}{ \backslashbox{Parameter}{Model} }    &   \multicolumn{3}{c|}{\modelPureType}  & \multicolumn{3}{c}{\modelHierarchical}  \\
         &  $d=4$  &  $d=10$ &  $d=30$ &  $d=4$  &  $d=10$ &  $d=30$  \\
         \hline
         \simcartcart      & FALSE & TRUE & TRUE & FALSE & TRUE & TRUE   \\
         \simreplace       & TRUE & TRUE & TRUE & FALSE & FALSE & TRUE   \\
         \simwidth         &  15  & 15 & 30  & 12 & 14 & 29 \\
         \simmtrycartcart  &  -  &  6 & 22 & - & 7 & 23     \\
         \simmtryrc        & 3 & 9 & 30 & 2 & 10 & 26 \\
         \simminnodesize   &  16 & 10 & 5 & 5 & 12 & 15 \\
    \end{tabular}
\\[0.5cm] 
    \centering
    \begin{tabular}{r|c c c|c c c}
         \multirow{2}{*}{ \backslashbox{Parameter}{Model} }    &   \multicolumn{3}{c|}{\modelAdditive}  & \multicolumn{3}{c}{\modelPureZwei}  \\
         &  $d=4$  &  $d=10$ &  $d=30$ &  $d=4$  &  $d=10$ &  $d=30$  \\
        \hline
         \simcartcart     & FALSE & TRUE & TRUE & FALSE & FALSE & FALSE     \\
         \simreplace      & TRUE & TRUE & TRUE & TRUE & TRUE & FALSE    \\
         \simwidth        & 12 & 15 & 16 & 13 & 15 & 25   \\
         \simmtrycartcart & - & 2 & 24 & - & - & -   \\
         \simmtryrc       & 2 & 8 & 24 & 4 & 10 & 30     \\
         \simminnodesize  & 14 & 11 & 8 & 23 & 13 & 22   \\
    \end{tabular}\\[0.5cm] \centering
        \begin{tabular}{r|c}
         \multirow{2}{*}{ \backslashbox{Parameter}{Model} }    &   \multicolumn{1}{c}{\modelPureDrei}   \\
         &  $d=6$   \\
         \hline
         \simcartcart     &  FALSE \\
         \simreplace      & TRUE \\
         \simwidth        & 9 \\
         \simmtrycartcart  & -   \\
         \simmtryrc       &  4    \\
         \simminnodesize  & 5   \\
    \end{tabular}
    
    \caption{Parameters used for \simrsrfnf (opt).}
    \label{tab:rsrf_nf_opt}
\end{table}

\begin{table}[hbt]
    \centering
    \begin{tabular}{r| c c c|c c c}
         \multirow{2}{*}{ \backslashbox{Parameter}{Model} }    &   \multicolumn{3}{c|}{\modelPureType}  & \multicolumn{3}{c}{\modelHierarchical}  \\
         &  $d=4$  &  $d=10$ &  $d=30$ &  $d=4$  &  $d=10$ &  $d=30$  \\
         \hline
         \simcartcart     & FALSE & TRUE  & TRUE  & FALSE & TRUE &  TRUE    \\
         \simreplace      & TRUE  & TRUE  & FALSE & TRUE  & TRUE &  FALSE   \\
         \simwidth        & 8     & 12    & 28    & 11    & 14   &  24   \\
         \simmtryrandom   & 3     & 9     & 26    & 4     & 8    &  19  \\
         \simmtryrc       & 4     & 8     & 26    & 3     & 9    &  30     \\
         \simminnodesize  & 14    & 5     & 6     & 10    & 11   &  17   \\
    \end{tabular}
\\[0.5cm]    \centering
    \begin{tabular}{r| c c c |c c c}
         \multirow{2}{*}{ \backslashbox{Parameter}{Model} }    &   \multicolumn{3}{c|}{\modelAdditive}  & \multicolumn{3}{c}{\modelPureZwei}  \\
         &  $d=4$  &  $d=10$ &  $d=30$ &  $d=4$  &  $d=10$ &  $d=30$  \\
         \hline
         \simcartcart     & FALSE & FALSE & TRUE & FALSE & FALSE & FALSE \\
         \simreplace      & TRUE  & TRUE  & TRUE & TRUE  & FALSE & TRUE  \\
         \simwidth        & 14    & 12    & 12   & 3     & 13    & 24   \\
         \simmtryrandom   & 4     & 9     & 22   & 4     & 8     & 24     \\
         \simmtryrc       & 2     & 10    & 26   & 2     & 10    & 28     \\
         \simminnodesize  & 12    &  7    & 30   & 20    & 13    & 29   \\
    \end{tabular}
    \\[0.5cm] \centering
        \begin{tabular}{r|c}
         \multirow{2}{*}{ \backslashbox{Parameter}{Model} }    &   \multicolumn{1}{c}{\modelPureDrei}   \\
         &  $d=6$   \\
         \hline
         \simcartcart     &  FALSE \\
         \simreplace      & FALSE \\
         \simwidth        &  15 \\
         \simmtryrandom   &  5 \\
         \simmtryrc       &  4    \\
         \simminnodesize  &  9  \\
    \end{tabular}
    
    \caption{Parameters used for \simrsrfaf (opt).}
    \label{tab:rsrf_af_opt}
\end{table}

\begin{table}[hbt]
    \centering
    \begin{tabular}{r| c c c|c c c}
         \multirow{2}{*}{ \backslashbox{Parameter}{Model} }    &   \multicolumn{3}{c|}{\modelPureType}  & \multicolumn{3}{c}{\modelHierarchical}  \\
         &  $d=4$  &  $d=10$ &  $d=30$ &  $d=4$  &  $d=10$ &  $d=30$  \\
         \hline
         \texttt{npairs}     &  14     & 153    & 749   & 7      &  110  & 450 \\
         \texttt{replace}      & TRUE  & FALSE  & FALSE &  FALSE & FALSE & FALSE \\
         \texttt{min.node.size}  &  20 &   11   &  11   &  10    &  8    &  17   \\
    \end{tabular}
    \\[0.5cm]    \centering
    \begin{tabular}{r|c c c|c c c}
         \multirow{2}{*}{ \backslashbox{Parameter}{Model} }    &   \multicolumn{3}{c|}{\modelAdditive}  & \multicolumn{3}{c}{\modelPureZwei}  \\
         &  $d=4$  &  $d=10$ &  $d=30$ &  $d=4$  &  $d=10$ &  $d=30$  \\
         \hline
         \texttt{npairs}         & 23    &  33     &  99       &     2  &  151  &  30  \\
         \texttt{replace}        & TRUE  &  FALSE  &  FALSE    & FALSE  & FALSE & FALSE  \\
         \texttt{min.node.size}  & 13    &  14     &  18       &  16    &  26   &  28  \\
    \end{tabular}    \\[0.5cm]    \centering
      \begin{tabular}{r | c}
         \multirow{2}{*}{ \backslashbox{Parameter}{Model} }    &   \multicolumn{1}{c}{\modelPureDrei}    \\
         &  $d=6$  \\
         \hline
         \texttt{npairs}     & 99  \\
         \texttt{replace}        & TRUE     \\
         \texttt{min.node.size}  & 22    \\
    \end{tabular}
    \caption{Parameters used for \simif (opt).}
    \label{tab:if_opt}
\end{table}

\begin{table}[hbt]
    \centering
    \begin{tabular}{r| c c c|c c c}
         \multirow{2}{*}{ \backslashbox{Parameter}{Model} }    &   \multicolumn{3}{c|}{\modelPureType}  & \multicolumn{3}{c}{\modelHierarchical}  \\
         &  $d=4$  &  $d=10$ &  $d=30$ &  $d=4$  &  $d=10$ &  $d=30$  \\
         \hline
         \texttt{mtry}           & 4 & 10 & 30 & 3 & 6 & 9 \\
         \texttt{replace}      & TRUE  & FALSE   & FALSE  & TRUE   &   TRUE & TRUE   \\
         \texttt{min.node.size}  &  5   & 5     & 7   & 8    &6     & 12    \\
    \end{tabular}
    \\[0.5cm]    \centering
    \begin{tabular}{r|c c c|c c c}
         \multirow{2}{*}{ \backslashbox{Parameter}{Model} }    &   \multicolumn{3}{c|}{\modelAdditive}  & \multicolumn{3}{c}{\modelPureZwei}  \\
         &  $d=4$  &  $d=10$ &  $d=30$ &  $d=4$  &  $d=10$ &  $d=30$  \\
         \hline
         \texttt{mtry}     & 2 & 7 & 26 & 2 &5  & 20 \\
         \texttt{replace}        & TRUE   & TRUE & TRUE  &  TRUE & TRUE  & TRUE   \\
         \texttt{min.node.size}  & 5    &  15    &  18  &   10  &   8  &  30  \\
    \end{tabular}\\[0.5cm]    \centering
    \begin{tabular}{r|c}
         \multirow{2}{*}{ \backslashbox{Parameter}{Model} }    &   \multicolumn{1}{c}{\modelPureDrei}    \\
         &  $d=6$  \\
         \hline
         \texttt{mtry}     & 5  \\
         \texttt{replace}        & TRUE     \\
         \texttt{min.node.size}  & 6    \\
    \end{tabular}
    \caption{Parameters used for \simrf (opt).}
    \label{tab:rf_opt}
\end{table}

\begin{table}[b]
    \centering
    \begin{tabular}{r|c c c|c c c}
         \multirow{2}{*}{ \backslashbox{Parameter}{Model} }    &   \multicolumn{3}{c|}{\modelPureType}  & \multicolumn{3}{c}{\modelHierarchical}  \\
         &  $d=4$  &  $d=10$ &  $d=30$ &  $d=4$  &  $d=10$ &  $d=30$  \\
         \hline
         \texttt{mtry}              &   4    & 9       & 29    & 3    & 9        & 29 \\
        \texttt{num.random.splits}  &  3     &       3 &     6 &      3 &    3   & 9 \\
         \texttt{replace}           &  FALSE & FALSE   & FALSE & FALSE  & FALSE  & FALSE   \\
         \texttt{min.node.size}     & 12     &  5      &   5   &   8    &   5    &   9 \\
    \end{tabular}
    \\[0.5cm]    \centering
    \begin{tabular}{r|c c c|c c c}
         \multirow{2}{*}{ \backslashbox{Parameter}{Model} }    &   \multicolumn{3}{c|}{\modelAdditive}  & \multicolumn{3}{c}{\modelPureZwei}  \\
         &  $d=4$  &  $d=10$ &  $d=30$ &  $d=4$  &  $d=10$ &  $d=30$  \\
         \hline
         \texttt{mtry}     & 3 & 7 & 29 & 2 & 7 & 28 \\
         \texttt{num.random.splits}   & 5 & 3 & 3 & 1 & 1 & 1 \\
         \texttt{replace}      &   TRUE&FALSE   & FALSE &  FALSE & TRUE  & TRUE   \\
         \texttt{min.node.size}  &   6  &  10    &  16  & 10    & 6    & 15   \\
    \end{tabular}
    \\[0.5cm] \centering
       \begin{tabular}{r | c}
         \multirow{2}{*}{ \backslashbox{Parameter}{Model} }    &   \multicolumn{1}{c}{\modelPureDrei}   \\
         &  $d=6$    \\
         \hline
         \texttt{mtry}     & 1  \\
         \texttt{num.random.splits}   &  5  \\
         \texttt{replace}      & FALSE  \\
         \texttt{min.node.size}  &   5  \\
    \end{tabular}
    \caption{Parameters used for \simet (opt).}
    \label{tab:et_opt}
\end{table}

\subsection{Some remarks on the parameters in \rsrf}\label{subsec:remark_parameters}
Below, we include some remarks on the parameter choices for \rsrf. Nonetheless, we want to point out that this discussion should be considered as heuristic and a deeper analysis on how to choose the hyper-parameters is beyond the scope of this paper. From our simulations we see that best results were usually obtained for large choices of the width parameter $\width$. However, a closer look at the tables in the supplementary material reveals that, for the pure interaction models, \rsrf\ improves upon Random Forests also for small values of $\width$. For instance, in \modelPureDrei, choosing a small $\width = 3$ for $\simrsrfaf$ already achieved an MSE of $\approx 0.25$ whereas the error for \simrf\ is larger than $0.5$ (see \Cref{fig:plotF_new}). Regarding the mtry-parameters in \rsrf, we note that, in order to reduce the number of tuning parameters, one could instead consider a single mtry parameter by setting $\mtrycartcart = \mtryrandomcart$ for $\simrsrfnf$, and similarly for $\simrsrfaf$. We did so in the applications on the real dataset.\\
Motivated from our simulations, we do not believe that the node size is particularly important and suggest it to choose rather small, e.g. $10$. Lastly, let us briefly discuss the parameter $\cartcart$. We observe that for large $d=30$ in models $\modelHierarchical$, $\modelAdditive$, it is advantageous to set this parameter to \quot{true} (this was the case in almost all of the top $20$ parameter setups for both $\simrsrfnf$, $\simrsrfaf$). Contrary, in the pure model $\modelPureZwei$ for $\simrsrfaf$, it was set to \quot{false} in $15$ out of the top $20$ setups (and in $19$ out of the top $20$ settings in $\modelPureType$ with $d=30$). We note that the choice for $\cartcart$ should also be connected to the width parameter $\width$. The larger the width, the less influential is $\cartcart$. 

\section{Appendix to the illustrations on real data examples}\label{sec:appendix_realdata}
In this section, we present the parameter choices used for the real datasets. Furthermore, we complement \Cref{sec:realdata} by presenting additional results.
\subsection{Parameter choices}\label{subsec:parametrs_realdata}

\Cref{tab:parameters_realdata} and \Cref{tab:parameters_realdata_noisecov} show the parameters used when applying the different algorithms on the real datasets (see \Cref{sec:realdata} in the main text).
\begin{remark}\label{remark:npairs_not_scale}
    For the sake of clarity, we shortly elaborate on why we used a larger range for \texttt{npairs} for Interaction Forests in the \calhousing\ example. This dataset has a much larger sample size than the other ones. Note that for a fixed \texttt{mtry} parameter in Random Forests, the number of candidate splits grows proportionally with the number of observations in the current node in a tree. Contrary, this is not the case in Interaction Forests. Here, for a fixed \texttt{npairs} parameter, the number of candidate splits does not change with the number of observations in the current node. 
\end{remark}
\Cref{tab:chosen_parameters} summarizes the parameters that were chosen during tuning, i.e. within the inner cross-validation loops, exemplarily for the two datasets \abalone\ and \robot\ and the algorithms from the main text. In most of the cases, these parameters do not vary strongly, which one can also observe for the other datasets. For \simif, we see that in some cases, both larger and smaller values for \texttt{npairs} are chosen, e.g. in the abalone dataset.

\begin{table}[h]
    \centering
    \begin{tabular}{c l l}
        Algorithm & Parameter & Value / Range \\
         \hline  
        \simrsrfnfRealdata 
                   & \simwidth        & $30$ \\
                   & \simmtrycartcart $=$ \simmtryrc & $1,2,\dots, d$ \\
                   & \simminnodesize  & $5$ \\
                   & \simreplace       & True \\
                   & \simntrees       & $100$ \\
                   &\simcartcart   & True \\
                   & \mtrymode        & \mtrymodenf \\ 
         \hline 
            \simrfRealdata & \texttt{num.trees} & $500$ \\
                   & \texttt{min.node.size}     & $5$ \\
                   & \texttt{replace}           & True \\
                   & \texttt{mtry}              & $1,2,\dots ,d$ \\
         \hline
           \simifRealdata &  \texttt{num.trees} & $500$ \\     
                  & \texttt{min.node.size} & $5$ \\
                   & \texttt{replace} & True \\
                   & \texttt{npairs} & $1,10,25,50\dots,250$ \\
                   & &(additionally $300,400,\dots 700$\\
                   & & for \calhousing) \\
         \hline
            \simetRealdata &  \texttt{num.trees} & $500$ \\     
                  & \texttt{min.node.size} & $5$ \\
                   & \texttt{replace} & True \\
                    & \texttt{num.random.splits}    & $1,2,\dots ,10$ \\
                    & \texttt{mtry}    & $1,2,\dots ,d$ \\
                   & \texttt{splitrule} & \texttt{extratrees}
    \end{tabular}
    \caption{Parameter settings used for the real data examples. For \simetRealdata, any combination of \texttt{num.random.splits} and \texttt{mtry} was used. We note that, for \simifRealdata\ in the \calhousing\ dataset, we chose a larger range for \texttt{npairs} to allow for a larger number of candidate splits in each node, because the sample size for this dataset is much larger than the other sample sizes, see also \Cref{remark:npairs_not_scale}}
    \label{tab:parameters_realdata}
\end{table}

\begin{table}[h]
    \centering
    \begin{tabular}{c l l}
        Algorithm & Parameter & Value / Range \\
         \hline  
        \simrsrfnfRealdata 
                   & \simwidth        & $30$ \\
                   & &(resp. $45$ for \\
                   & & \calhousing\ (HD))\\
                   & \simmtrycartcart $=$ \simmtryrc & $10,20,\dots, 50,d$ \\
                   & &(resp. $10,30,50,d$ for \\
                   & & \calhousing\ (HD)) \\
                   & \simminnodesize  & $5$ \\
                   & \simreplace       & True \\
                   & \simntrees       & $100$ \\
                   &\simcartcart   & True \\
                   & \mtrymode        & \mtrymodenf \\ 
         \hline 
            \simrfRealdata & \texttt{num.trees} & $500$ \\
                   & \texttt{min.node.size}     & $5$ \\
                   & \texttt{replace}           & True \\
                   & \texttt{mtry}              & $5,10,\dots,55 ,d$ \\
         \hline
           \simifRealdata &  \texttt{num.trees} & $500$ \\     
                  & \texttt{min.node.size} & $5$ \\
                   & \texttt{replace} & True \\
                   & \texttt{npairs} & $1,10,25,50\dots,250,300,400,\dots,700$ \\
                   & &(resp. $50,100, 200,\dots, 1000$ \\
                   & & for \calhousing\ (HD)) \\
         \hline
            \simetRealdata &  \texttt{num.trees} & $500$ \\     
                  & \texttt{min.node.size} & $5$ \\
                   & \texttt{replace} & True \\
                    & \texttt{num.random.splits}    & $1,2,\dots ,10$ \\
                    & \texttt{mtry}    & $5,10,\dots ,55$ \\
                   & \texttt{splitrule} & \texttt{extratrees}
    \end{tabular}
    \caption{Parameter settings used for the real data examples with artificially added covariables. For \simetRealdata, any combination of \texttt{num.random.splits} and \texttt{mtry} was used. We note that, for \simifRealdata\ in the \calhousing\ dataset, we chose a larger range for \texttt{npairs} to allow for a larger number of candidate splits in each node, because the sample size for this dataset is much larger than the other sample sizes, see also \Cref{remark:npairs_not_scale}.}
    \label{tab:parameters_realdata_noisecov}
\end{table}

\subsection{Additional numerical results on real data}\label{subsec:realdata_additional_simulations}
\begin{figure}[t]
     \centering
     \begin{subfigure}[b]{0.32\textwidth}
            \centering
            \caption{\concrete}
         \includegraphics[width=\textwidth]{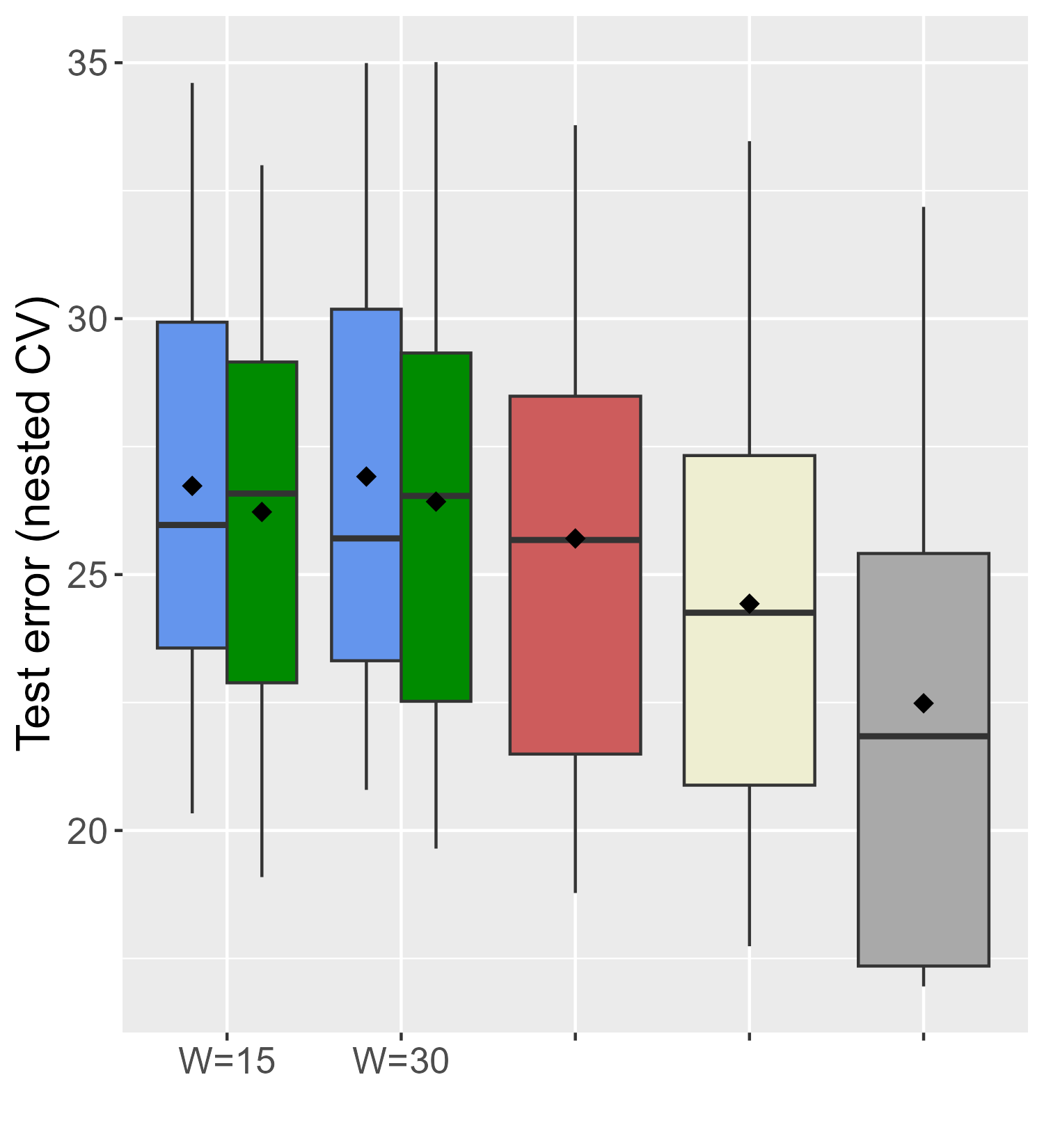}
     \end{subfigure}
     \hfill
     \begin{subfigure}[b]{0.32\textwidth}
         \centering
         \caption{\airfoil}
         \includegraphics[width=\textwidth]{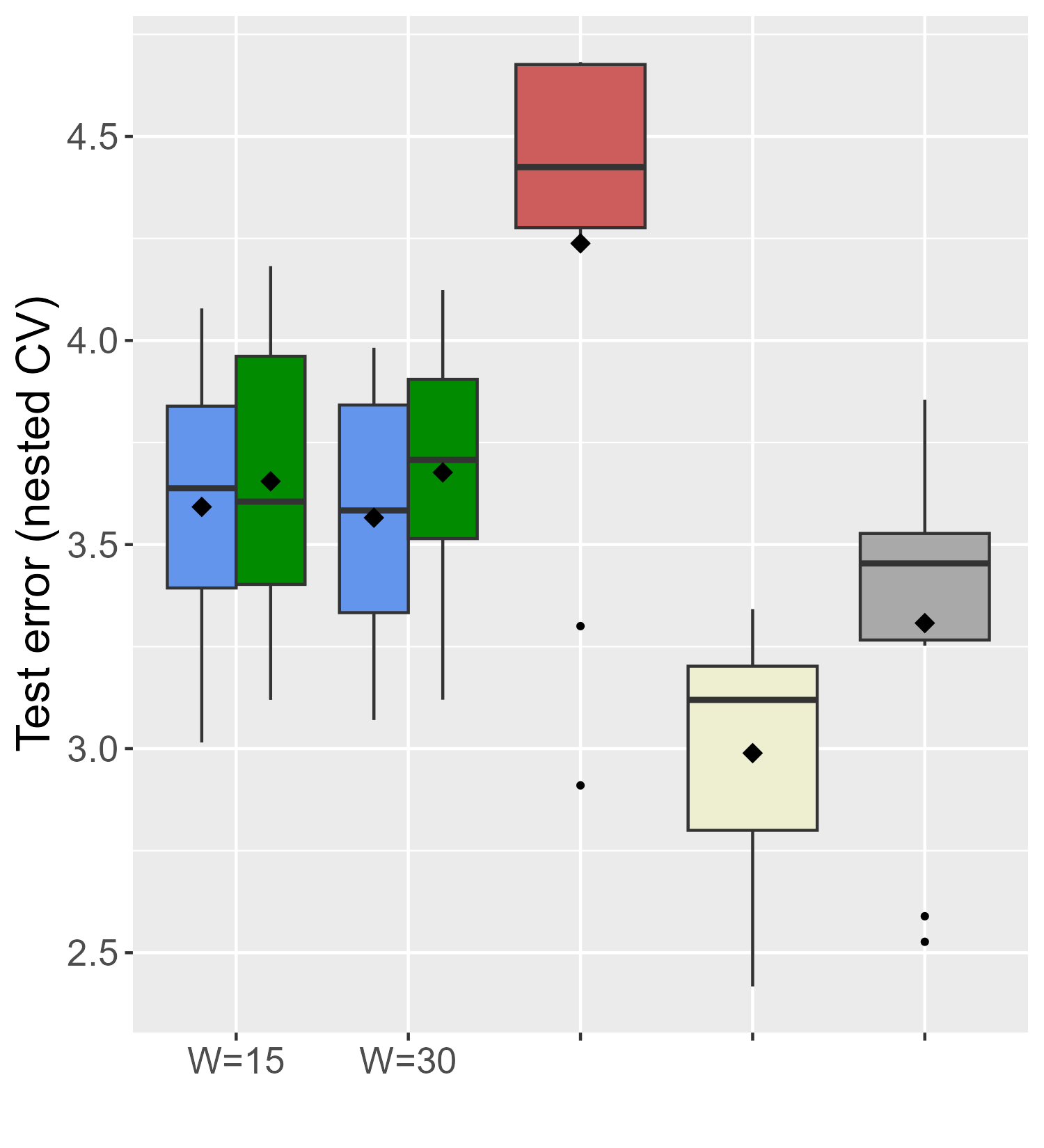}
     \end{subfigure}
     \begin{subfigure}[b]{0.32\textwidth}
         \centering
          \caption{\abalone}
         \includegraphics[width=\textwidth]{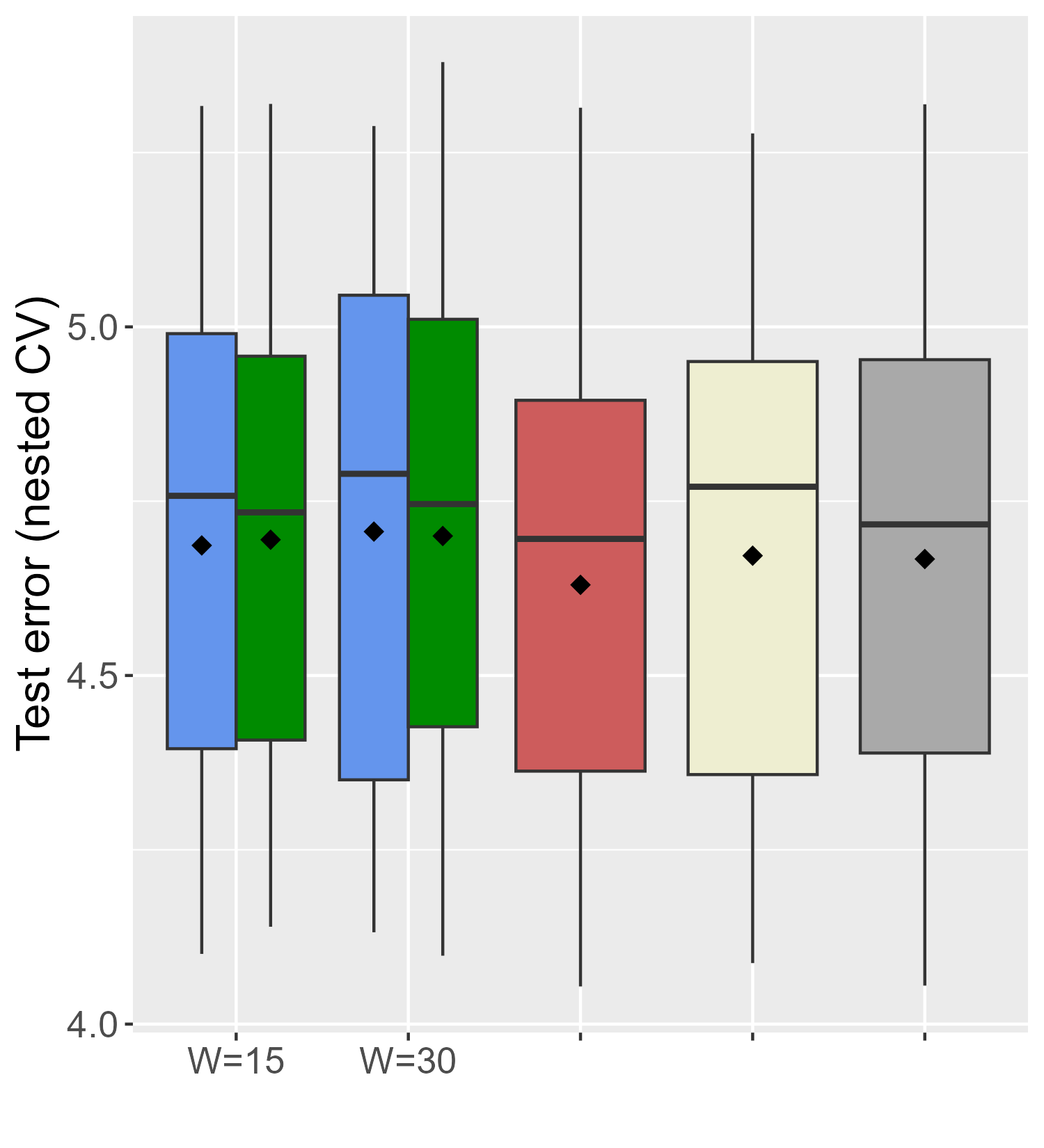}
     \end{subfigure}
       \hfill
    \begin{subfigure}[b]{0.32\textwidth}
         \centering
           \caption{\robot}
         \includegraphics[width=\textwidth]{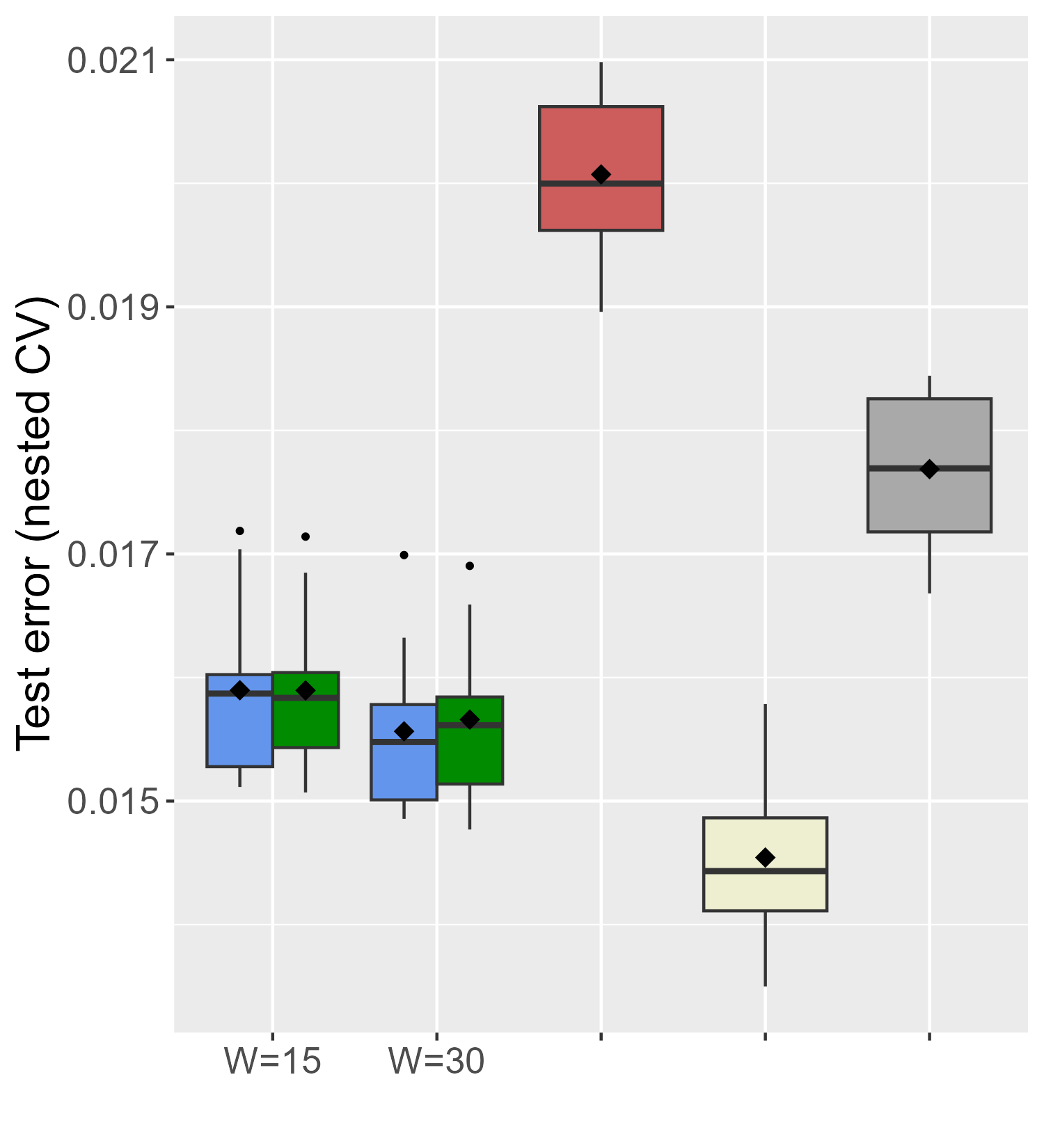}
     \end{subfigure} 
     \begin{subfigure}[b]{0.32\textwidth}
         \centering
           \caption{\calhousing}
         \includegraphics[width=\textwidth]{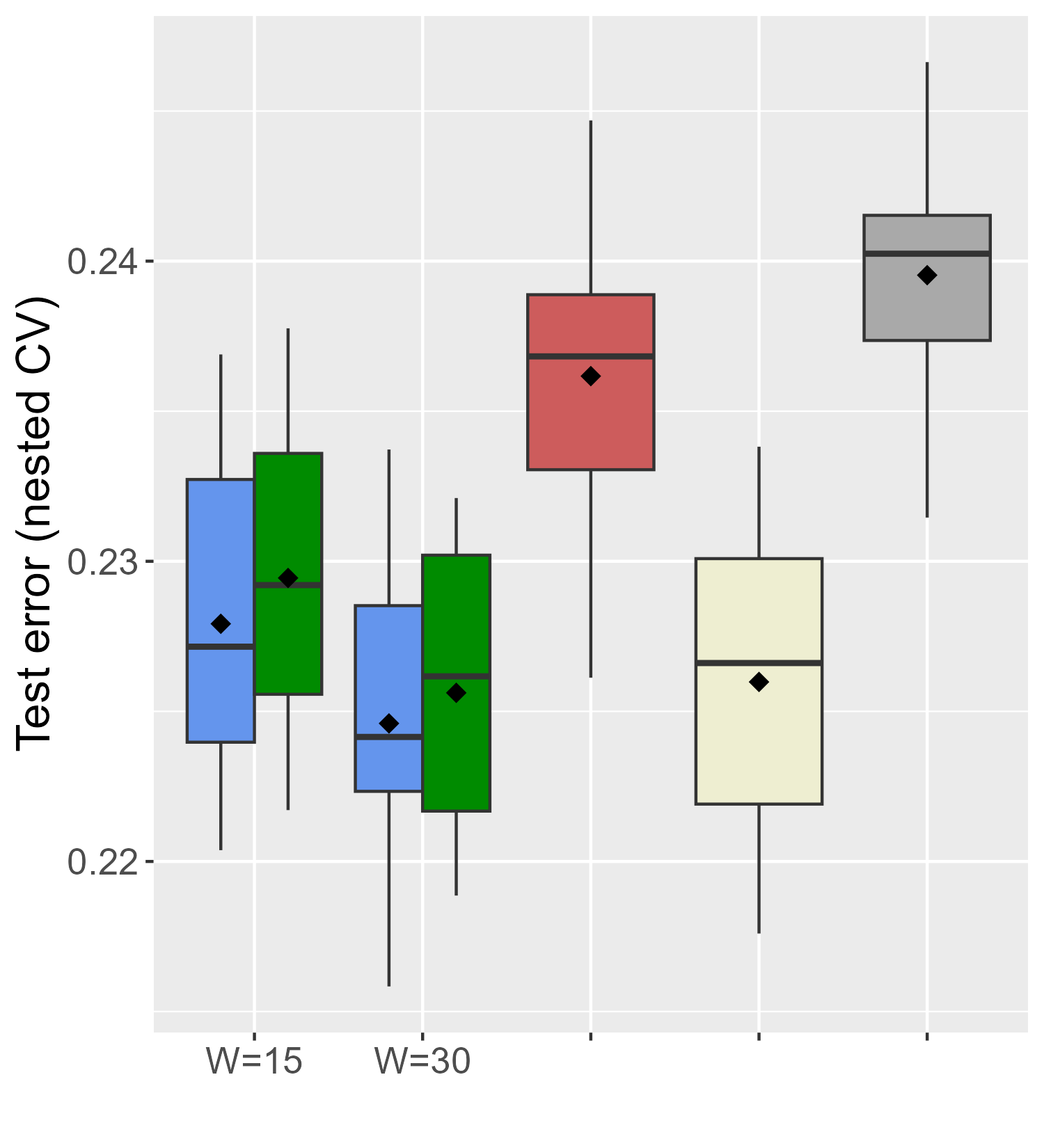}
     \end{subfigure} \hfill
       \begin{subfigure}[b]{0.32\textwidth}
         \centering
           \caption{Plot Legend}
         \includegraphics[width=\textwidth]{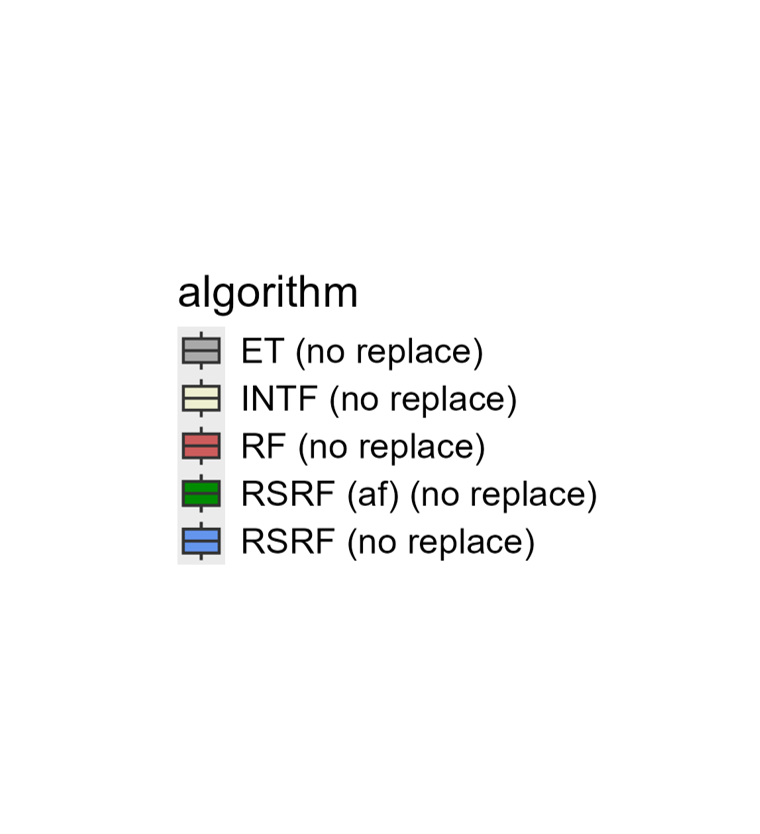}
     \end{subfigure}
        \caption{Additional results for the real data examples. The boxplots show squared errors from nested CV for the real datasets. The black diamonds represent the mean of the squared errors.}
        \label{fig:boxplots_additional}
\end{figure}

\begin{figure}[t]
     \centering
     \begin{subfigure}[b]{0.32\textwidth}
            \centering
            \caption{\concrete\ (HD)}
         \includegraphics[width=\textwidth]{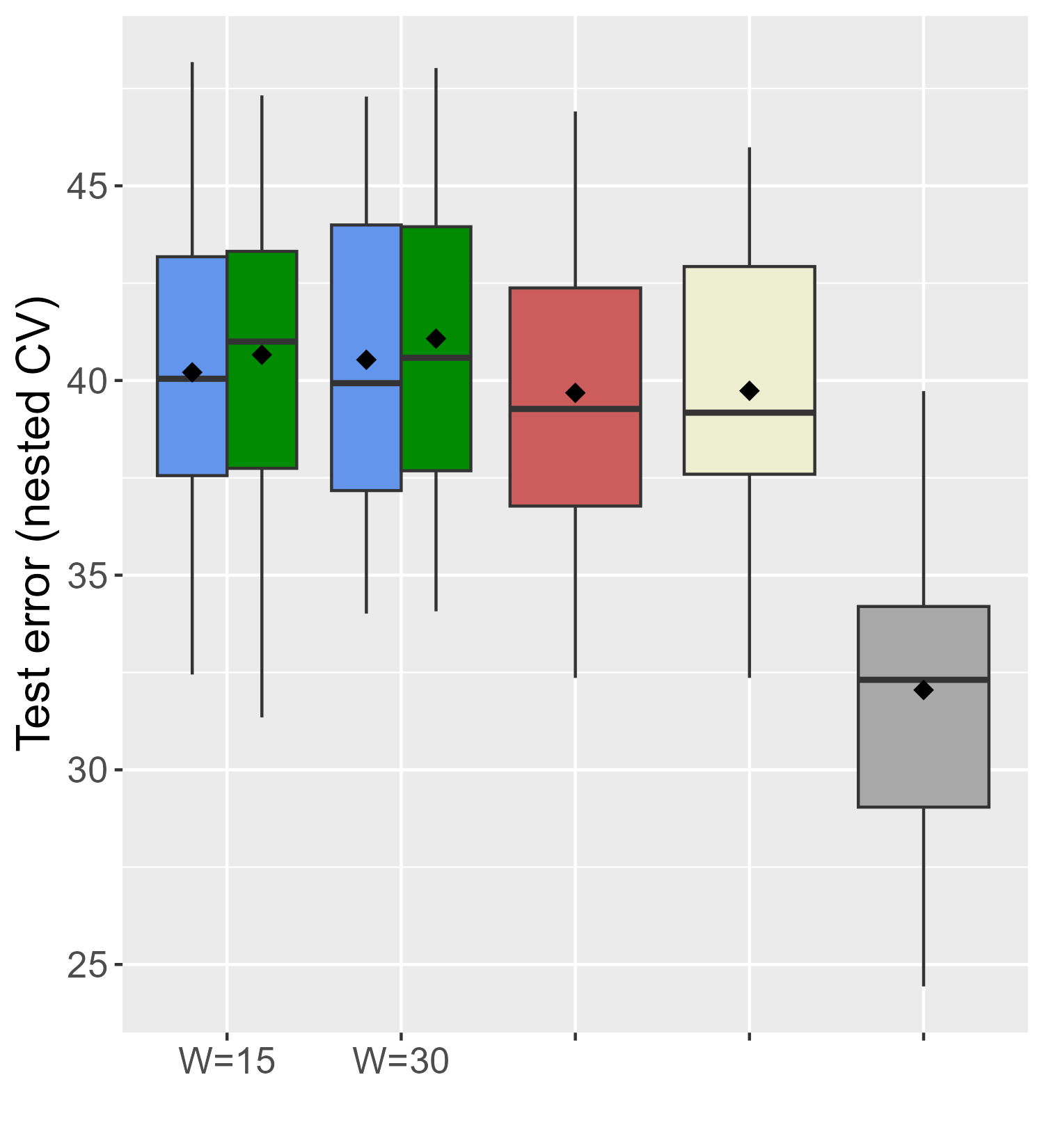}
     \end{subfigure}
     \hfill
     \begin{subfigure}[b]{0.32\textwidth}
         \centering
         \caption{\airfoil\ (HD)}
         \includegraphics[width=\textwidth]{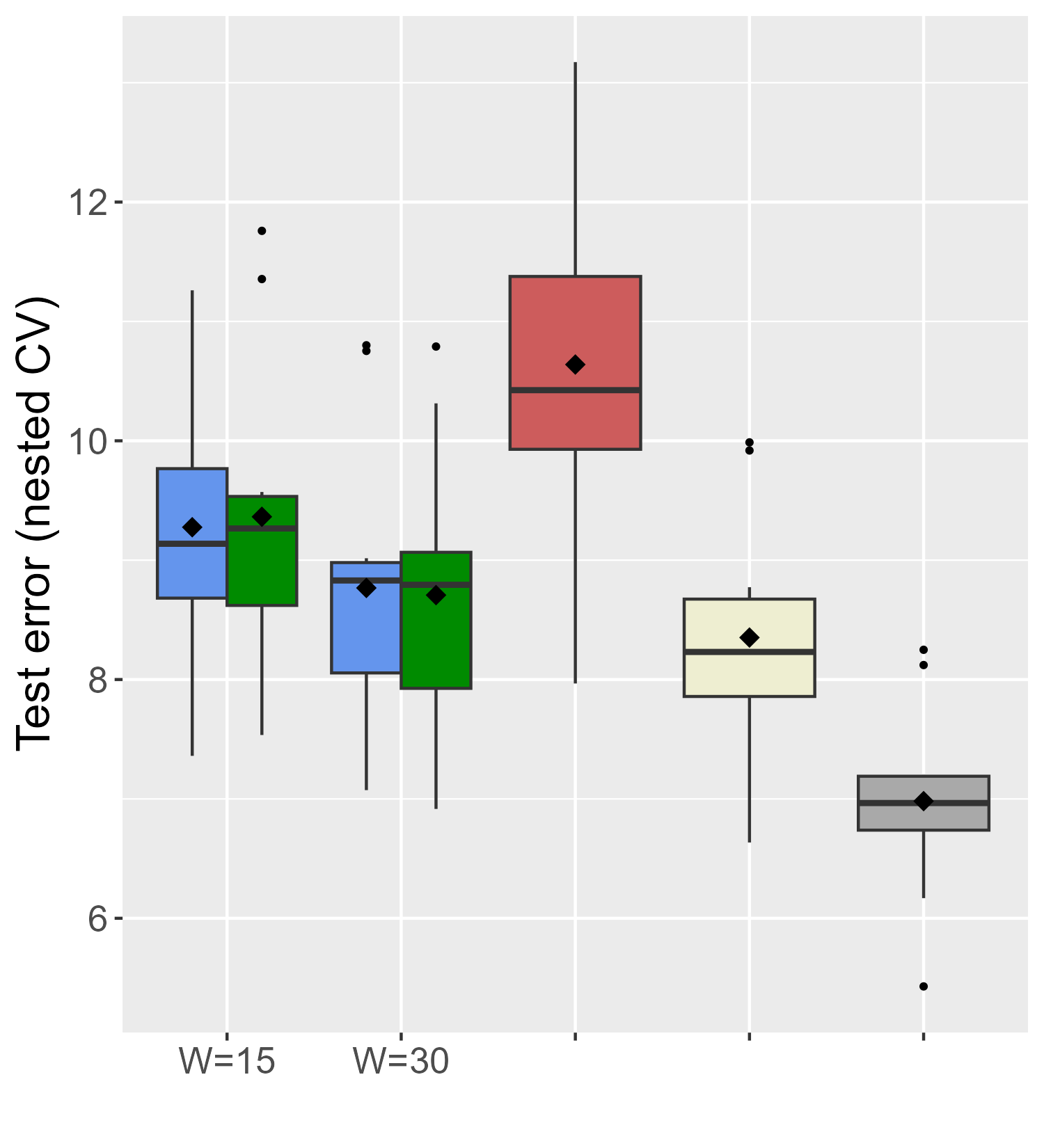}
     \end{subfigure}
     \begin{subfigure}[b]{0.32\textwidth}
         \centering
          \caption{\abalone\ (HD)}
         \includegraphics[width=\textwidth]{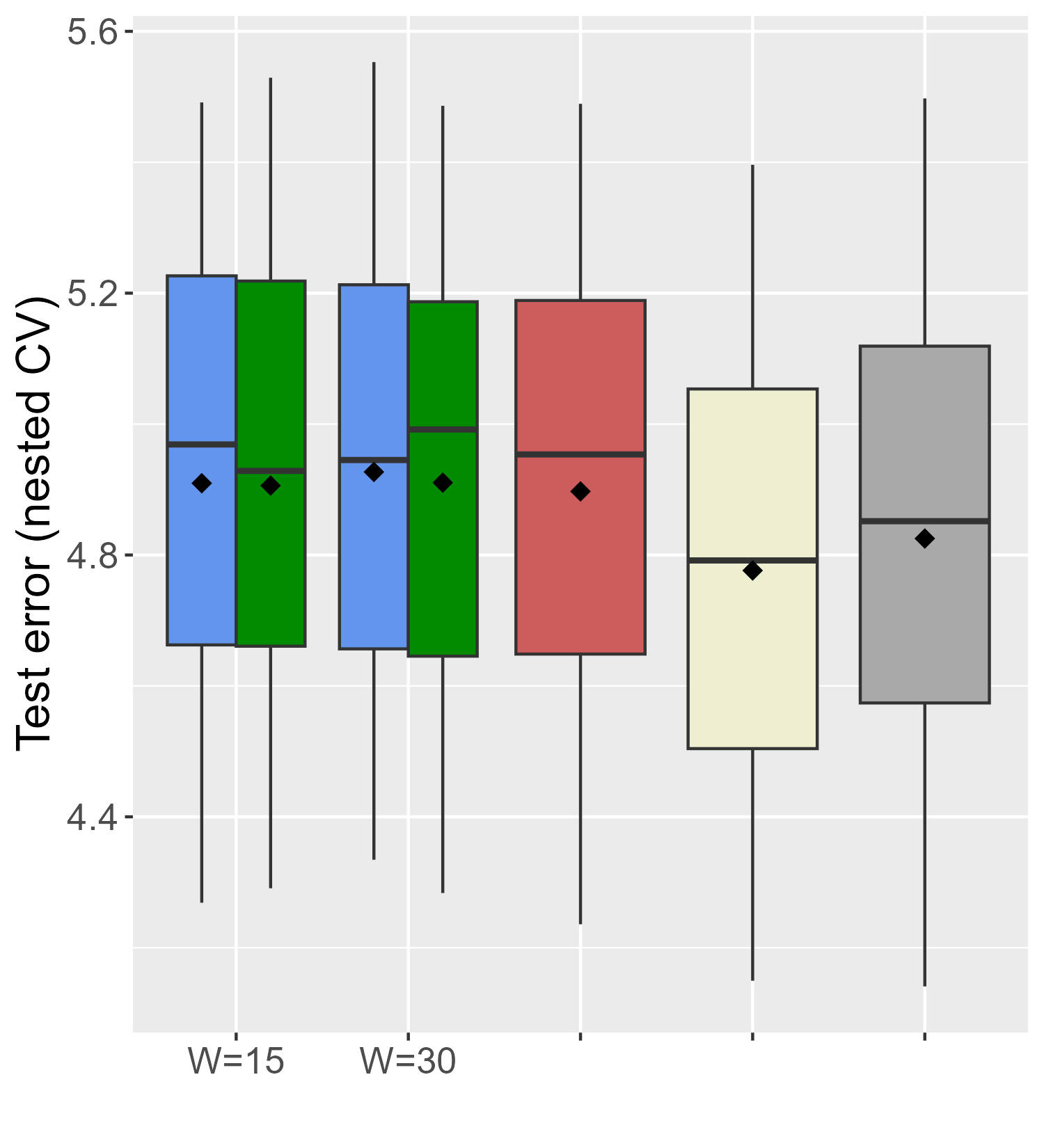}
     \end{subfigure}
       \hfill
    \begin{subfigure}[b]{0.32\textwidth}
         \centering
           \caption{\robot\ (HD)}
         \includegraphics[width=\textwidth]{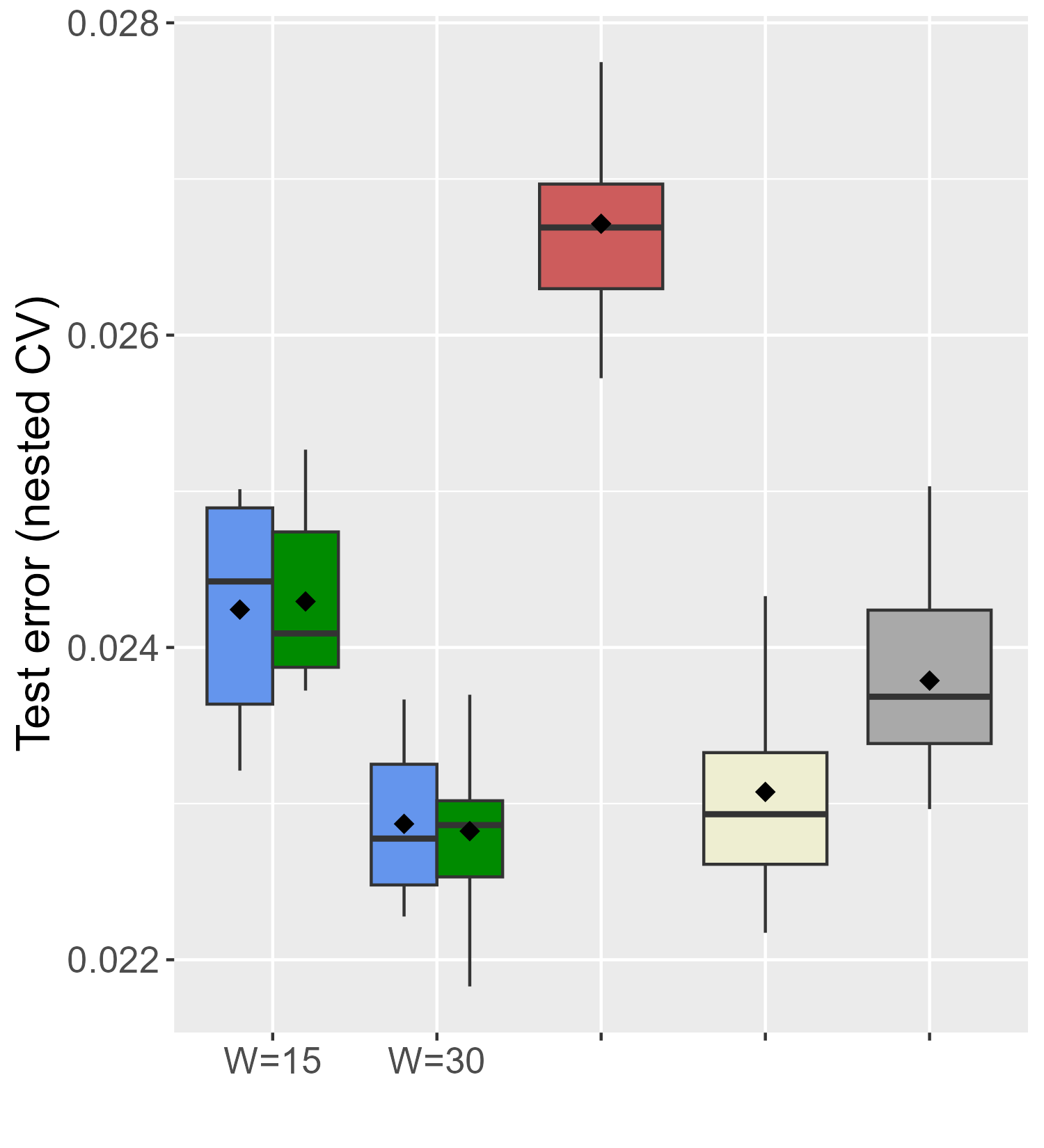}
     \end{subfigure} 
     \begin{subfigure}[b]{0.32\textwidth}
         \centering
           \caption{\calhousing\ (HD)}
         \includegraphics[width=\textwidth]{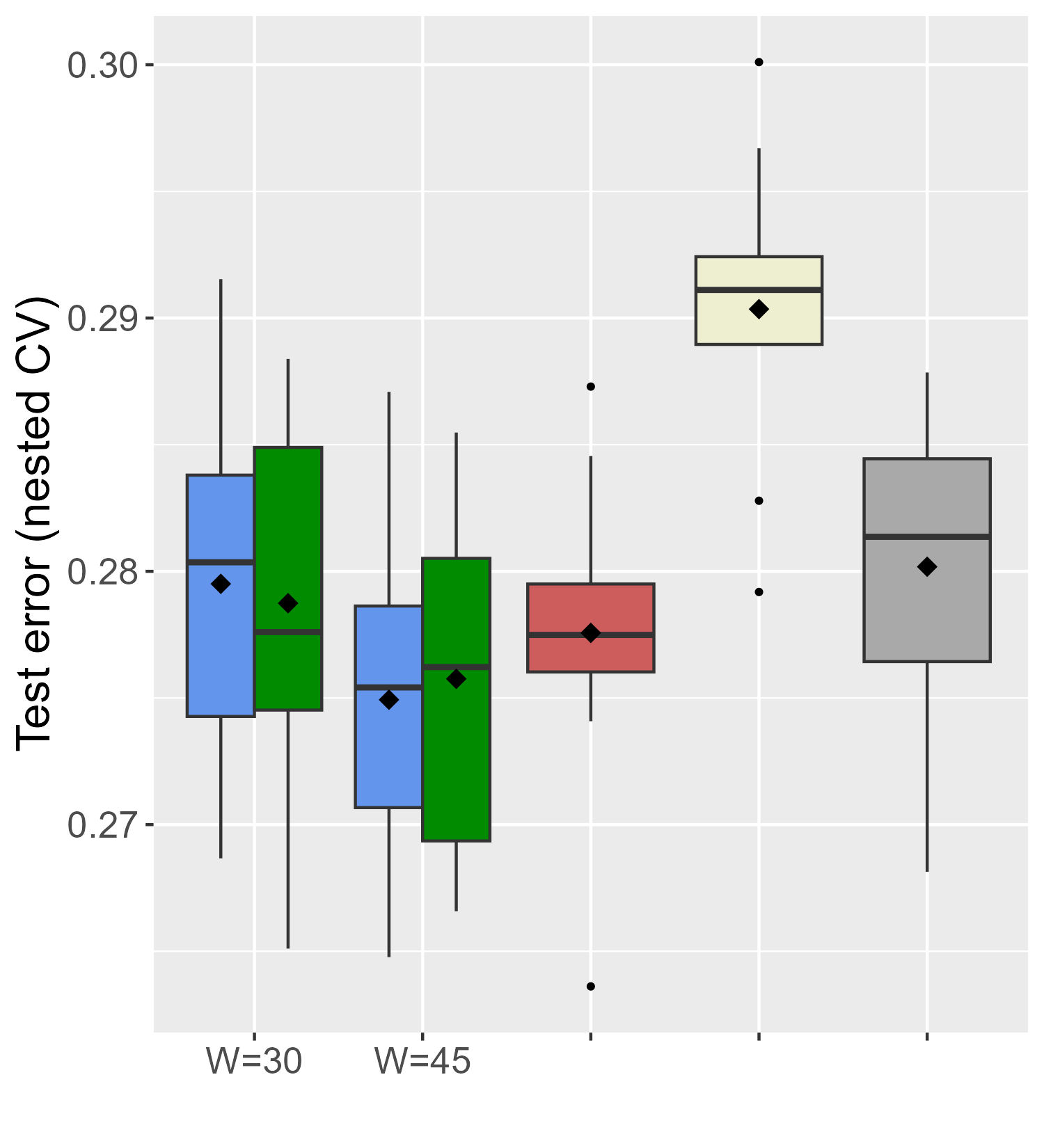}
     \end{subfigure} \hfill
       \begin{subfigure}[b]{0.32\textwidth}
         \centering
           \caption{Plot Legend}
         \includegraphics[width=\textwidth]{figures_dataexamples/legend_appendix_print.png}
     \end{subfigure}
        \caption{Additional results for the real data examples (HD). The boxplots show squared errors from nested CV for the real datasets. The black diamonds represent the mean of the squared errors.}
        \label{fig:boxplots_hd_additional}
\end{figure}

\begin{figure}[t]
     \centering
     \begin{subfigure}[b]{0.32\textwidth}
            \centering
            \caption{\concrete}
         \includegraphics[width=\textwidth]{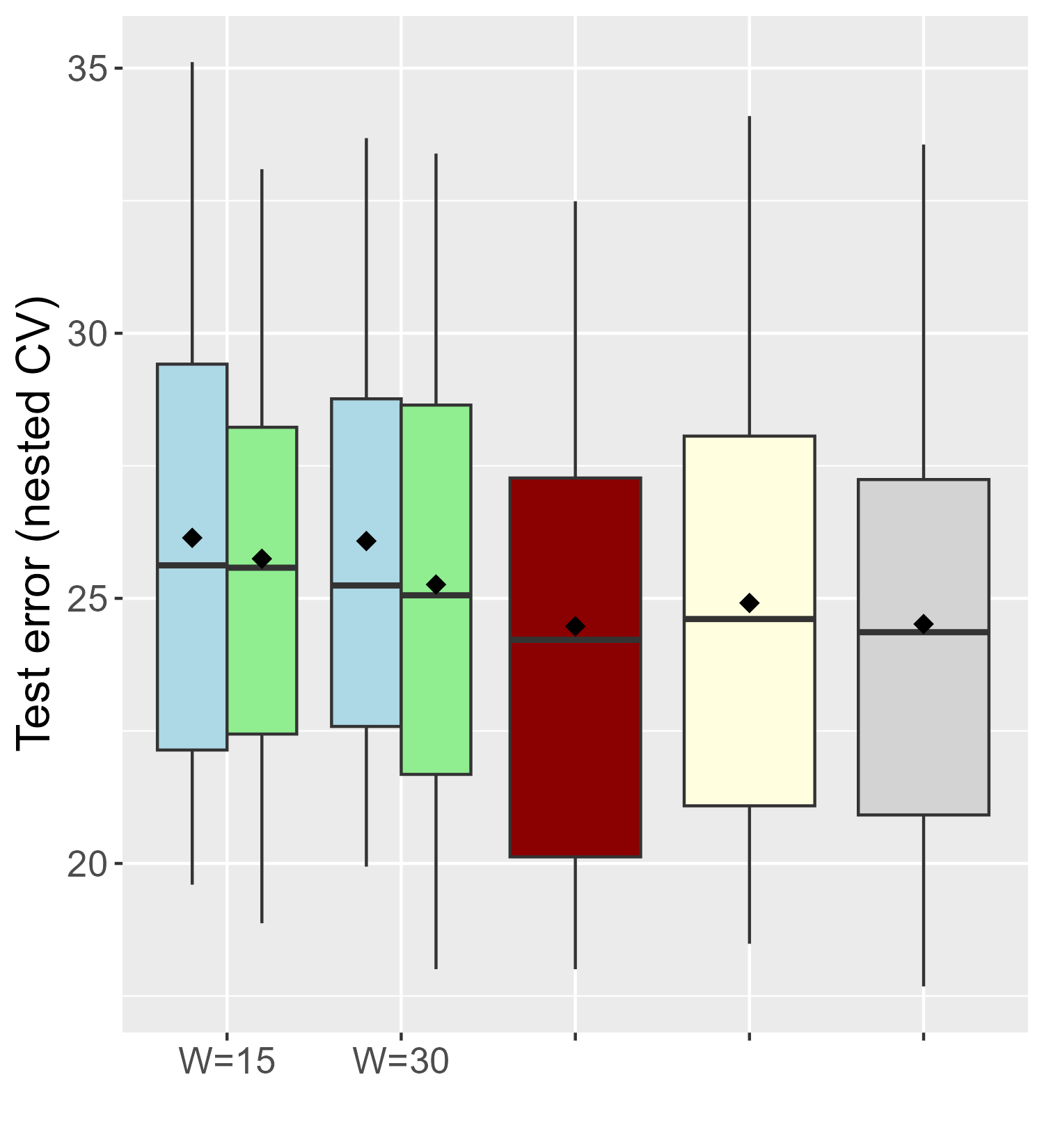}
     \end{subfigure}
     \hfill
     \begin{subfigure}[b]{0.32\textwidth}
         \centering
         \caption{\airfoil}
         \includegraphics[width=\textwidth]{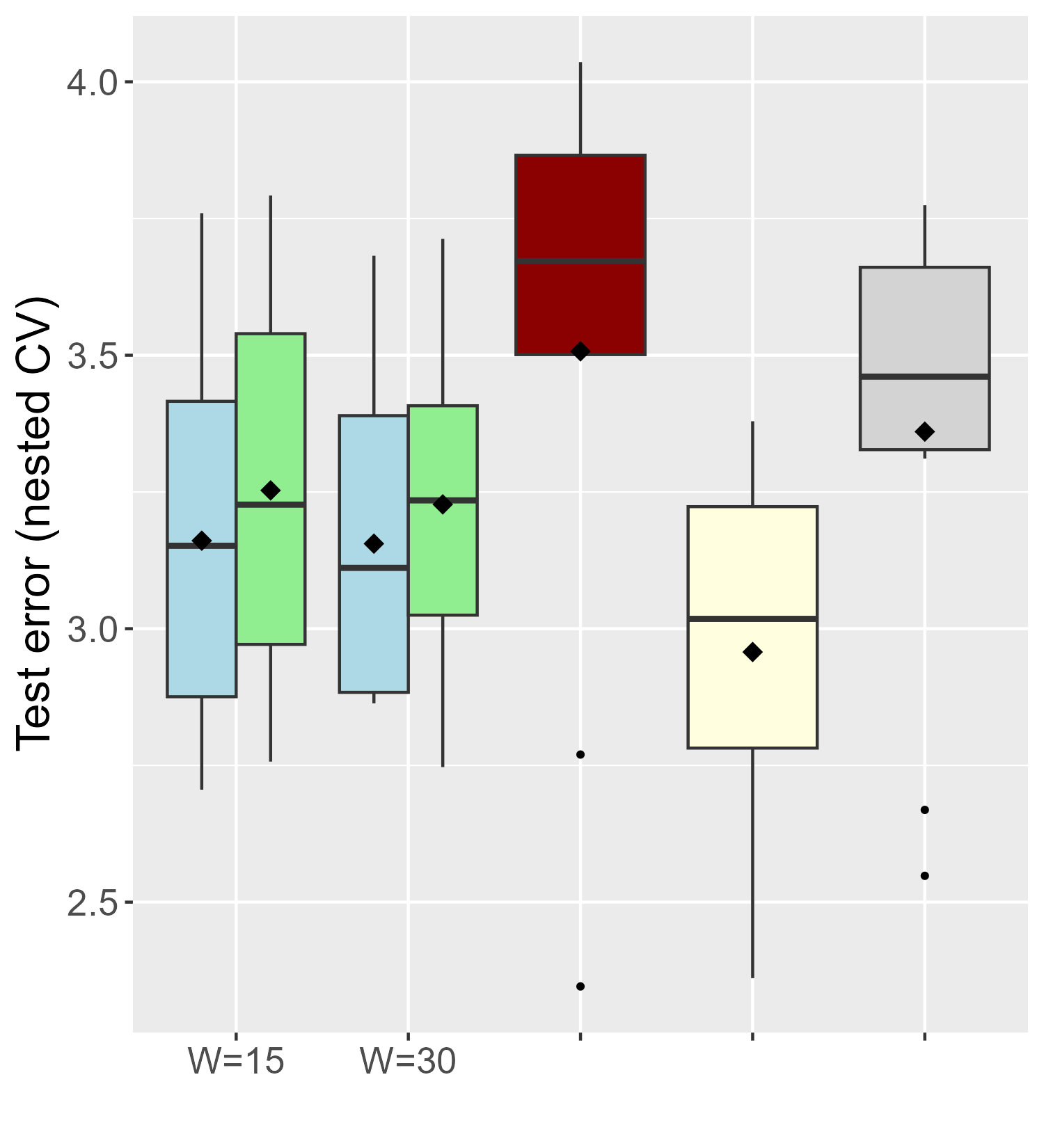}
     \end{subfigure}
     \begin{subfigure}[b]{0.32\textwidth}
         \centering
          \caption{\abalone}
         \includegraphics[width=\textwidth]{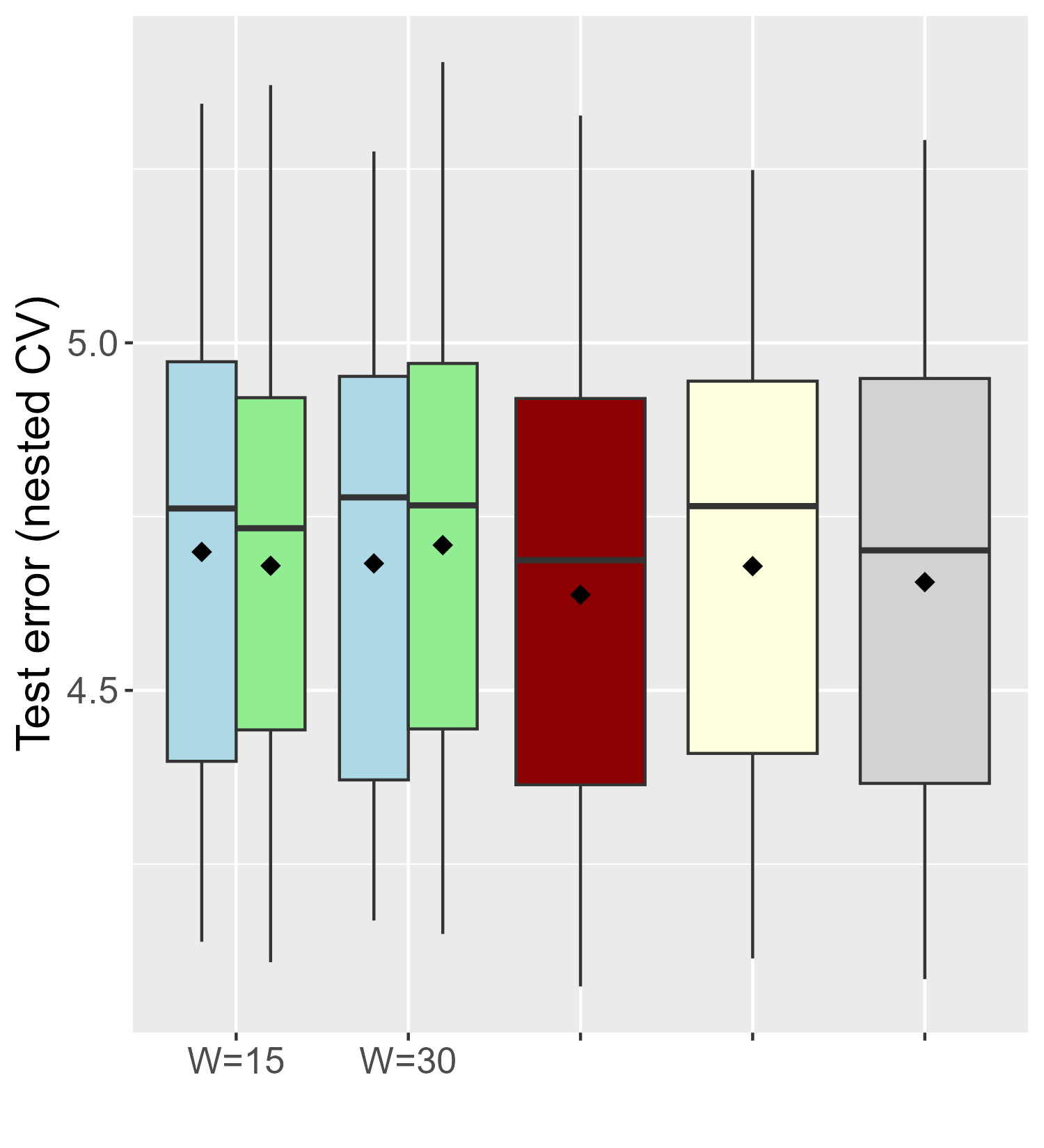}
     \end{subfigure}
       \hfill
    \begin{subfigure}[b]{0.32\textwidth}
         \centering
           \caption{\robot}
         \includegraphics[width=\textwidth]{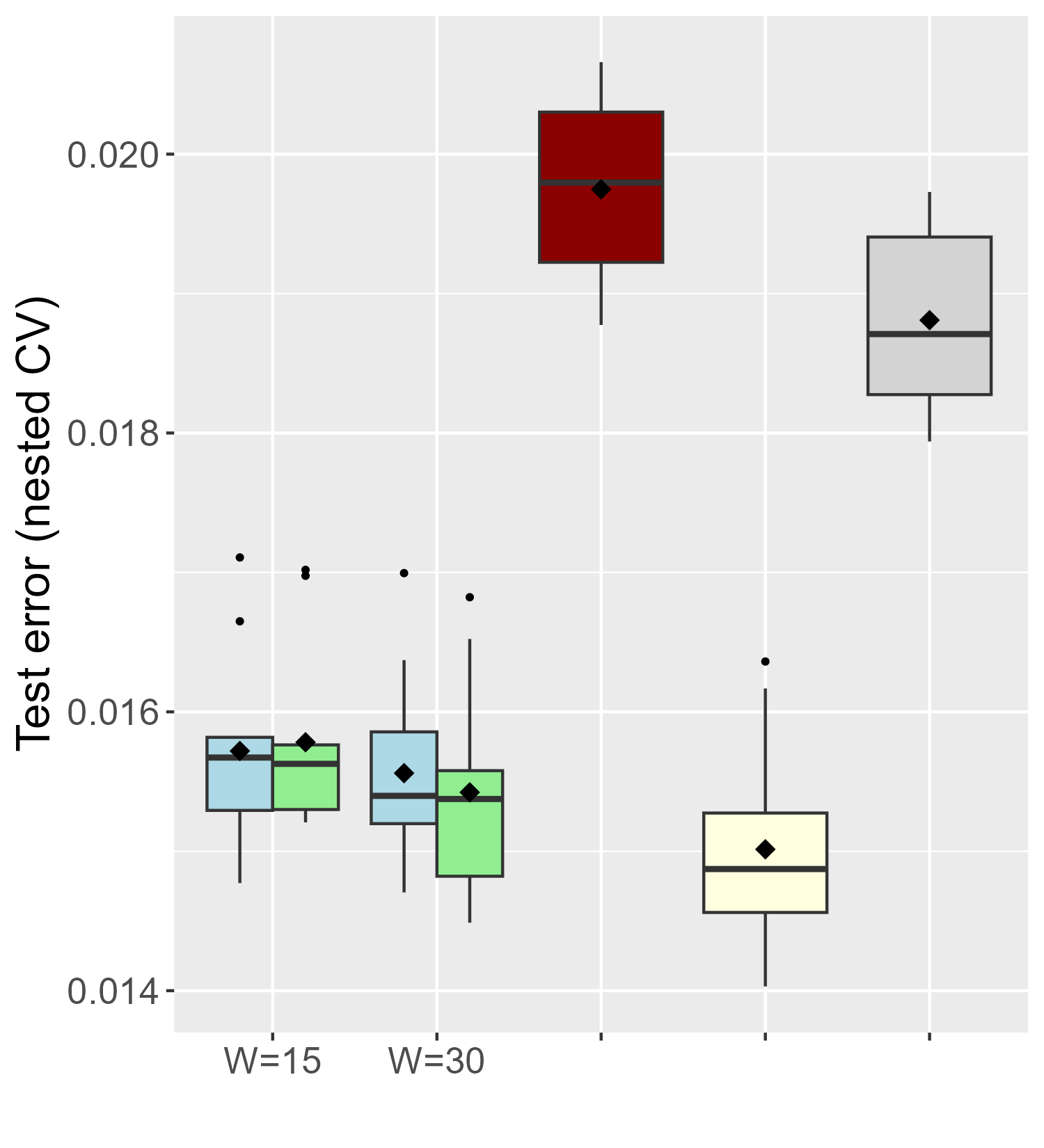}
     \end{subfigure} 
     \begin{subfigure}[b]{0.32\textwidth}
         \centering
           \caption{\calhousing}
         \includegraphics[width=\textwidth]{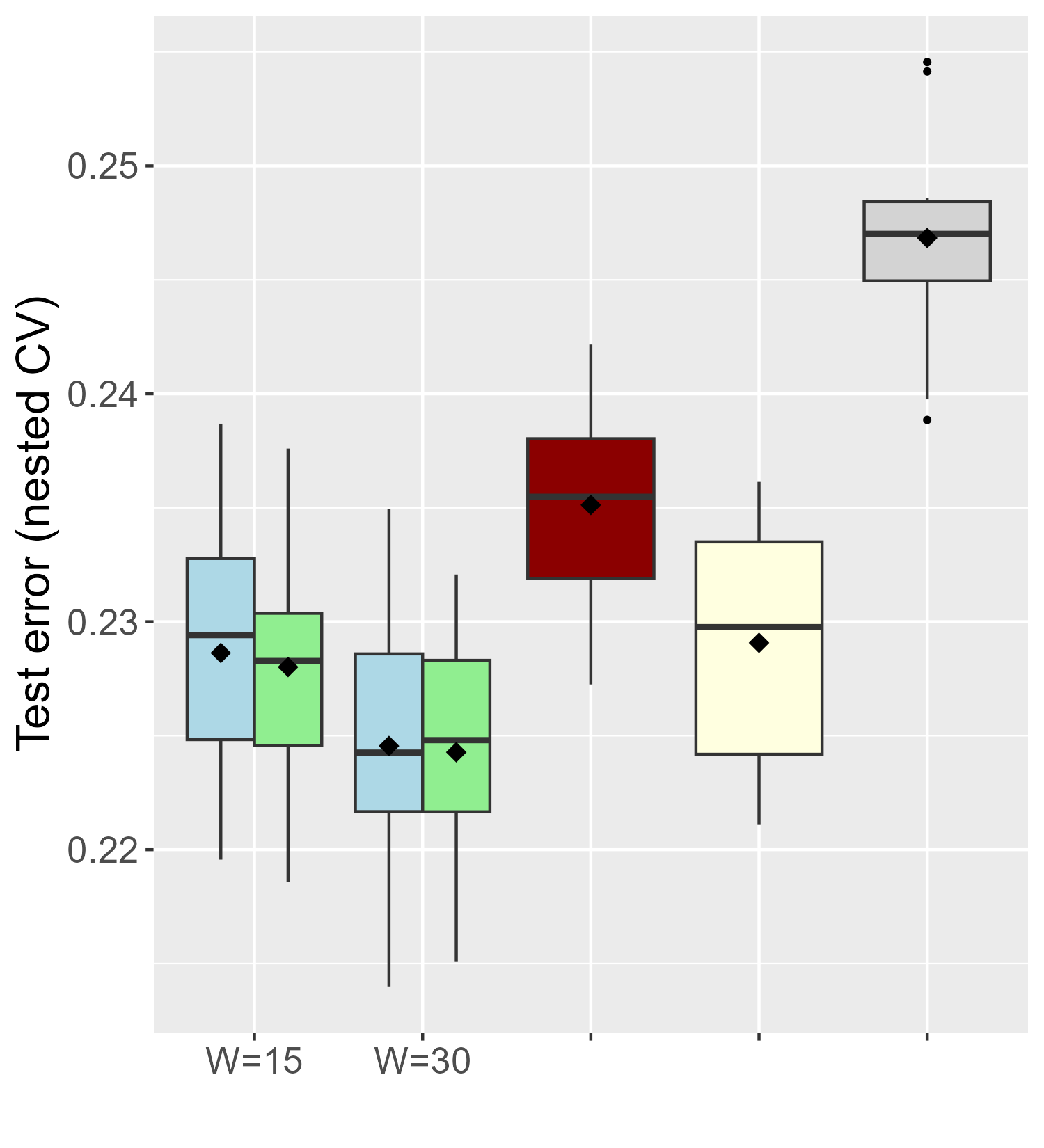}
     \end{subfigure} \hfill
       \begin{subfigure}[b]{0.32\textwidth}
         \centering
           \caption{Plot Legend}
         \includegraphics[width=\textwidth]{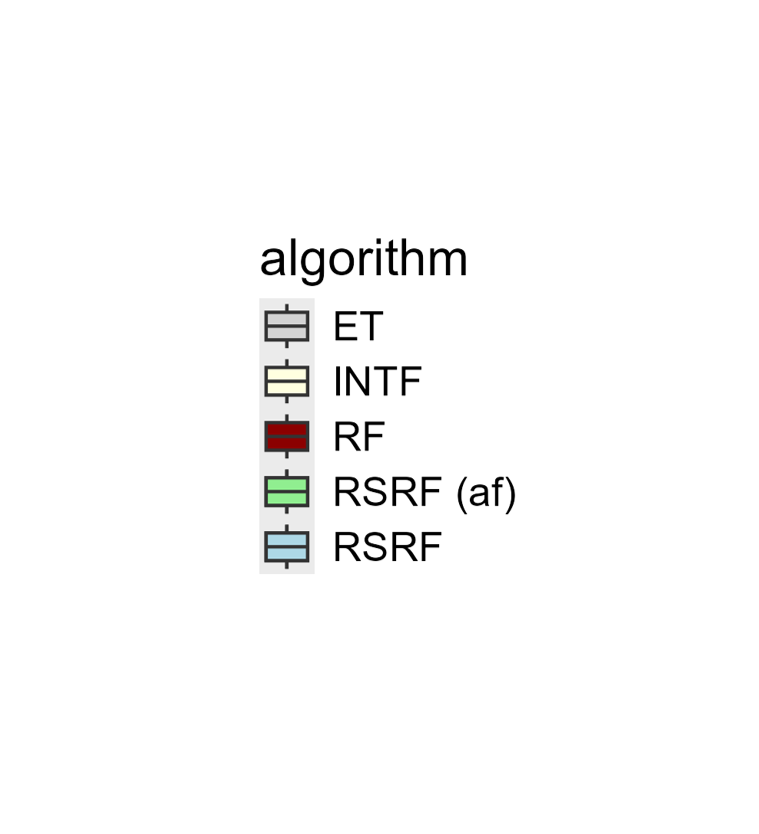}
     \end{subfigure}
        \caption{Results as shown in \Cref{fig:boxplots} in the main text, but including results for \simrsrfaf\ and the results when setting the width value to $15$ for \simrsrfnfRealdata / \simrsrfafRealdata. The boxplots show squared errors from nested CV for the real datasets. The black diamonds represent the mean of the squared errors.}
        \label{fig:boxplots_additional2}
\end{figure}

\begin{figure}[t]
     \centering
     \begin{subfigure}[b]{0.32\textwidth}
            \centering
            \caption{\concrete\ (HD)}
         \includegraphics[width=\textwidth]{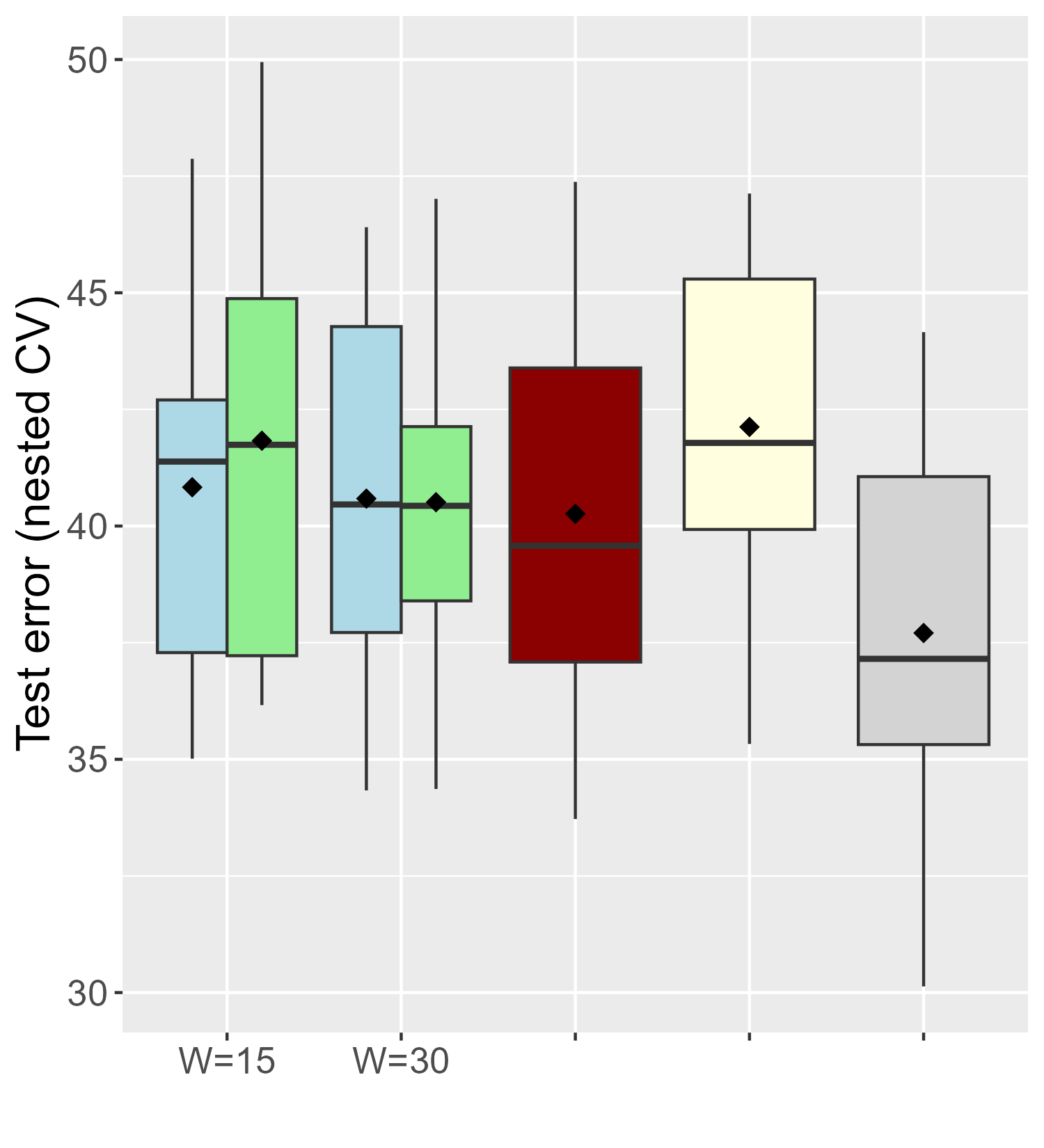}
     \end{subfigure}
     \hfill
     \begin{subfigure}[b]{0.32\textwidth}
         \centering
         \caption{\airfoil\ (HD)}
         \includegraphics[width=\textwidth]{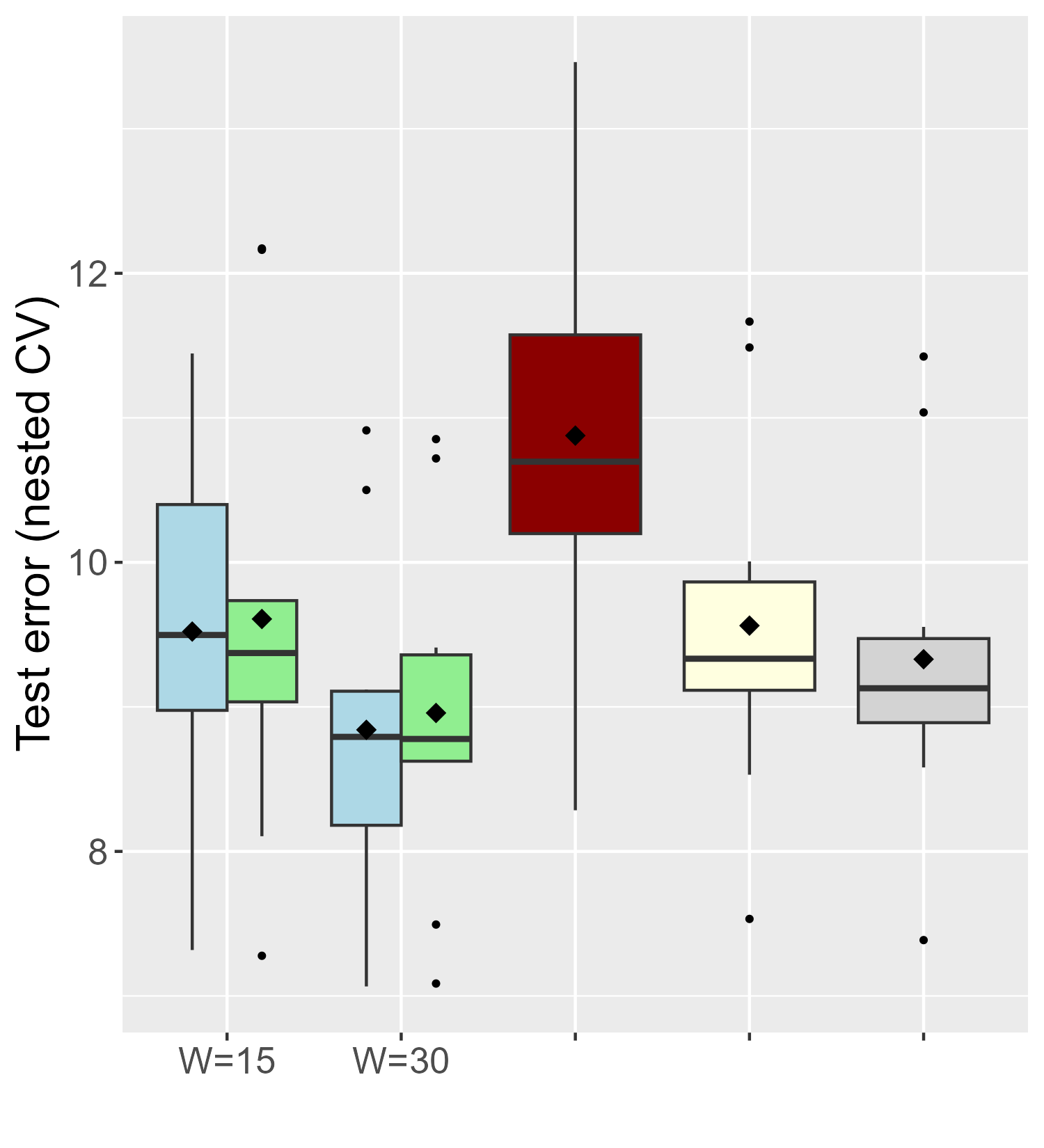}
     \end{subfigure}
     \begin{subfigure}[b]{0.32\textwidth}
         \centering
          \caption{\abalone\ (HD)}
         \includegraphics[width=\textwidth]{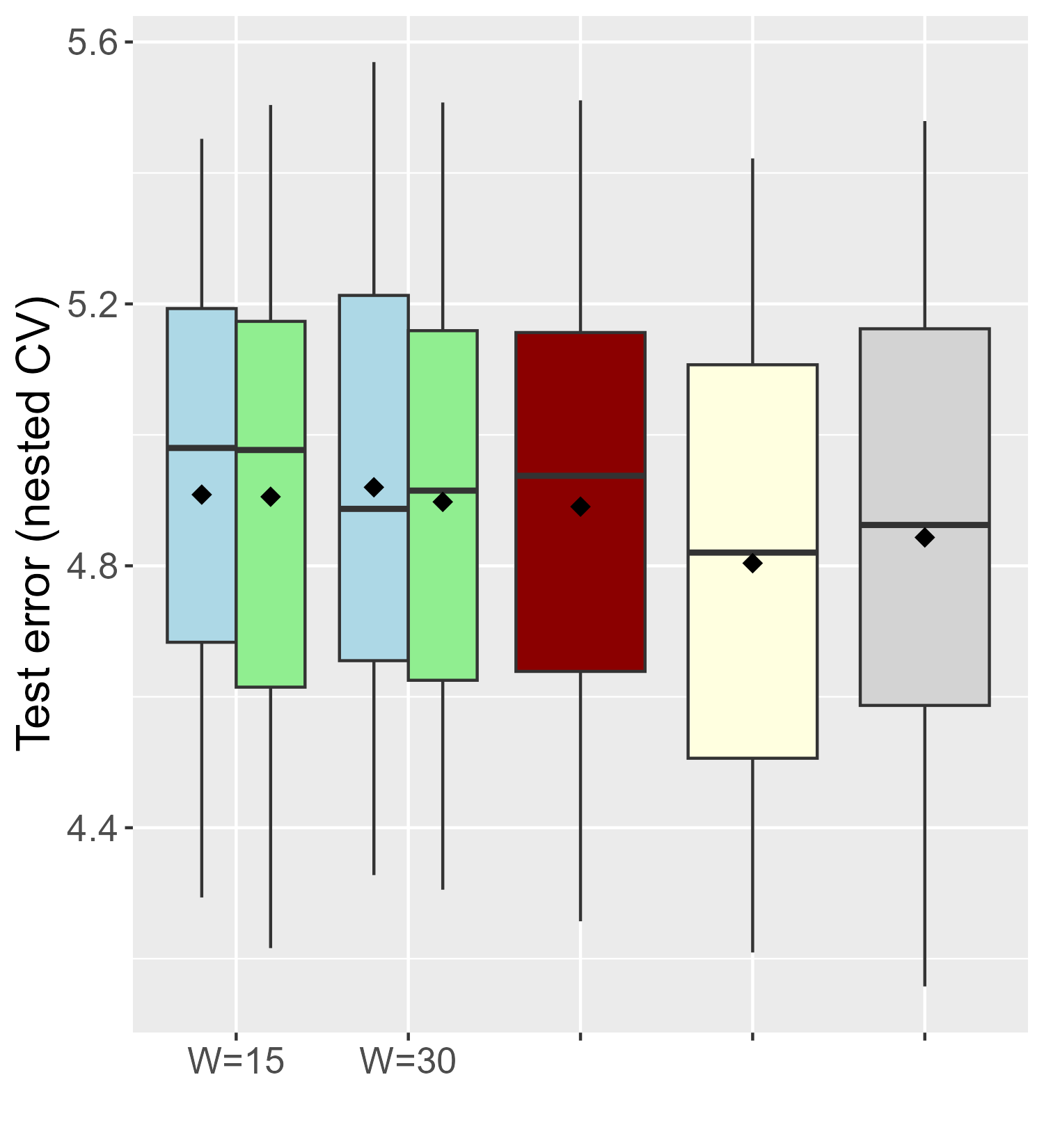}
     \end{subfigure}
       \hfill
    \begin{subfigure}[b]{0.32\textwidth}
         \centering
           \caption{\robot\ (HD)}
         \includegraphics[width=\textwidth]{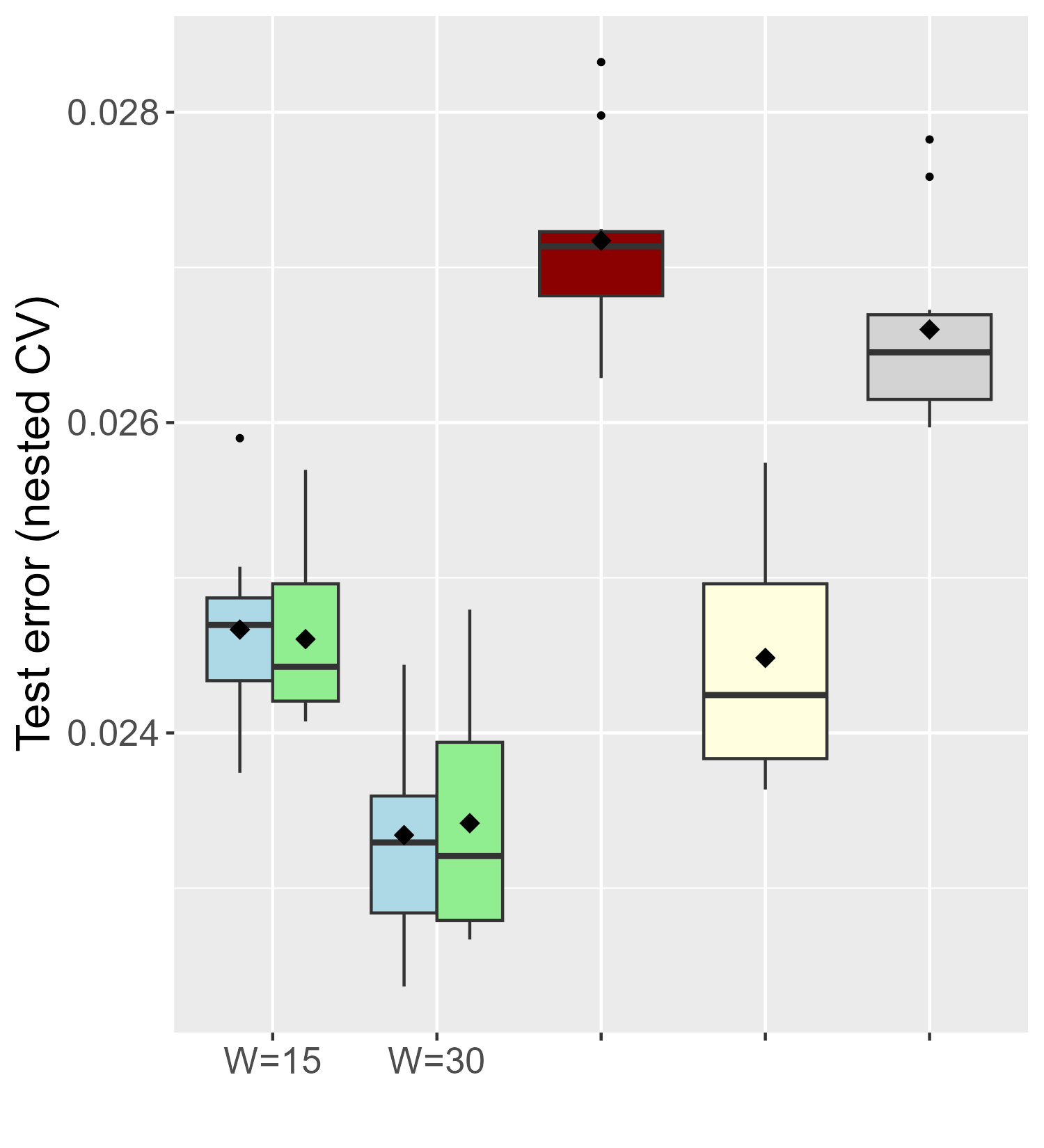}
     \end{subfigure} 
     \begin{subfigure}[b]{0.32\textwidth}
         \centering
           \caption{\calhousing\ (HD)}
         \includegraphics[width=\textwidth]{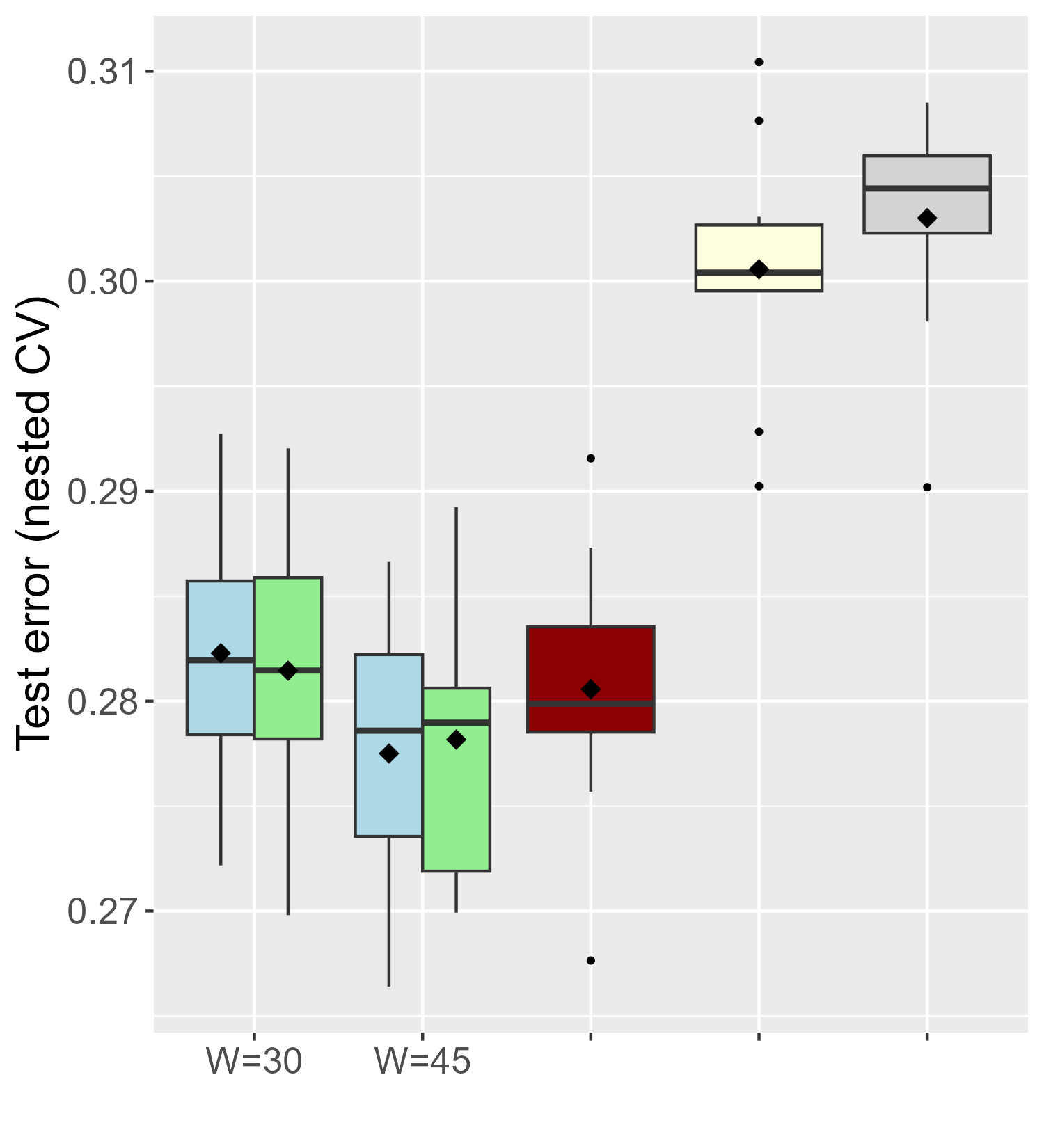}
     \end{subfigure} \hfill
       \begin{subfigure}[b]{0.32\textwidth}
         \centering
           \caption{Plot Legend}
         \includegraphics[width=\textwidth]{figures_dataexamples/legend_appendix2_print.png}
     \end{subfigure}
        \caption{Results as shown in \Cref{fig:boxplots_hd} in the main text, but including results for \simrsrfaf\ and the results when setting the width to $15$ for \simrsrfnfRealdata / \simrsrfafRealdata\ (resp. to $30$ for the \calhousing\ (HD) dataset). The boxplots show squared errors from nested CV for the real datasets. The black diamonds represent the mean of the squared errors.}
        \label{fig:boxplots_hd_additional2}
\end{figure}

We applied the algorithms on the real datasets from \Cref{sec:realdata} under additional setups. These additional settings are three-fold: First, we include the variant \simrsrfafRealdata\ as described in \Cref{sec:appendix_simulation}. Second, we include variants of the different algorithms which do not use bootstrap resampling. Third, for \simrsrfnfRealdata\ and \simrsrfafRealdata, we include results when choosing a smaller value for the width parameter.\\
In general, the parameters are as in \Cref{tab:parameters_realdata} resp. \Cref{tab:parameters_realdata_noisecov}, but with the following modifications. For \simrfnoreplaceRealdata, \simifnoreplaceRealdata, \simrsrfnfnoreplaceRealdata\ and \simrsrfafnoreplaceRealdata, subsamples instead of bootstrap samples were used (i.e. \simreplace\ is set to false). For \simetnoreplaceRealdata, neither bootstrap resampling nor subsampling was used (i.e. \simreplace\ is set to false and \texttt{sample.fraction} is set to $1$).  This is in accordance with the original version of Extremely Randomized Trees. For \simrsrfafnoreplaceRealdata, \mtrymode=\texttt{fixed} and the mtry-parameter setting is replaced by \mtryrandom=\mtryrandomcart. For \simrsrfnf, \simrsrfaf, \simrsrfnfnoreplaceRealdata\ and \simrsrfafnoreplaceRealdata, we additionally show results when the \simwidth\ parameter is set to $15$, resp. $30$ for \calhousing\ (HD).
The results can be found in \Cref{fig:boxplots_additional} and \Cref{fig:boxplots_hd_additional} for the algorithms tagged with \texttt{(noreplace)}, while the others can be found in \Cref{fig:boxplots_additional2} and \Cref{fig:boxplots_hd_additional2}.\\
In the remainder of this section, we summarize the results.\\
In \Cref{fig:boxplots,fig:boxplots_additional} resp. \Cref{fig:boxplots_hd,fig:boxplots_hd_additional}, we see that the subsampling version of Interaction Forests \simifnoreplaceRealdata\ led to better results than its bootstrap version \simif\ for \airfoil\ (HD), \calhousing\ (HD), \robot\ (HD), while for the other examples, \simifnoreplaceRealdata\ performed similarly or only slightly better than \simif. From \Cref{fig:boxplots} and \Cref{fig:boxplots_additional}, we observe that \simetnoreplaceRealdata\ performed better than \simet. In the \airfoil\ example, the bootstrap versions \simrsrfnf, \simrsrfaf\ and \simrf\ performed better than the corresponding subsampling versions, while for the other data examples the results were similar. 
Similarly, in the (HD)-settings, see \Cref{fig:boxplots_hd} and \Cref{fig:boxplots_hd_additional}, \simrsrfnf, \simrsrfaf, \simrf\ do not show big differences compared to their bootstrap counterparts. Furthermore, we do not see differences in the performance between the two variants \simrsrfnf\ and \simrsrfaf, see \Cref{fig:boxplots_additional2} and \Cref{fig:boxplots_hd_additional2}. The same observation can be made for \simrsrfnfnoreplaceRealdata\ and \simrsrfafnoreplaceRealdata\ in \Cref{fig:boxplots_additional,fig:boxplots_hd_additional}. For \simrsrfnf, \simrsrfaf, \simrsrfnfnoreplaceRealdata, \simrsrfafnoreplaceRealdata, we observe from \Cref{fig:boxplots_additional,fig:boxplots_additional2} in the non-HD case, that for the examples \robot\ and \calhousing, the larger value $W=30$ for the width parameter led to slightly better results, compared to $W=15$, while performing similarly in the other examples. For the (HD)-versions, a similar observation can be made from \Cref{fig:boxplots_hd_additional,fig:boxplots_hd_additional2} where the results for larger width were as good as the ones for smaller width, being better for \airfoil\ (HD), \robot\ (HD) and \calhousing\ (HD).\\
We recall from \Cref{fig:boxplots} and \Cref{fig:boxplots_hd} that, in both settings with and without additional feature vectors, \simrsrfnf\ and \simif\ showed a better overall performance than \simrf. Similarly, it can also be observed that \simrsrfnfnoreplaceRealdata, \simrsrfafnoreplaceRealdata\ and  \simifnoreplaceRealdata\ performed in general better than \simrfnoreplaceRealdata\ (see \Cref{fig:boxplots,fig:boxplots_additional2}). Comparing \simrsrfnfnoreplaceRealdata\ and \simifnoreplaceRealdata\ with \simrfnoreplaceRealdata, we see that the first two show better or strongly better performance in three examples in \Cref{fig:boxplots_additional} and in two examples in \Cref{fig:boxplots_additional2}. \simrfnoreplaceRealdata\ only outperformed \simifnoreplaceRealdata\ for \calhousing\ (HD). \simrfnoreplaceRealdata\ was slightly better than \simrsrfnfnoreplaceRealdata\ and \simrsrfafnoreplaceRealdata\ in the \concrete\ and \airfoil\ example in \Cref{fig:boxplots_additional}, and in one of the settings for \calhousing\ (HD) in \Cref{fig:boxplots_additional2}.\\
Finally, in addition to the observation that the version \simetnoreplaceRealdata\ is superior to \simetRealdata, we see that it also outperformed any of the other methods in three cases (irrespectively of whether the other methods used bootstrap samples or subsamples; see both \Cref{fig:boxplots,fig:boxplots_hd} and \Cref{fig:boxplots_additional2,fig:boxplots_hd_additional2}).

\begin{table}[t]
    \centering
    \begin{tabular}{D{0.6cm}|p{3cm} | p{4.7cm}| p{3.3cm}}
& & \multicolumn{2}{c}{Chosen values per dataset} \\
     & Parameter &    \abalone & \robot
\\
    \hline

        \simif\ & \texttt{npairs} & $10$$(6)$ \newline $25$$(2)$ \newline $175$, $225$ &  $175$, $225$ (3 each) \newline $125$, $250$ ($2$ each) \\
        \hline
         \simrf &  \texttt{mtry} & $3$, $4$ (5 each) & $6$, $8$ (2 each) \newline $4$, $5$ \\
         \hline 
         \simet &          \texttt{mtry}  &     \multirow{2}{*}{\small \begin{tabular}{p{0.5pt}|p{0.5pt}|p{0.5pt}|p{0.5pt}|p{0.5pt}|p{0.5pt}|p{0.5pt}|p{0.5pt}|p{0.5pt}|p{0.5pt}}
              3&3&3&3&3&4&4&4&5&6  \\
              4&4&8&9&9&5&5&8&5&5 
         \end{tabular} } &  always $8$  \\ 
      &\texttt{num.random.splits} &  & $5$$(4)$, $4$$(2)$, $8$$(2)$, $9$$(2)$ \\ \hline 
      \simrsrfnf & \simmtryrc \newline =\simmtrycartcart &
      
      $1$, $2$ ($3$ each) \newline $3$$(2)$ \newline $4$, $6$ & $7$, $8$ ($4$ each) \newline $6$$(2)$
    \end{tabular}

    \caption{The ten tuning parameters chosen in the inner-CV search when performing two times nested CV (five inner and five outer folds) for the datasets \abalone\ and \robot, and the algorithms \simif, \simrf, \simet\ and \simrsrfnf (with width $30$). In brackets, the number of occurrences is shown.}
    \label{tab:chosen_parameters}
\end{table}

\end{document}